\documentclass[11pt,letterpaper]{report}
\usepackage[width=150mm,top=35mm,bottom=35mm,bindingoffset=6mm]{geometry} 
\usepackage{multirow}
\usepackage[table,xcdraw]{xcolor}
\usepackage{amsmath,amssymb}

\usepackage[T1]{fontenc}
\usepackage{titlesec}

\titleformat{\section}
  {\normalfont\sffamily\Large\bfseries}
  {\thesection}{1em}{}
  
\newcolumntype{Z}{>{\centering\hspace{0pt}\arraybackslash}p{1.5cm}}

\usepackage[nottoc]{tocbibind}
	\settocbibname{References}
\usepackage[titletoc,title]{appendix}
\usepackage{fancyhdr}

\usepackage{multirow}
\usepackage{tabularx,tabulary}

\usepackage[newcommands]{ragged2e} 
\usepackage{booktabs} 
\usepackage{subcaption}
\usepackage{graphicx}
\usepackage{amsmath, amssymb,amsthm}
\usepackage{dsfont}
\usepackage{arydshln}
\usepackage{textcomp}
\usepackage{ragged2e}
\usepackage{xfrac} 
\usepackage{pdflscape} 
\usepackage{afterpage}
\usepackage{color} 
\usepackage{mdwlist} 
\usepackage{diagbox} 
\usepackage{titlesec} 
\usepackage{mathtools}
\usepackage{bm}
\usepackage{nccmath}
\usepackage{algorithmic}
\usepackage{cite}
\usepackage{nccmath}
\usepackage{float}
\usepackage{hyperref}
\usepackage{threeparttable}  
\usepackage{doi}
\usepackage{commath}
\PassOptionsToPackage{hyphens}{url}\usepackage{hyperref}
\hypersetup{urlcolor=blue}
\hypersetup{
    colorlinks=true,
    linkcolor={red!0!black},
    citecolor={green!0!black},
    urlcolor={blue!0!black}
}

\usepackage[capitalise,nameinlink]{cleveref}
\usepackage{enumitem}

\usepackage{placeins}
\usepackage{afterpage}
\usepackage[ruled,vlined]{algorithm2e}

\usepackage{tikz}
\usetikzlibrary{shapes.geometric, arrows}
\tikzstyle{startstop} = [rectangle, rounded corners, minimum width=2cm, minimum height=1cm,text centered, draw=black, fill=red!30]
\tikzstyle{i} = [trapezium, trapezium left angle=70, trapezium right angle=110, minimum width=3cm, minimum height=1cm, text width=10cm, text centered, draw=black, fill=blue!30]
\tikzstyle{o} = [trapezium, trapezium left angle=70, trapezium right angle=110, minimum width=3cm, minimum height=1cm, text width=5.4cm, text centered, draw=black, fill=blue!30]
\tikzstyle{process} = [rectangle, minimum width=3cm, minimum height=1cm, text width=7cm, text centered, draw=black, fill=orange!30]
\tikzstyle{processno} = [rectangle, minimum width=7cm, minimum height=1cm, text width=7cm, text centered, draw=black, fill=orange!30]
\tikzstyle{decision} = [diamond, minimum width=3cm, minimum height=1cm, text centered, draw=black, fill=green!30]
\tikzstyle{arrow} = [thick,->,>=stealth]

\usepackage[intoc]{nomencl}
\makenomenclature
\usepackage{etoolbox}
\renewcommand\nomgroup[1]{%
  \item[\bfseries
  \ifstrequal{#1}{A}{Subscripts}{%
  \ifstrequal{#1}{B}{Topology}{%
  \ifstrequal{#1}{F}{Functions}{{}}%
  \ifstrequal{#1}{C}{Frames of Reference}{{}}%
  \ifstrequal{#1}{S}{Scalars}{{}}%
  \ifstrequal{#1}{Z}{Matrices}{{}}%
  \ifstrequal{#1}{V}{Vectors}{}}}%
]}

\graphicspath{{images/}} 

\titleformat{\section}{\normalfont\Large\bfseries}{\thesection}{1em}{}

\titleformat{\chapter}{\normalfont\huge\bfseries}{\thechapter}{1em}{}

\let\oldhat\hat
\renewcommand{\hat}[1]{\oldhat{\mathbf{#1}}}

\makeatletter
\newcommand*{\transpose}{%
  {\mathpalette\@transpose{}}%
}
\newcommand*{\@transpose}[2]{%
  \raisebox{\depth}{$\m@th#1\intercal$}%
}
\makeatother

\title{\LARGE\textbf{MAPPING AND LOCALIZATION \\ USING
LIDAR FIDUCIAL MARKERS}}
\author{\Large{\color{black} YIBO LIU} \\[12pt]
\date{\vfill\normalsize{A DISSERTATION SUBMITTED TO THE FACULTY OF GRADUATE STUDIES IN PARTIAL FULFILLMENT OF THE REQUIREMENTS FOR THE DEGREE OF}\\ \vfill
\large{DOCTOR OF PHILOSOPHY} \\ \vfill
\normalsize{GRADUATE PROGRAM IN\\EARTH AND SPACE SCIENCE\\
YORK UNIVERSITY\\
TORONTO, ONTARIO\\
JANUARY 2025\\ \vfill
\copyright \hspace{1mm} YIBO LIU, 2025}}}

\begin{document}
\maketitle

\pagenumbering{roman}
\addtocounter{page}{1}


\chapter*{Abstract}
\addcontentsline{toc}{chapter}{Abstract}
Light Detection and Ranging (LiDAR) sensors are crucial for autonomous systems, yet the advancements and applications of LiDAR fiducial markers (LFMs) remain limited compared to the widespread use of visual fiducial markers (VFMs) in camera-based applications. Bridging this technological gap is of great significance for robotics and computer vision applications, but challenging due to the unstructured, sparse nature of 3D LiDAR point clouds and the 2D-specific design of previous fiducial marker algorithms. In this dissertation, a novel framework for mapping and localization using LFMs is proposed to benefit a variety of real-world applications, including the collection of 3D assets and training data for point cloud registration, 3D map merging, Augmented Reality (AR), and many more. In particular, this dissertation investigates the development of LFM and the utilization of LFM in mapping and localization from the following three aspects.
\par
First, in response to the absence of an LFM system as convenient and reliable as VFMs, an Intensity Image-based LiDAR Fiducial Marker (IFM) system is developed. The markers are thin, letter-sized sheets of paper or board with patterns compatible with popular VFM systems, without affecting the 3D geometry of the environment. A marker detection method is introduced that locates the 3D fiducials in the point cloud through the intensity image. Then, an algorithm is presented that uses the detected 3D fiducials to estimate the LiDAR 6-degree-of-freedom (6-DOF) pose.
\par
Second, to tackle the limitations of the vanilla IFM method, an improved algorithm for detecting LFMs is proposed. This enhancement extends marker detection from the single-view point cloud to the 3D map and increases the detectable distance of markers, thereby making downstream tasks such as 3D map merging feasible. In particular, a new pipeline is designed to analyze a 3D point cloud from both intensity and geometry perspectives. This approach differs from conventional 3D object detection methods, which rely solely on 3D geometric features and can only detect spatially distinguishable objects.

\par
Third, a novel approach for mapping and localization using LFMs is developed. It processes unordered, low-overlap point clouds by registering them into a single point cloud for mapping and determining their relative poses for localization. Initially, an improved adaptive threshold marker detection method is developed to provide robust detection results when viewpoints among point clouds change dramatically. Subsequently, a maximum a-posteriori (MAP) problem is formulated for registering the multiview point clouds, with a two-level graph framework developed to address it. The first-level graph, constructed as a weighted graph, is designed to efficiently and optimally infer initial values of point cloud poses from the unordered set. The second-level graph is constructed as a factor graph for the global optimization of point cloud poses, marker poses, and marker corner positions. In addition, a new dataset named Livox-3DMatch is collected using the proposed approach and incorporated into the training of the state-of-the-art learning-based multiview point cloud registration methods, resulting in evident improvements across various benchmarks.
\par
Extensive experiments with various LiDAR models in diverse indoor and outdoor scenes demonstrate the effectiveness and superiority of the proposed framework. The framework serves as an efficient, low-cost, and reliable tool for robotics and computer vision applications, including AR, 3D asset and training data collection, degraded scene reconstruction, Global Positioning System (GPS)-denied localization, and 3D map merging.


\clearpage

 \chapter*{Dedication}
 \begin{center}
\addcontentsline{toc}{chapter}{Dedication}
    \vspace*{\fill}
    \textit{To my wife, son, and parents}
    \vspace*{\fill}
\end{center}

{\chapter*{Acknowledgments}
\addcontentsline{toc}{chapter}{Acknowledgments}
First, I would like to express my sincere gratitude to my supervisor, Prof. Jinjun Shan. Although the research focus of our group is mainly on control theory, he has generously allowed me the flexibility to explore topics in robotics and computer vision. He has kindly provided me with such extensive support, including resources for study and research and personal life. I also want to thank my committee members, Prof. Armenakis and Prof. Jianguo Wang, as well as the oral exam committee members, Prof. Zhu, Prof. Ping Wang, Prof. Hu, and Prof. Waslander, for their insightful feedback, which strengthened my dissertation.
Thanks to my colleagues at SDCNLab—Ti, Marc, Hassan, Samira, Mingfeng, Penghai, Hao, Junjie, Xiaoyu, and Yida—for their kindness and support. I am grateful to my co-authors—Hunter, Amal, Kelly, Andrew, and Robert—for their valuable contributions.
I would like to acknowledge the senior researchers—Shiyuan, Adeel, Brian, Guile, Shuo, Han, Yuan, Yang, Binbin, Tongtong, and Bingbing—whose guidance shaped my research perspective.
I extend my appreciation to Haiping Wang, Hongpei Yin, Jiahe Cui, Yicong Fu, Jie Li, and Professor Hao Fan for their assistance.
Lastly, I am deeply thankful for my parents’ unwavering support and to my wife, Siyu Wu, for her contributions to our family that allowed me to focus on my research. My son, Orion Wu Liu, inspires me every day.
This journey was impossible without all of you.
\clearpage}

{
  \hypersetup{linkcolor=black,
              citecolor=black}
{\addcontentsline{toc}{chapter}{Table of Contents}
\renewcommand{\contentsname}{Table of Contents}
\tableofcontents
\listoftables
\begingroup
\renewcommand{\color}[1]{}
\listoffigures
\endgroup
\clearpage}
}

\nomenclature[V]{$\mathbf{p}_{a}$, $\mathbf{p}_{b}$}{3D coordinates of a point in the coordinate systems $\{a\}$ and $\{b\}$}
\nomenclature[V]{$\mathbf{p}_{L}$}{An observed point in the 3D point cloud}
\nomenclature[S]{\(x,y,z\)}{3D coordinates}
\nomenclature[S]{\(u,v\)}{2D image coordinates}
\nomenclature[S]{\(\theta,\phi,r\)}{Sphercial coordinates expressed in the LiDAR coordinate system}
\nomenclature[S]{\(x_{L},y_{L},z_{L}\)}{Cartesian coordinates expressed in the LiDAR coordinate system}
\nomenclature[S]{$i$}{Intensity value}
\nomenclature[S]{$t_{L}$}{Marker size of a LiDARTag}
\nomenclature[S]{$\omega$}{Thickness of the object to which the LiDATag is attached}
\nomenclature[Z]{$\mathbf{T}$}{Extrinsic matrix}
\nomenclature[Z]{$\mathbf{R}$}{Rotation matrix}

\nomenclature[Z]{$\mathbf{B}$}{Random orthonormal matrix }

\nomenclature[Z]{$\mathbf{A}\mathbf{A}^{T}$}{Random positive definite matrix}

\nomenclature[Z]{$\mathbf{H}\mathbf{A}^{T}$}{Covariance matrix of $\mathbf{q}$ and $\mathbf{q}^{\prime}$}

\nomenclature[Z]{$\mathbf{X}$}{Orthonormal matrix which is the product of $\mathbf{V}$ and $\mathbf{U}^{T}$}

\nomenclature[Z]{$\mathbf{U}\mathbf{\Lambda}\mathbf{V}^{T}$}{SVD of $\mathbf{H}$}

\nomenclature[V]{$\mathbf{t}$}{Translation vector}
\nomenclature[V]{$\mathbf{0}^{1 \times 3}$}{Zero vector with the dimension of $1 \times 3$}
\nomenclature[B]{\(\mathbb{R}^{n \times m}\)}{Set of real numbers of a dimension $n \times m$}
\nomenclature[F]{($\cdot$)}{Operation for transmitting 3D points between frames of reference}
\nomenclature[F]{$det()$}{Determinant of a matrix}
\nomenclature[B]{$SO(3)$}{Special Orthogonal Group}
\nomenclature[B]{$SE(3)$}{Lie Group}
\nomenclature[B]{$\mathfrak{so(\mathrm{3})}$}{Lie algebra associated with $SO(3)$}
\nomenclature[V]{$\xi$}{Lie algebra coordinates}
\nomenclature[F]{$\log(\cdot)$}{Matrix logarithm}
\nomenclature[F]{$Trace()$}{Trace of a square matrix}
\nomenclature[F]{$\vee$}{Map operator that finds the unique vector $\xi \in \mathbb{R}^{3\times1}$ corresponding to a given skew-symmetric matrix $\log(\mathbf{R})\in \mathbb{R}^{3 \times 3}$}

\nomenclature[C]{$\{L\}$}{LiDAR coordinate system}
\nomenclature[C]{$\{I\}$}{Image coordinate system}
\nomenclature[C]{$\{G\}$}{Global/World coordinate system}

\nomenclature[V]{$\mathbf{u}$}{Projection of $\mathbf{p}_{L}$ on the image plane}

\nomenclature[F]{$\lceil \; \rfloor$}{Rounding a value to the nearest integer}

\nomenclature[S]{$\Theta_{a}$}{Angular resolution in the azimuth direction}

\nomenclature[S]{$\Theta_{i}$}{Angular resolution in the inclination direction}

\nomenclature[S]{$u_{o}$}{Offset in the azimuth direction}

\nomenclature[S]{$v_{o}$}{Offset in the inclination direction}

\nomenclature[S]{$I_{w}$}{Width of the intensity image}
\nomenclature[S]{$I_{h}$}{Height of the intensity image}

\nomenclature[S]{$P_{w}$}{Maximum angular width of the point cloud}

\nomenclature[S]{$P_{h}$}{Maximum angular height  of the point cloud}

\nomenclature[S]{$\Theta_{h}$}{Angular resolution in the azimuth direction given by the user manual}

\nomenclature[S]{$\Theta_{v}$}{Angular resolution in the inclination direction given by the user manual}

\nomenclature[V]{$\mathbf{{u}^{r}}$}{2D pixel with range information}
\nomenclature[V]{$\mathbf{{u}_{k}}$}{2D fiducial that is detected but corresponds to an unobserved region in the raw intensity image}
\nomenclature[S]{$u_{k},v_{k}$}{2D image coordinates of $\mathbf{{u}_{k}}$}
\nomenclature[V]{$\mathbf{{p}_{k}}$}{3D point corresponding to $\mathbf{{u}_{k}}$}
\nomenclature[S]{$x_{k},y_{k},z_{k}$}{3D coordinates of $\mathbf{{p}_{k}}$}

\nomenclature[S]{\(\theta_{k},\phi_{k},r_{k}\)}{Sphercial coordinates of $\mathbf{{p}_{k}}$}

\nomenclature[S]{\(\theta_{k},\phi_{k},r_{k}\)}{Sphercial coordinates of $\mathbf{{p}_{k}}$}

\nomenclature[S]{\(\theta_{d},\phi_{d},r_{d}\)}{Sphercial coordinates of $\mathbf{{p}_{d}}$}

\nomenclature[S]{\(\theta_{u},\phi_{u},r_{u}\)}{Sphercial coordinates of $\mathbf{{p}_{u}}$}

\nomenclature[V]{$\mathbf{{p}_{u}}$, $\mathbf{{p}_{d}}$}{A pair of 3D points corresponding to the observed pixels that are symmetric about $\mathbf{{u}_{k}}$}

\nomenclature[S]{$x_{u},y_{u},z_{u}$}{3D coordinates of $\mathbf{{p}_{u}}$}

\nomenclature[S]{$x_{d},y_{d},z_{d}$}{3D coordinates of $\mathbf{{p}_{d}}$}

\nomenclature[B]{$\alpha$}{Plane determined by $\mathbf{{p}_{u}}$, $\mathbf{{p}_{k}}$, and $\mathbf{{p}_{d}}$}

\nomenclature[B]{$\beta$}{Plane where the marker is located}

\nomenclature[B]{$\textbf{\textit{l}}$}{Intersection of $\alpha$ and $\beta$}

\nomenclature[V]{$\mathbf{{O}_{L}}$}{Origin of the LiDAR coordinate system}

\nomenclature[F]{$\mathrm{diag}()$}{Operation for creating a diagonal matrix}

\nomenclature[S]{$\mu$}{Ratio}

\nomenclature[Z]{$\mathbf{{M}_{1}}$, $\mathbf{{M}_{2}}$}{Weight matrices of $\mathbf{{p}_{d}}$ and $\mathbf{{p}_{u}}$}

\nomenclature[V]{$\mathbf{f}$}{3D coordinates of a fiducial}

\nomenclature[B]{$\mathcal{{P}_{L}}$}{Set of detected 3D fiducials expressed in $\{L\}$}

\nomenclature[B]{$\mathcal{{P}_{W}}$}{Set of detected 3D fiducials expressed in $\{W\}$}

\nomenclature[V]{$\mathbf{{f}_{1}}, \ \cdots, \ \mathbf{{f}_{n}}$}{Detected 3D fiducials expressed in $\{L\}$}


\nomenclature[V]{$\mathbf{{f}_{1}}^{\prime}, \ \cdots, \ \mathbf{{f}_{n}}^{\prime}$}{Predefined 3D fiducials expressed in $\{W\}$}

\nomenclature[F]{$I(\mathbf{p})$}{Function of intensity with respect to $\mathbf{p}$}

\nomenclature[F]{$\hat{I}(\mathbf{p})$}{Approximated function of intensity with respect to $\mathbf{p}$}

\nomenclature[Z]{$\mathbf{A}$}{Coefficient matrix of $\hat{I}(\mathbf{p})$}

\nomenclature[S]{$b$}{Coefficient of $\hat{I}(\mathbf{p})$}

\nomenclature[S]{$\bar I$}{Mean of the intensity values of points in $\mathcal{{P}_{I}}$}

\nomenclature[V]{$\mathbf{{p}_{0}}$}{A given 3D point}

\nomenclature[B]{$\mathcal{{P}_{I}}$}{Point set composed of the neighbouring points around $\mathbf{{p}_{0}}$}

\nomenclature[V]{$\mathbf{{p}_{1}},\cdots,\mathbf{{p}_{n}}$}{Neighbouring points around $\mathbf{{p}_{0}}$}

\nomenclature[V]{$|\nabla I|$}{Intensity gradients}

\nomenclature[S]{$a$}{Side length of the marker}
\nomenclature[S]{$t_{M}$}{Thickness of the marker}

\nomenclature[S]{$l,w,h$}{Length, width, and height of the OBB}
\nomenclature[S]{$L_{OBB}$}{Cuboid diagonal length of the OBB}

\nomenclature[S]{$S_{OBB}$}{Area of the 2D OBB}

\nomenclature[Z]{$\mathbf{T}_{OBB}$}{Pose that transforms 3D points from $\{W\}$ to the OBB frame}

\nomenclature[S]{$t_{b}$}{Amplification factor}

\nomenclature[B]{$\mathbf{P}$}{Intermediate plane}

\nomenclature[V]{$\mathbf{{p}_{G}}$}{Point transmitted to the origin of $\{W\}$}

\nomenclature[V]{$\mathbf{C}$}{Coefficient vector}
\nomenclature[V]{$\mathbf{I}_{in}$}{Response variable vector}
\nomenclature[V]{$\mathbf{E}$}{Regression coefficient vector}
\nomenclature[S]{$b$}{Intercept}

\nomenclature[Z]{$\mathbf{D}$}{Design matrix}

\nomenclature[B]{$\mathcal{{P}}_{j}$}{Set of points enclosed within the $j$-th OBB}

\nomenclature[B]{$\mathcal{P}^{G}_{j}$}{Set of points belonging to $\mathcal{{P}_{j}}$ transmitted to the origin of $\{W\}$}

\nomenclature[Z]{$\mathbf{T}_{in}$}{Transmission from $\{W\}$ to the intermediate plane}

\nomenclature[Z]{$\mathbf{I}$}{Identity matrix}

\nomenclature[B]{$\mathcal{P}^{\prime}_{j}$}{Set of points on the intermediate plane}

\nomenclature[B]{$\mathcal{P}_{loam}$}{Set of sampled data used by LOAM}
\nomenclature[B]{$\mathcal{P}_{Traj}$}{Set of sampled data used by Traj LO}

\nomenclature[B]{$\mathcal{C}_{loam}$}{Set of corner features extracted by LOAM}

\nomenclature[B]{$\mathcal{S}_{loam}$}{Set of plane features extracted by LOAM}

\nomenclature[S]{$t_{cor}$,$t_{sur}$}{Thresholds for judging corner and plane features}

\nomenclature[F]{$\kappa(\mathbf{p}_{i})$}{Function to calculate the curvature}

\nomenclature[F]{$\mathbf{N}(\mathbf{p}_{i})$}{Function to calculate the normal}

\nomenclature[F]{$\mathbf{T}(t)$}{Function of time representing the LiDAR trajectory}

\nomenclature[F]{$\tau(t_{i})$}{Function of time representing the LiDAR trajectory at the $i$-th time segment}

\nomenclature[B]{$\mathcal{E}_{reg}$}{Registration energy}

\nomenclature[B]{$\mathcal{E}_{kine}$}{Kinematics energy}

\nomenclature[B]{$\mathcal{E}_{marg}$}{Marginalization energy}

\nomenclature[B]{$Q$}{Queue for saving detected markers}

\nomenclature[S]{$Q_{l}$}{Length of $Q$}

\nomenclature[B]{$Q_{temp}$}{Temporary queue for saving detected markers}

\nomenclature[S]{$Q_{temp,l}$}{Length of $Q_{temp}$}

\nomenclature[B]{$\mathcal{I}$}{Raw intensity image}

\nomenclature[S]{$\lambda$, $\lambda^{*}$}{Threshold for binarization and Optimal threshold for binarization}

\nomenclature[S]{$Thr$}{Threshold for calculating recall}

\nomenclature[S]{$S$}{Search scope}
\nomenclature[S]{$\delta$}{Step size}

\nomenclature[F]{$\Psi( )$}{Function of binarization}

\nomenclature[F]{$\Gamma( )$}{Function of marker detection}

\nomenclature[B]{$f_{i}$}{LiDAR point cloud acquired at the $i$-th viewpoint}

\nomenclature[C]{$\{F\}_{i}$}{Local coordinate system of $f_{i}$. $\{F\}_{i}$ is aligned with $\{L\}$ when the frame is acquired}

\nomenclature[C]{$\{M\}_{j}$}{Marker coordinate system of $j$-th marker}

\nomenclature[V]{$\mathbf{p}^{j,s}$}{3D coordinates of the $s$-th coroner of the $j$-th marker expressed in $\{W\}$}

\nomenclature[V]{$^{j}\mathbf{p}^{j,s}$}{3D coordinates of the $s$-th coroner of the $j$-th marker expressed in $\{M\}_{j}$}

\nomenclature[V]{$\mathbf{q}_{j}$, $\mathbf{q}^{\prime}_{j}$}{Centroid-aligned coordinates}

\nomenclature[V]{$_{i}\mathbf{p}^{j,s}$}{3D coordinates of the $s$-th coroner of the $j$-th marker expressed in $\{F\}_{i}$}

\nomenclature[Z]{$\mathbf{T}_{i}$}{Pose that transforms 3D points from $\{F\}_{i}$ to $\{W\}$}

\nomenclature[Z]{$\mathbf{T}^{j}$}{Pose that transforms 3D points from $\{M\}_{j}$ to $\{W\}$}

\nomenclature[Z]{$\mathbf{T}^{j,k}$}{Pose that transforms 3D points from $\{M\}_{j}$ to $\{M\}_{k}$}

\nomenclature[Z]{$\mathbf{T}_{i,m}$}{pose that transforms 3D points from $\{F\}_{i}$ to $\{F\}_{m}$}

\nomenclature[Z]{$\mathbf{T}^{j}_{i}$}{Pose that transforms 3D points from $\{M\}_{j}$ to $\{F\}_{i}$}

\nomenclature[B]{$\mathcal{Z}$}{Set of measurements}

\nomenclature[B]{$\Theta$}{Set of variables}

\nomenclature[B]{$\Theta^{*}$}{Set of optimal variables}

\nomenclature[F]{$P(|)$}{Posterior probability}

\nomenclature[F]{$h_k()$}{Measurement function}

\nomenclature[B]{$z_k$}{Measurement}

\nomenclature[S]{$\|e\|_{\Sigma}^2$}{Squared Mahalanobis distance}

\nomenclature[Z]{$\Sigma$}{Covariance matrix}
\nomenclature[F]{$\ominus$}{Straight subtraction for elements}

\nomenclature[F]{$\mathrm{Rot}(\mathbf{T})$}{Extracting the rotation matrix from $\mathbf{T}$}

\nomenclature[F]{$\mathrm{Trans}(\mathbf{T})$}{Extracting the translation vector from $\mathbf{T}$}

\nomenclature[C]{$\{a\}$, $\{b\}$}{Two random coordinate systems}

\nomenclature[S]{${e}_{pp}$}{Point-to-point error}

\nomenclature[]{\(\)}{}

\nomenclature[]{\(\)}{}

\nomenclature[]{\(\)}{}

\nomenclature[]{\(\)}{}

\nomenclature[]{\(\)}{}

\nomenclature[]{\(\)}{}

\nomenclature[]{\(\)}{}

\nomenclature[]{\(\)}{}

\nomenclature[]{\(\)}{}

\nomenclature[]{\(\)}{}

\nomenclature[]{\(\)}{}

\nomenclature[]{\(\)}{}

\nomenclature[]{\(\)}{}

\nomenclature[]{\(\)}{}

\nomenclature[]{\(\)}{}

\nomenclature[]{\(\)}{}

\nomenclature[]{\(\)}{}

\nomenclature[]{\(\)}{}

\nomenclature[]{\(\)}{}

\nomenclature[]{\(\)}{}

\nomenclature[]{\(\)}{}

\nomenclature[]{\(\)}{}

\nomenclature[]{\(\)}{}

\nomenclature[]{\(\)}{}

\nomenclature[]{\(\)}{}

\nomenclature[]{\(\)}{}

\nomenclature[]{\(\)}{}

\nomenclature[]{\(\)}{}

\nomenclature[]{\(\)}{}

\nomenclature[]{\(\)}{}

\nomenclature[]{\(\)}{}

\nomenclature[]{\(\)}{}

\printnomenclature[2.3cm]

\clearpage

{\chapter*{List of Acronyms}
\addcontentsline{toc}{chapter}{List of Acronyms}
\begin{tabbing}

LiDAR ~~~~~~~~~~~~~~~~\= Light Detection and Ranging\\

LFMs \> LiDAR fiducial markers\\

VFMs \> Visual fiducial markers\\

IFM \> Intensity Image-based LiDAR Fiducial Marker\\

6-DOF \> 6-degree-of-freedom\\

MAP \> Maximum a-posteriori\\

GPS \> Global Positioning System\\

SfM \> Structure-from-Motion\\

SLAM \> Simultaneous Localization and Mapping\\

RGB-D \> Red Green Blue-Depth\\

IMUs \> Inertial Measurement Units\\

AR \> Augmented Reality\\

LOAM \> LiDAR Odometry and Mapping\\

LO \> LiDAR Odometry\\

ICP \> Iterative Closest Point\\

ROS \> Robot Operating System\\

FoV \> Field of View\\

SVD \> Singular Value Decomposition\\

MoCap \> OptiTrack Motion Capture\\

CPU \> Central Processing Unit\\

OBB \> Oriented Bounding Box\\

PCA \> Principal Component Analysis\\
 
PC \> Primary Component\\

LM \> Levenberg-Marquardt\\

ID \> Identification\\
RTK \> Real-Time Kinematic\\
RMSEs \> Root-mean-square errors\\

$\mathrm{RMSE}_{T}$ \> Root-mean-square error of translation\\

$\mathrm{RMSE}_{R}$ \> Root-mean-square error of translation rotation\\

GPU \> Graphics Processing Unit\\

CD \> Chamfer Distance\\

RR \> Registration Recall\\

\end{tabbing}

\clearpage
}

\pagenumbering{arabic}

\chapter{Introduction \label{Intro}}
\section{Motivation}
Light Detection and Ranging (LiDAR) sensors are crucial ranging sensors for autonomous systems \cite{loam,traj,rangenet}. In contrast to another vital sensor, the camera, the development and application of LiDAR fiducial markers (LFMs) are rare. Specifically, research on Visual Fiducial Marker (VFM) systems \cite{wang,olson,ap3,aruco,cctag}, fiducial marker systems developed for cameras, has a long history, with extensive experience and significant research achievements accumulated. VFMs provide environments with controllable artificial features, making feature extraction and matching simpler and more reliable. VFM systems have been applied in various camera-based applications, including Augmented Reality (AR) \cite{ar}, human-robot interaction \cite{wang,olson}, navigation \cite{yibo1,yibo2,liao}, multi-sensor calibration \cite{kalibr,lt2}, Structure-from-Motion (SfM) \cite{munoz,qingdao}, and visual Simultaneous Localization and Mapping (SLAM) \cite{munoz2019,shuo,shuo2}. \par
LiDAR-based mapping and localization \cite{loam,lloam,sdk,traj,sghr,multiway,mdgd} are fundamental in robotics and computer vision. Similar to visual SLAM \cite{shuo,shuo2,shuo3} and SfM \cite{munoz,qingdao} in camera-based mapping and localization, there are two main categories of solutions. The first category is LiDAR-based SLAM approaches \cite{loam,lloam,sdk,traj}, which process consecutive LiDAR scans in real time to map the scene and localize the LiDAR sensor simultaneously. The second category is multiview point cloud registration methods \cite{sghr,multiway,mdgd}. Compared with SLAM approaches that process LiDAR scans sequentially, these methods handle a set of unaligned point clouds all at once in an offline manner. The mapping is achieved by registering the multiview point clouds into one complete point cloud, and localization is accomplished by finding the relative pose between point clouds.
To clarify the motivation, this dissertation will answer three questions in this field:
\begin{itemize}
\item Given the success of VFM in camera-based applications, why is it still desirable to exploit the LiDAR sensor and LFM?

\item Considering the existence of previous LiDAR fiducial objects, why is it favorable to develop a new type of LiDAR fiducial marker system?

\item Given that previous LiDAR-based mapping and localization methods do not rely on fiducial markers, why is it advantageous to exploit LFM in this dissertation?
\end{itemize}
\par
\noindent\textbf{Response to the First Question.} There are two reasons why LiDAR sensors and LFM are irreplaceable by cameras and VFM. 
First, as a ranging sensor, the role of LiDAR in autonomous systems \cite{waabi,waymo,cd} cannot be replaced by the camera. While Red Green Blue-Depth (RGB-D) cameras can also measure depth, their effective range is usually less than 10 meters \cite{3dmatch,eth,scan}, whereas LiDAR can scan objects several hundred meters away \cite{loam,traj,rangenet}.
Additionally, due to the unique modality of LiDAR point clouds, data captured by LiDAR cannot be replaced by data acquired from cameras. Extensive learning-based models \cite{rangenet,cd,mending} rely on LiDAR data for training.
Second, unlike VFM detection, which is sensitive to ambient light \cite{ap3,aruco,cctag}, LiDAR-based object detection is robust to unideal illumination conditions \cite{lt,iilfm}, such as purely dark or overexposed environments. Third, VFM detection suffers from rotational ambiguity \cite{munoz2018,qingdao,munoz2019} and requires discarding ambiguous measurements, whereas LiDAR-based planar object pose estimation is free from this issue \cite{lt}.
\par

\noindent\textbf{Response to the Second Question.}
There have been some fiducial objects developed for LiDAR sensors \cite{cal,cal2,a4,lt} for calibration purposes. There are three reasons why it is desired to develop a new type of LFM system. First, most existing LiDAR fiducial objects \cite{cal,cal2,lt,a4} are designed for specific LiDAR models and are not generalizable to a wide range of solid-state and mechanical LiDARs. Second, unlike popular VFMs such as AprilTag \cite{ap3} and ArUco \cite{aruco}, which are thin-sheet objects attached to surfaces without affecting the 3D geometry of the environment, most existing LiDAR fiducial objects \cite{cal,cal2,lt} are thick boards placed on tripods, acting as additional 3D objects that alter the environment. These extra 3D objects are undesirable in applications such as scene reconstruction and data collection. Third, a typical calibration board \cite{cal,cal2,a4} does not have complex patterns generated from elaborately designed encoding-decoding algorithms like VFMs \cite{wang,olson,ap3,aruco,cctag}, which are robust against false positives and negatives. 
\par
Thus, the first major objective of this dissertation is to develop a unified LFM system that is applicable to a variety of LiDAR sensors, while being as convenient to use as VFMs: printed on sheets of paper/board and attached to surfaces without affecting the 3D environment. The unified system is desirable because we do not need to redesign the algorithm/marker every time the sensor is changed, and it is beneficial for sensor fusion of different models. Additionally, the system should incorporate a reliable encoding-decoding algorithm, like VFMs \cite{wang,olson,ap3,aruco,cctag}, for robust detection. 
\par
\noindent\textbf{Response to the Third Question.}
The existing point cloud registration methods, including geometry-based methods \cite{teaser,kiss,lloam,random} and learning-based methods \cite{sghr,pre,mdgd,se3et,geotransformer}, rely on shared geometric features between point clouds to align them.
Hence, these methods struggle in scenes with scarce shared features, such as extremely low overlap cases and degraded scenes \cite{degradation}. 
Consequently, the existing methods \cite{mdgd,sghr,se3et,geotransformer,teaser} are unsuitable for tasks such as capturing the complete 3D shape of a novel object from a sparse set of scans taken from dramatically different viewpoints, gathering data from unseen scenarios for training, reconstructing degraded scenes, or merging large-scale 3D maps with low overlap.
In contrast, the introduction of LFM makes feature extraction and matching simpler and more reliable. Thanks to this, the proposed method is robust to any unseen scenarios with extremely low overlap, making it a convenient, efficient, and low-cost tool for diverse applications that pose significant challenges to existing methods.
To help readers understand this point, a case with similar motivation in a camera-based application is presented: Although marker-free SfM methods \cite{colmap,dust3r} exist, when constructing the Omniobject3D \cite{omni} dataset—a collection of posed multiview images of real-world objects—VFM-based SfM is employed to efficiently and robustly estimate the relative pose between images using a chessboard composed of AprilTags \cite{ap3}. 
%
One may consider adding external sensors, such as Inertial Measurement Units (IMUs), to the LiDAR sensor to tackle challenging textureless scenes or cases with dramatic viewpoint changes. However, this involves multi-sensor calibration and synchronization. In comparison, LFMs made from sheets of paper or board are low-cost, easy to use, and reliable. This motivation is the same as that of VFM-based SfM \cite{munoz,qingdao} and visual SLAM \cite{munoz2019,shuo} studies.
\par
In summary, the main objective of this dissertation is to develop a framework for mapping and localization using LFMs. This framework will serve as a fundamental tool for robotics and computer vision, enabling a variety of real-world applications, including AR, 3D asset collection from sparse scans, training data collection in unseen scenes, reconstruction of degraded scenes, localization in GPS-denied environments, and merging large-scale low overlap maps. In particular, considering that training data collection for point cloud registration is time-consuming and expensive, and existing datasets \cite{3dmatch,eth,scan} are limited to indoor scenes captured using RGB-D cameras, it is desirable to construct a new training dataset in a straightforward and low-cost manner to benefit learning-based point cloud registration methods \cite{sghr,mdgd}. 
\section{Related Work and Challenges}
\label{Challenges}
\subsection{Visual Fiducial Markers}
VFMs \cite{wang,olson,ap3,aruco,cctag,arc,arc2} are particular objects, commonly planar, that provide the environment with controllable artificial features. They were originally developed for augmented reality \cite{ar}, but have been widely applied in various robotics and computer vision applications, such as human-robot interaction \cite{wang,olson}, navigation \cite{yibo1,yibo2,liao}, multi-sensor calibration \cite{kalibr,lt2}, SfM \cite{munoz,qingdao}, and visual SLAM \cite{munoz2019,shuo,shuo2}. The previous research on the VFM system mainly focuses on the following four aspects: (1) A higher detection rate. Line segmentation methods have been continuously upgraded alongside VFM systems \cite{olson,wang}. In addition to line segmentation, non-square shape detection methods, such as ellipse detection algorithms \cite{arc,arc2,cctag}, have also emerged and made valuable contributions. These advancements aim to improve the detection of line segments, candidate quads \cite{ap3,aruco}, or circles \cite{arc,arc2,cctag}, thereby enhancing the marker detection rate under varying ambient light and/or motion blur. (2) A lower false positive rate. The encoding-decoding algorithms of the latest VFM systems \cite{ap3,aruco} have been upgraded compared to earlier generations \cite{wang,olson}, making the identification of candidate markers more reliable and reducing false positives during decoding. (3) A lower computational time. Again, through the amelioration of methods for graphic segmentation and algorithms adopted in encoding-decoding algorithms, the VFM systems are accelerated. For instance, the speed of the third generation of Apirltag \cite{ap3} is almost 5 times faster than that of the second generation \cite{wang}. (4) Resolving the rotation ambiguity problem. The rotational ambiguity means a planar marker could project onto the same pixels from two different poses when the perspective effect is weak \cite{ippe}. Much research \cite{yibo,ch2020} has been conducted to solve this problem by adding external sensors or changing the maker from 2D to 3D. Unfortunately, rotational ambiguity is still an unresolved problem for 2D VFM detection without external information.
\par
However, the aforementioned VFM algorithms were developed for 2D images, not 3D LiDAR point clouds. Given the differences between 2D image and 3D point cloud modalities, it is challenging—though beneficial and desirable—to transfer the advancements made in VFM research to LiDAR-based applications \cite{lt}.
\subsection{LiDAR Fiducial Objects}
\subsubsection{LiDAR Fiducial Object Patterns}
Most existing LiDAR fiducial objects \cite{cal,cal2,a4} are designed for calibration purposes. Fig. \ref{tags}(a) shows a typical calibration board, a thick board with holes and/or regions covered by high-intensity materials. While a calibration board can provide fiducials (holes, corners, and high-intensity regions) and it is feasible to assign specific indexes to the fiducials, a calibration board is fundamentally different from fiducial marker systems \cite{ap3,aruco}. This is because the design of calibration board patterns does not incorporate the core element of a fiducial marker system—the encoding-decoding algorithm. Consequently, calibration boards \cite{cal,cal2} do not support rich patterns with unique identifications (IDs) and the robust detection provided by encoding-decoding algorithms \cite{wang,olson,ap3,aruco,cctag,arc,arc2} as shown in Figs. \ref{tags}(b)(c).
The development of VFMs and LiDAR fiducial objects had been disjointed until the proposal of LiDARTag \cite{lt}, as illustrated in Fig. \ref{tags}(d).
LiDARTag, as the first fiducial marker system for LiDAR sensors, integrates the encoding-decoding algorithm of AprilTag 3 \cite{ap3} into LiDAR fiducial object detection. Consequently, it inherits the rich patterns and robust detection features of AprilTag 3. However, LiDARTag only studies square markers with patterns belonging to AprilTag 3 and does not support square markers with different pattern sources, for instance, ArUco \cite{aruco}, or non-square markers, such as CCTag \cite{cctag}. Developing a general algorithm that fully incorporates the achievements of VFM research while not requiring a specific marker pattern and shape is challenging. As shown in Fig. \ref{tags}(e), the other widely used LiDAR fiducial objects are prisms \cite{station}, which are designed to reflect laser beams and produce very high intensity, making them easy to detect. However, the cost of prisms is much higher than that of VFMs, and they also do not contain the encoding-decoding algorithm to support rich patterns that are robust to false positives/negatives.
\subsubsection{LiDAR Fiducial Object Placement}
\begin{figure}[H] 
	\centering
	\includegraphics[width=0.8\linewidth]{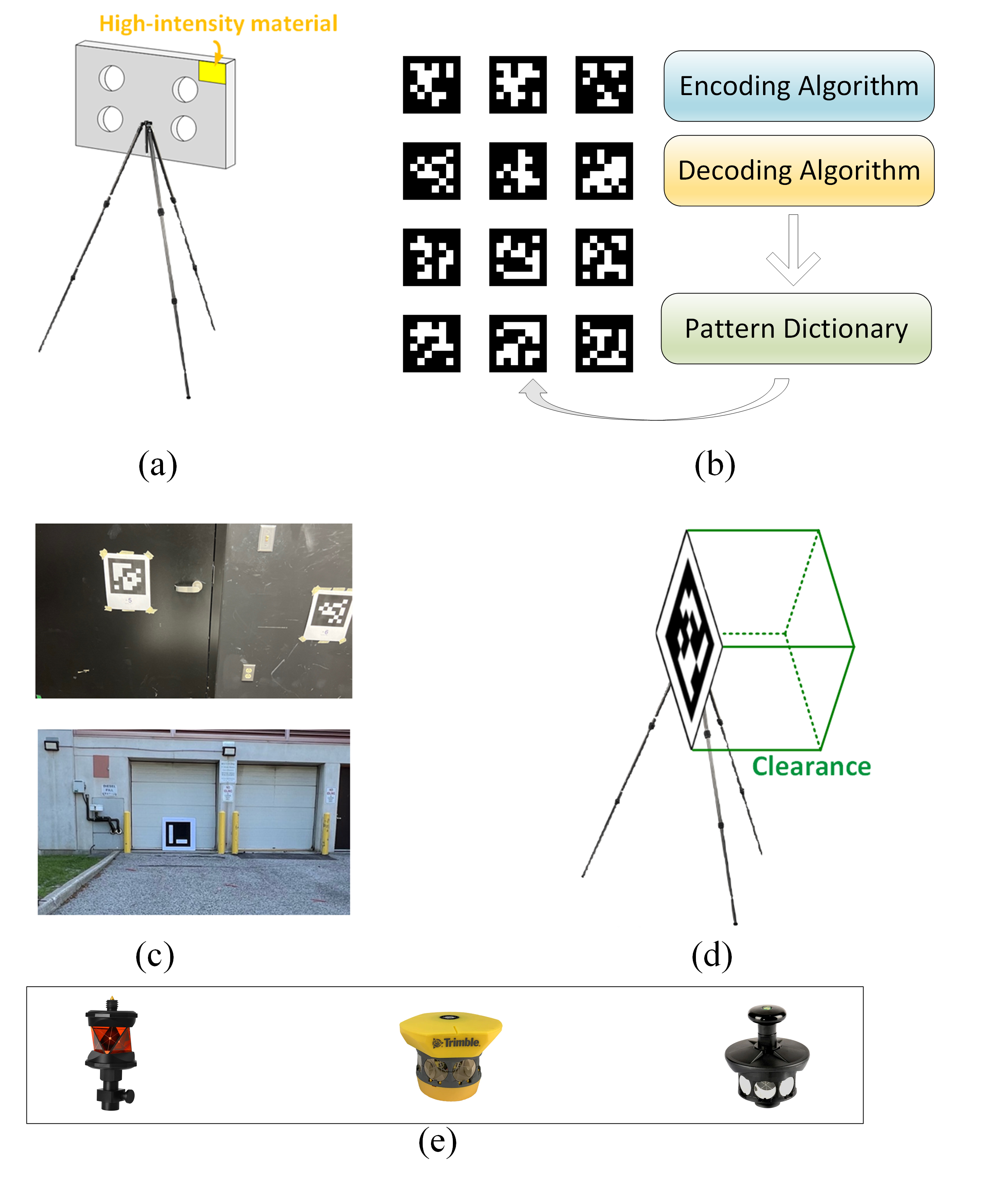}
	\caption{ Comparison of (a) a typical calibration board, (b)(c) VFM, (d) LiDARTag, and (e) prisms in terms of object patterns and placement. 
 }
	\label{tags}
\end{figure}

3D LiDAR point clouds are sparse and unstructured compared to 2D images, making it challenging to develop an LFM system \cite{lt}. Existing LiDAR fiducial objects \cite{cal1,cal2}, including LiDARTag \cite{lt}, use conventional geometry-based clustering methods \cite{edge,single,rangenet}, requiring spatial distinction. 
Thus, as seen in Fig. \ref{tags}(a) and (d), they are often placed on tripods, introducing extra 3D objects that impact the environment. 
For instance, LiDARTag adopts the single-linkage clustering algorithm \cite{single} and requires that the marker must have at least $t_{L}\sqrt{2}/4$ clearance around it, where $t_{L}$ represents the marker's size. If the marker is attached to a wall or a box, it is necessary to ensure that $\omega > t_{L}\sqrt{2}/4$, where $\omega$ is the thickness of the marker's 3D object. Now, consider the case where ten LiDARTags are placed in an indoor environment. This situation would result in a room crowded with tripods and large markers. 
Although the sizes of prisms \cite{station} (see Fig. \ref{tags}(e)) are smaller than those of calibration boards or LiDARTags, they still act as additional objects.
Moreover, in applications such as training data collection for learning-based point cloud registration methods \cite{sghr,mdgd}, these additional 3D objects are undesirable as they wreck the 3D geometry of the scene. However, considering the different modalities of 3D LiDAR point clouds and 2D images, it is challenging to develop an LFM system as convenient as VFMs (see Fig. \ref{tags}(c)). 

\subsection{LiDAR-based Mapping and Localization}
\subsubsection{LiDAR-based SLAM}
LiDAR Odometry and Mapping (LOAM) \cite{loam} is a milestone work in LiDAR-based SLAM, inspiring numerous follow-up studies \cite{floam,lloam,aloam,sdk,kiss,traj}. It splits the SLAM problem into two parallel tasks: LiDAR Odometry (LO) and mapping. LO is performed on the front end at high frequency to extract feature points (\textit{e.g.} corners, surfaces, \textit{etc}.) and use them to estimate the LiDAR pose in real-time. Mapping operates on the back end at a lower frequency to refine the odometry results and construct a more accurate map. LiDAR-based SLAM methods \cite{loam,floam,lloam,aloam,sdk,kiss,traj} focus on sequentially processing consecutive LiDAR scans and are designed for use cases that particularly require real-time response. Applications with the following three features are not use cases for LiDAR-based SLAM: (1) require processing a set of point clouds all at once, (2) involve an unordered set and/or point clouds with low overlap (dramatic viewpoint changes), and (3) allow for offline operation. In contrast, multiview point cloud registration methods are designed for these cases.

\subsubsection{Multiview Point Cloud Registration}
Multiview point cloud registration \cite{sghr,multiview,multiway,mdgd} is also a popular solution for LiDAR-based mapping and localization. It directly processes a set of point clouds, which can be unordered and have low overlap, in an offline manner. In particular, mapping is achieved by registering the multiview point clouds into a complete point cloud, while localization is done by determining the relative pose between them. The mapping and localization framework introduced in this dissertation belongs to this category.
\par
Geometry-based point cloud registration methods, such as variants of Iterative Closest Point (ICP) methods \cite{kiss,lloam} and Teaser++ \cite{teaser}, mainly focus on designing efficient and robust algorithms for describing geometry and extracting geometric features (\textit{e.g.}, points, lines, and surfaces/normals) to find correspondence between point clouds and align them. In learning-based point cloud registration methods like Predator \cite{pre} and SGHR \cite{sghr}, neural networks are designed to learn features representing the geometry \cite{features,multiview,3dmatch,se3et,geotransformer}, and then learn to utilize these features for registering the point clouds. However, most existing methods \cite{kiss,lloam,features,3dmatch} apply only to high overlap scenarios, making them less robust in practical applications \cite{pre}. Some recent learning-based research proposes multiview registration methods \cite{sghr,multiview,multiway} and studies low overlap cases \cite{pre,sghr}. Despite their promising performance on benchmarks \cite{3dmatch,eth,scan}, the generalization of learning-based methods to unseen scenarios (\textit{i.e.}, out-of-distribution cases of training data) remains problematic. Moreover, the existing methods \cite{mdgd,sghr,se3et,geotransformer,teaser} face challenges in handling degraded scenes, such as repetitive structures and textureless environments. In camera-based applications \cite{munoz,qingdao,munoz2019}, VFMs have been utilized to tackle challenging degraded scenes and low-overlap cases, while exploiting LFMs for multiview point cloud registration remains an open problem.
\subsection{Training Data Collection for Point Cloud Registration}
Training data is crucial for the development of learning-based point cloud registration methods \cite{sghr,mdgd}, but collecting it is time-consuming and expensive. For instance, in previous dataset collections \cite{3dmatch,eth,scan}, external sensors such as cameras, IMUs, and GPS are employed to obtain ground truth poses among point clouds. This requires careful multi-sensor calibration and synchronization, which can be time-consuming and labor-intensive. A common approach \cite{sghr,pre} to evaluating a learning-based point cloud registration model is to train it on 3DMatch \cite{3dmatch} and test it on various benchmarks, including 3DMatch \cite{3dmatch}, ETH \cite{eth}, and ScanNet \cite{scan}. However, the 3DMatch benchmark is mainly constructed from RGB-D camera captures of indoor scenes \cite{3dmatch}. Collecting a new training dataset with outdoor scenes and point clouds captured by a LiDAR sensor for point cloud registration is beneficial for improving the performance of learning-based methods. Unfortunately, this is challenging due to the lack of an efficient, reliable, and easy-to-use tool for the task.

\section{Contributions}
This dissertation makes several contributions to addressing the technological gap in utilizing LFMs for mapping and localization introduced in Section \ref{Challenges}. Specifically, the main contributions of this dissertation are as follows:
\begin{itemize}

\item Intensity Image-Based LiDAR Fiducial Marker (IFM) System. A new IFM system is introduced that supports various LiDAR models and integrates seamlessly with existing VFM patterns \footnote{Please note that the focus of our research is enabling LiDAR sensors to detect thin-sheet markers with patterns from VFM research, rather than proposing new patterns.}. This system offers a practical and cost-effective solution by allowing users to generate markers with standard printing methods, making the process as convenient as current VFM systems.

\item Advanced LFM Detection Method. A novel approach is developed to detect 3D fiducial markers through intensity images. This enables the integration of VFM systems into the IFM system, enhancing its flexibility compared to existing systems like LiDARTag, which are limited to specific marker patterns.

\item Improved Pose Estimation. A new pose estimation technique for LiDAR is proposed, achieving higher accuracy than traditional VFM-based camera systems. The method also eliminates rotation ambiguity and provides robustness against lighting variations.

\item Enhanced Localization of Thin-Sheet LFMs. Limitations of the vanilla IFM are addressed. A method is introduced to extend LFM localization to 3D LiDAR maps and increase the detection distance of LFMs.

\item Joint Analysis of Intensity and Geometry. A novel pipeline is proposed for detecting planar fiducial markers by jointly analyzing 3D point clouds from both intensity and geometry perspectives. This differs from conventional methods, which rely solely on geometric features and are ineffective for detecting planar objects indistinguishable from their background.

\item Framework for Mapping and Localization. A comprehensive framework for mapping and localization using LFMs is developed. This framework registers unordered multiview point clouds with low overlap, providing a reliable tool for various applications, such as 3D asset collection from sparse scans, training data collection in unseen scenes, degraded scene reconstruction, GPS-denied environment localization, and 3D map merging.

\item Livox-3DMatch Dataset. A new training dataset, Livox-3DMatch, is created, augmenting the original 3DMatch data. This expanded dataset improves the performance of state-of-the-art learning-based methods across multiple benchmarks.

\item Adaptive Threshold Detection Algorithm. A viewpoint-robust LFM detection algorithm is developed, ensuring reliable detection in varying real-world conditions.
\end{itemize}
These contributions collectively advance the use of LFMs in robust and efficient LiDAR-based mapping and localization systems, addressing critical challenges and unlocking new possibilities in robotics and computer vision.

\section{Dissertation Organization}
This dissertation is organized into five chapters, in addition to the Introduction.
%
Chapter \ref{IFM} is about the development of a general fiducial marker system for the LiDAR sensor, named IFM. Section \ref{3.1} presents an overview of the IFM system, including its overall pipeline and advantages compared to closely related prior work LiDARTag \cite{lt}. Chapter \ref{pre} mainly introduces the preliminary knowledge of three-dimensional transformation, Lie group and Lie algebra, and LiDAR technology. Section \ref{3.2} introduces a method for detecting 3D fiducials in a point cloud through the intensity image. Section \ref{3.3} presents the estimation of the 6-degree-of-freedom (6-DOF) LiDAR pose using the IFM. Section \ref{3.4} reports the qualitative and quantitative experimental results and analysis regarding the efficiency and limitations of the vanilla IFM. The developments listed in this chapter, and related materials have been published in the journal article \cite{iilfm}. The open-source implementation, based on C++ and Robot Operating System (ROS), as well as the datasets, are available at \url{https://github.com/York-SDCNLab/IILFM}.
\par
Chapter \ref{improve} discusses an algorithm that addresses the two limitations of the vanilla IFM: its inapplicability to 3D LiDAR maps and its limited detectable range. Section \ref{4.1} presents an overview of the proposed algorithm, including a comparison to the vanilla IFM in terms of the pipeline for processing point clouds and applicable scenes. Section \ref{4.2} discusses a new method that jointly analyzes point clouds from both intensity and geometry perspectives to find point clusters that cloud contain fiducial markers.
Section \ref{4.3} presents a method for localizing fiducials within the point clusters via a designed intermediate plane. Section \ref{new4.4} introduces how the 3D maps are constructed. Section \ref{4.4} reports the experimental results demonstrating the improvements brought by the proposed method over the vanilla IFM. The developments listed in this chapter, and related materials have been published in the journal article \cite{access}. The open-source implementation based on C++ is available at \url{https://github.com/York-SDCNLab/Marker-Detection-General}.
\par
Chapter \ref{multi} is about the development of a framework exploiting LFMs for mapping and localization.
Section \ref{5.1} provides an overview of the proposed framework, including its overall pipeline and the rich downstream robotic and computer vision applications. Section \ref{new5.2} introduces the methodology.
Section \ref{dataset} introduces the construction of a new training dataset for point cloud registration, named Livox-3DMatch. Section \ref{5.7} reports the experimental results that validate the proposed framework and demonstrate its rich applications in robotic and computer vision. The developments listed in this chapter and related materials are accepted in the journal article \cite{lpr}. 
The open-source implementation, based on C++ and Python, as well as the datasets, are available at \url{https://github.com/yorklyb/L-PR}.
\par
Finally, the conclusions and future work are presented in Chapter \ref{conclusion}.


\chapter[Intensity Image-based LiDAR Fiducial Marker System]{Intensity Image-based LiDAR Fiducial \\Marker System \label{IFM}}
\section{Overview} \label{3.1}
\begin{figure}[H] 
	\centering
	\includegraphics[width=0.8\linewidth]{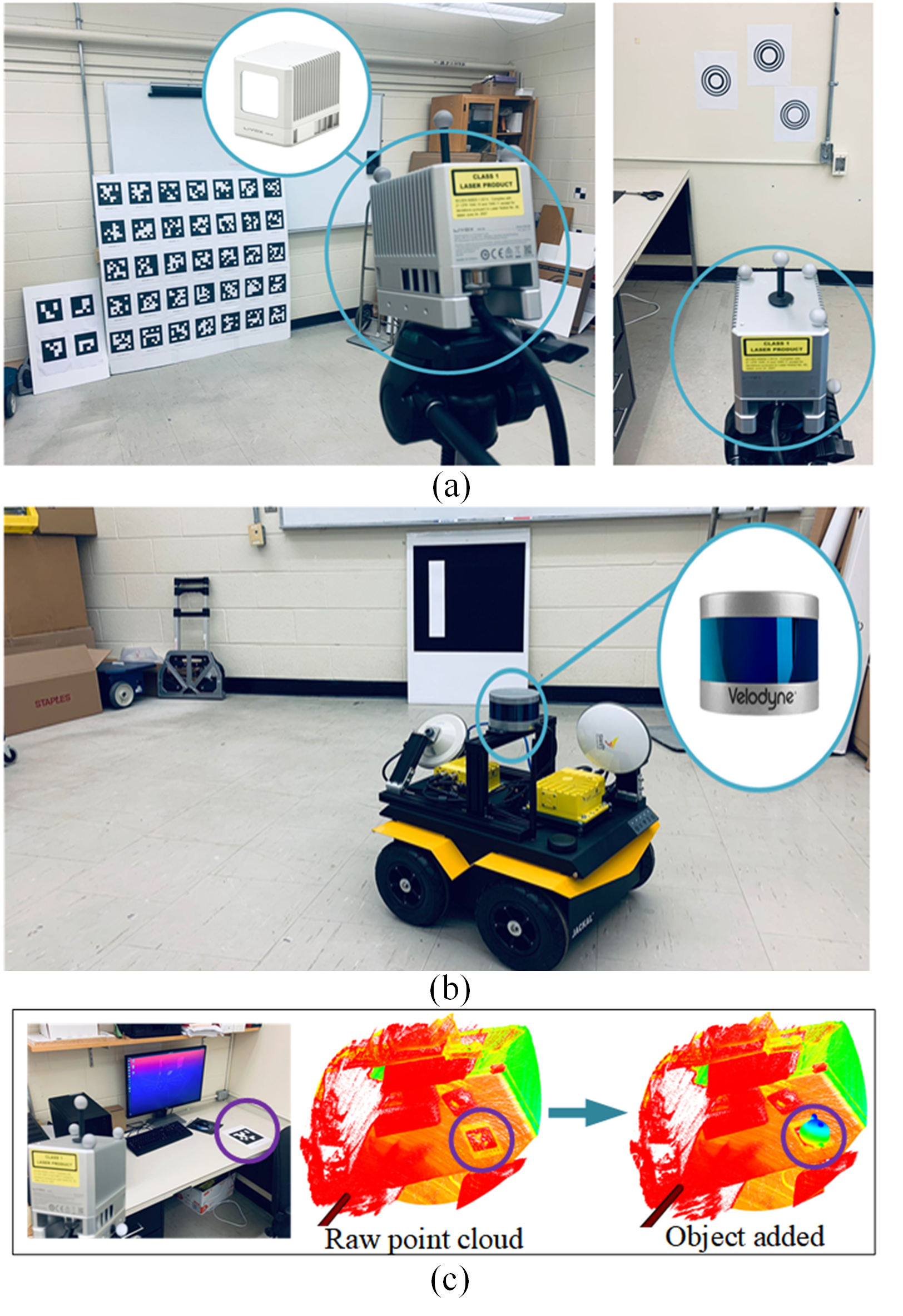}
	\caption{An illustration of the flexibility and generalizability of the proposed IFM.
 }
	\label{flex}
\end{figure}

This chapter introduces the vanilla IFM system. As illustrated in Fig. \ref{flex}, the proposed IFM is a general system with ample flexibility. As seen, unlike previous LiDAR fiducial objects, the markers are thin-sheet objects attached to other surfaces without affecting the 3D geometry of the environment. In particular, Figs. \ref{flex}(a,b,c) show that different VFM systems, such as (a) AprilTag \cite{ap3}, (b) CCTag \cite{cctag}, and (c) ArUco \cite{aruco}, can be easily embedded. Moreover, the system is applicable to both solid-state LiDAR Figs. \ref{flex}(a, b) and mechanical LiDAR Fig. \ref{flex}(c). An AR demo using the proposed system is shown in Fig. \ref{flex}(d): the teapot point cloud is transmitted to the location of the marker in the LiDAR's point cloud based on the pose provided by the IFM system. \par
\begin{figure}[H] 
	\centering
	\includegraphics[width=1.0\linewidth]{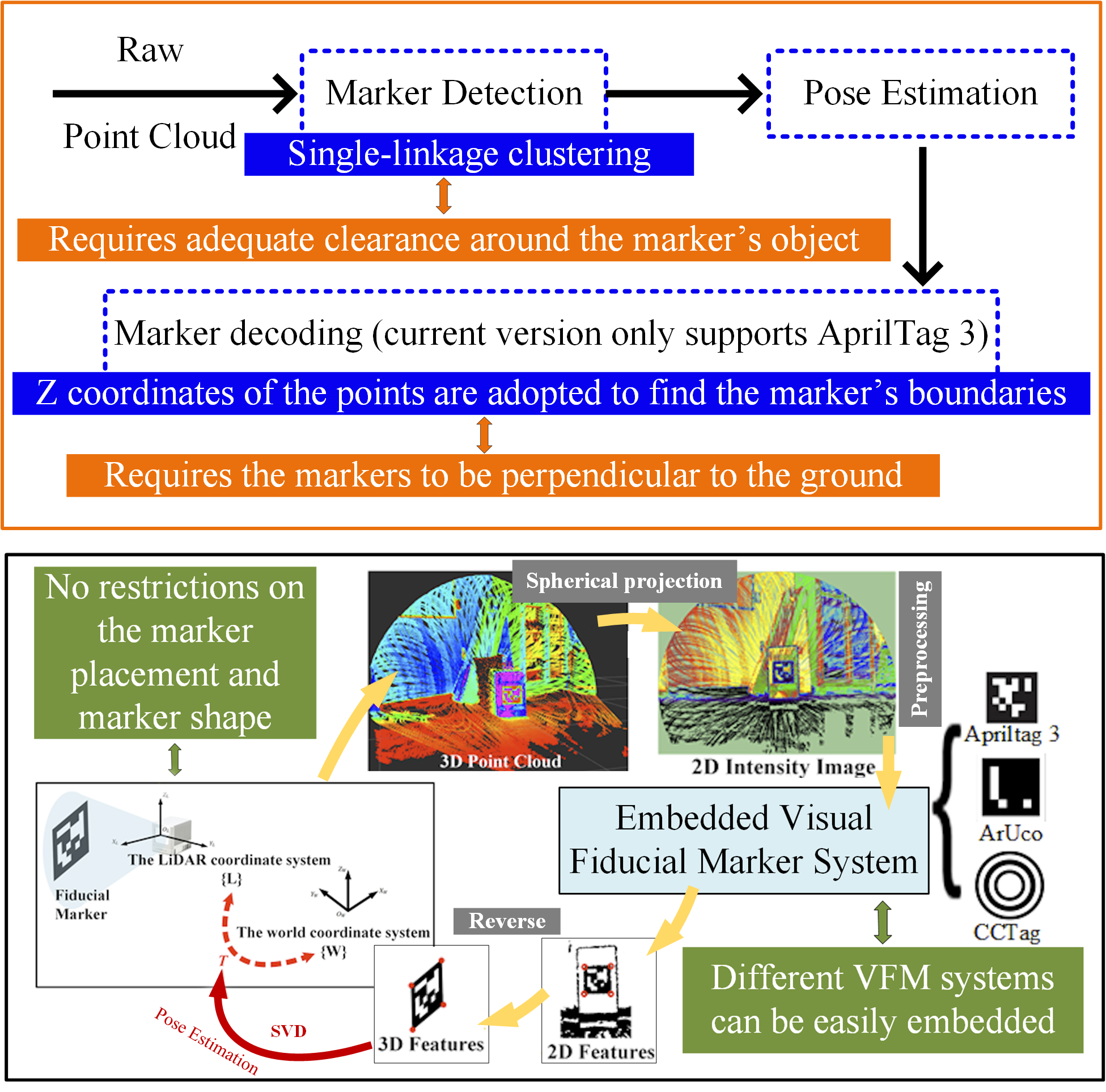}
	\caption{Comparison of LiDARTag \cite{lt} (top) and the proposed IFM (bottom).
 }
	\label{tagoverview}
\end{figure}

LiDARTag \cite{lt} is a highly relevant prior work. Fig. \ref{tagoverview} shows a comprehensive comparison between LiDARTag and the proposed system. The comparison regards three aspects. 
\begin{enumerate}[label=(\roman*)]
	 \item Convenience of usage. The LiDARTag \cite{lt} system adopts the single-linkage clustering in the tag detection stage that brings restrictions on marker placement. Namely, there must be adequate clearance around the marker's object, which makes the marker an extra object added to the spatial environment.  Furthermore, according to the implementation of LiDARTag, the method to find the boundaries of the markers in the tag decoding stage requires that the marker be perpendicular to the ground. In contrast, owing to the fact that the IFM system does not adopt the clustering method based on spatial features, it has no restrictions on marker placement. That is, the user can place the marker as they do with the conventional VFM.
	\item Extensibility. The current version of LiDARTag only supports markers with patterns belonging to the AprilTag 3 family \cite{ap3}. In the IFM system, however, different VFM systems (square and non-square) function directly since the marker detection is executed on the 2D intensity image.
	\item User-friendly design. The LiDARTag system \cite{lt}, though definitely well-programmed, adopts many settings that are different from the implementations of VFM systems. For example, users have to utilize the marker's local frame as the inertial coordinate system by default to define the sensor (LiDAR) pose. However, the development of the proposed system follows the successful VFM systems, such as AprilTag \cite{ap3} and ArUco \cite{aruco}. For instance, we leave the authority of defining the inertia coordinate system to the users as \cite{ap3,aruco} do. As a result, users who are familiar with the popular VFM systems can commence with our system comfortably.
\end{enumerate}

\section{Preliminaries} \label{pre}
\subsection{Three-dimensional Transformation} \label{threedimen}
Suppose that $\mathbf{p}_{a} = [x,y,z]^{T}$ is a 3D point expressed in the coordinate system $\{a\}$. To express the point in another coordinate system $\{b\}$ as $\mathbf{p}_{b} = [x^{\prime},y^{\prime},z^{\prime}]^{T}$, translation and rotation transformations need be applied to $\mathbf{p}_{a}$. In particular, the extrinsic matrix, $\mathbf{T}\in\mathbb{R}^{4\times4}$, is employed to describe the 6-DOF pose:

\begin{equation}
\mathbf{T} = \left[\begin{array}{cc}
\mathbf{R} & \mathbf{t} \\
\mathbf{0}^{1 \times 3} & 1 \label{transform}
\end{array}\right],
\end{equation} 
where $\mathbf{t}\in\mathbb{R}^{3\times1}$ denotes the translation vector and $\mathbf{R}\in \mathbb{R}^{3\times3}$ denotes the rotation matrix. Afterwards, $\mathbf{p}_{b}$ is obtained as follows:
\begin{equation}
\left[\begin{array}{c}
\mathbf{p}_b \\
1
\end{array}\right]=\mathbf{T}\left[\begin{array}{c}
\mathbf{p}_a \\
1
\end{array}\right]. \label{dot}
\end{equation} \par
To simplify the notation, the operator ($\cdot$) is introduced to denote:
\begin{equation}
\mathbf{p}_b=\mathbf{T} \cdot \mathbf{p}_a. 
\end{equation} \par
Please note that ($\cdot$) is not a dot product as $\mathbf{T}$ is $4 \times 4$ and $\mathbf{p}$ is $3\times 1$. So it is different from a regular dot product. Eq. (\ref{dot}) shows what it means.

\subsection{Lie Group and Lie Algebra} \label{lie}
The rotation matrix $\mathbf{R}\in SO(3)$, which is the special orthogonal group representing rotations \cite{barfoot}:
\begin{equation}
SO(3) = \{ \mathbf{R}\in\mathbb{R}^{3\times3} | \ \mathbf{R}\mathbf{R}^{T}=\mathbf{1},det(\mathbf{R}) =1 \}.
\end{equation} \par
The pose matrix $\mathbf{T} \in SE(3)$, which is the special Euclidean group representing poses \cite{barfoot}:
\begin{equation}
SE(3) = \{ \mathbf{T} = \left[\begin{array}{cc}
\mathbf{R} & \mathbf{t} \\
0 & 1 
\end{array}\right] \in \mathbb{R}^{4 \times 4} | \ \mathbf{R} \in SO(3), \mathbf{t}\in \mathbb{R}^{3 \times 1}\}.
\end{equation} \par
$SO(3)$ and $SE(3)$ are two specific Lie groups. Every matrix Lie group is associated with a Lie algebra \cite{barfoot}. The Lie algebra $\mathfrak{so(\mathrm{3})}$ that is associated with $SO(3)$ is defined as follows:
\begin{equation}
\begin{aligned}
& SO(3) \rightarrow \mathfrak{so(\mathrm{3})}: \\
& \log(\mathbf{R}) = \frac{\theta}{2\mathrm{sin}(\theta)}(\mathbf{R}-\mathbf{R}^{T}),  \\ 
& \xi = \log(\mathbf{R})_{\vee}, 
\end{aligned}
\end{equation}
where $\theta= \arccos \frac{1}{2}(Trace(\mathbf{R})-1)$. $\log(\cdot)$ is the matrix logarithm and $\xi \in \mathbb{R}^{3\times1}$ is the Lie algebra coordinates. $\vee$ is the \textit{vee} map operator that finds the unique vector $\xi \in \mathbb{R}^{3\times1}$ corresponding to a given skew-symmetric matrix $\log(\mathbf{R})\in \mathbb{R}^{3 \times 3}$ \cite{barfoot,tagslam}.

\subsection{LiDAR Technology}
\begin{figure}[H] 
	\centering
	\includegraphics[width=1.0\linewidth]{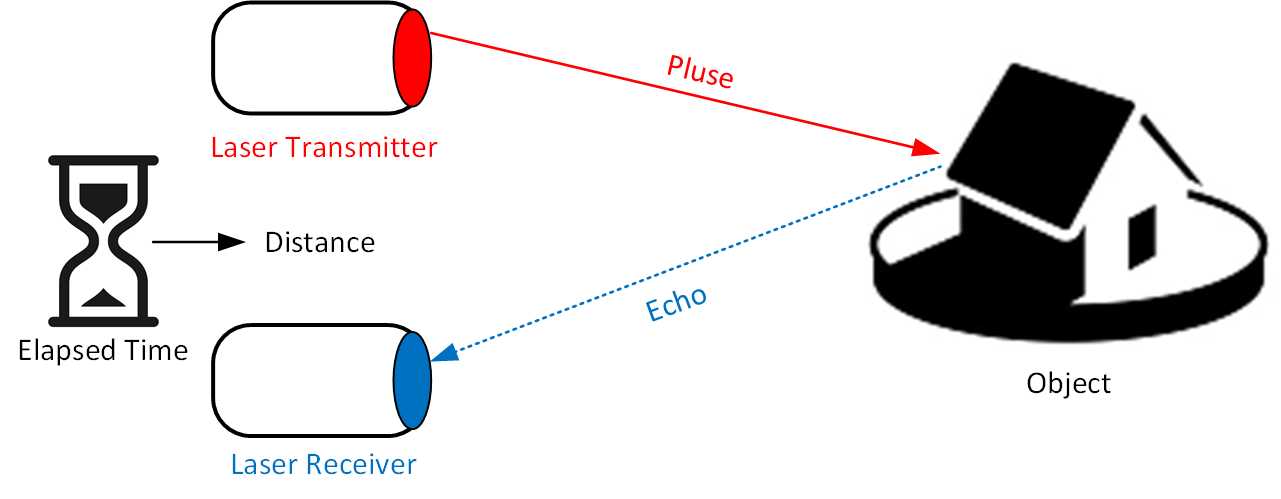}
	\caption{ The schematic diagram of the working principle of LiDAR.
 }
	\label{lidarwork}
\end{figure}

LiDAR is a technology that uses laser pulses to measure the distance between the sensor and objects in the environment. It creates precise three-dimensional representations of the scanned scene. 
The LiDAR sensor emits laser beams, and by measuring the time it takes for the reflected light to return to the sensor, the positions of 3D points in space are calculated.
In addition to the geometric data, a LiDAR sensor can also capture intensity values, which reflect the strength of the returned signal. These intensity values are influenced by factors such as surface material, texture, and angle of incidence. The inclusion of intensity information enhances the ability to distinguish between different types of surfaces, improving tasks like object detection, scene understanding, and feature recognition. Fig.~\ref{lidarwork} illustrates the schematic diagram of the working principle of LiDAR.

\section{Marker Detection} \label{3.2}
\label{marker-detection}
\subsection{Generation of the Intensity Image} \label{twotwoone}
As introduced in LiDARTag \cite{lt}, the reason LiDAR fiducial objects act as additional 3D objects, abandoning the free-placement advantage of VFMs, is due to the gap between structured images and unstructured point clouds.
Nevertheless, this gap is not insurmountable. In particular, there is a notable hot trend in LiDAR-based 3D object detection/segmentation \cite{rangenet,lasernet}: Neural Networks that are originally developed for 2D object detection/segmentation can be utilized to detect/segment the objects in the range/intensity image(s) of the 3D LiDAR point cloud. This indicates that the range/intensity image(s) generated from the LiDAR point cloud can be a pathway to transfer the research accomplishments on the 2D image to the 3D point cloud. Following this inspiration, and considering that the black-and-white marker is explicitly visible in the point cloud rendered by intensity, the intensity image is utilized for LFM development.
\par
\begin{figure}[H]
	\centering
	\includegraphics[width=0.9\linewidth]{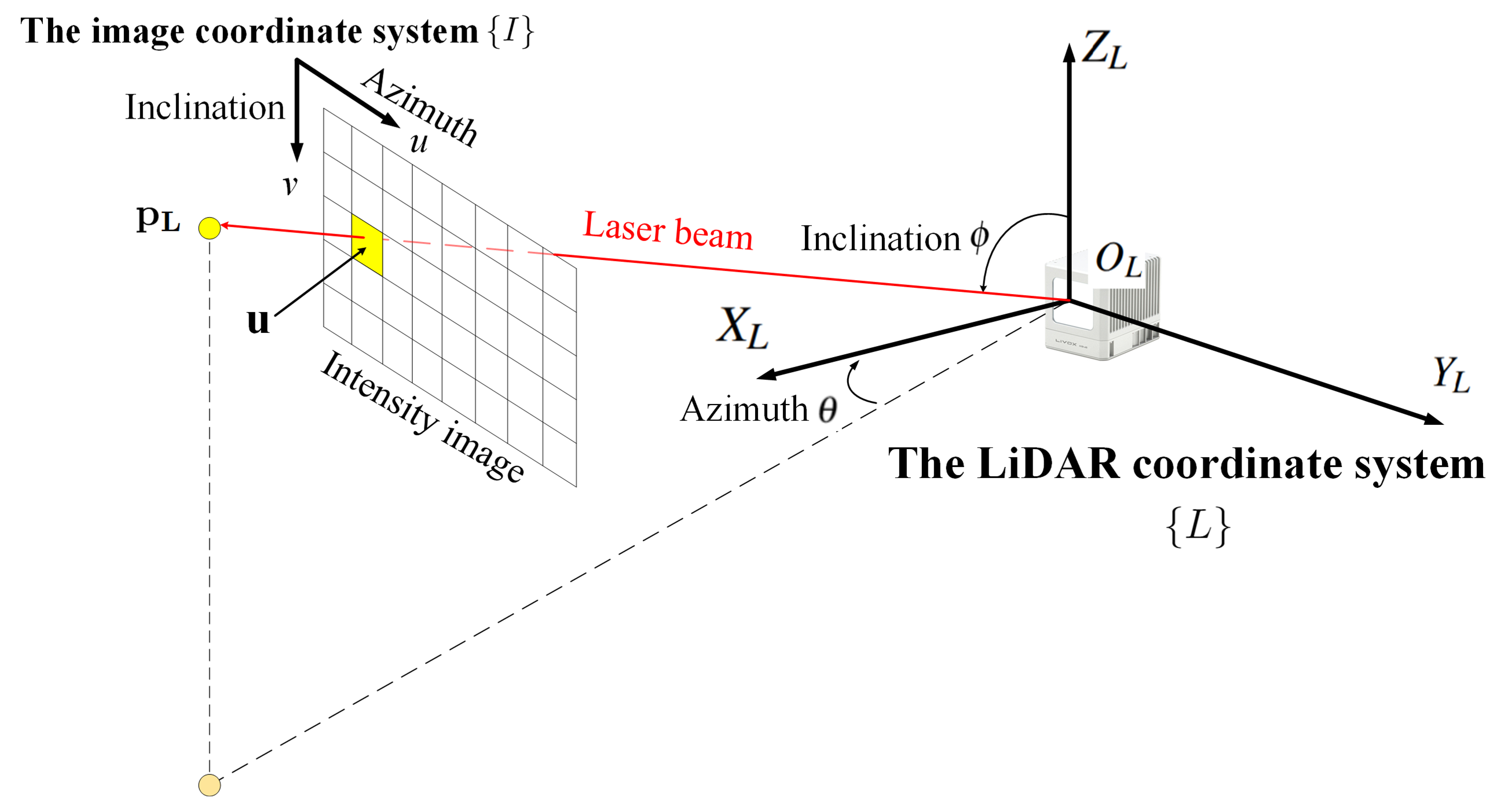}	
	\caption{An illustration of the coordinate systems and notations.}
	\label{notation}
\end{figure}

The generation of an intensity image from a given unstructured point cloud can be summarized as transferring all the 3D points in the point cloud onto a 2D image plane by spherical projection and rendering the corresponding pixels with intensity values. Fig. \ref{notation} shows the coordinate systems and notations. $\mathbf{p_{L}}=[x_{L},y_{L},z_{L},i]^{T}$ is an observed point in the 3D point cloud, with ${[x_{L},y_{L},z_{L}]^{T}}$ denoting its Cartesian coordinates w.r.t. the LiDAR coordinate system $\{L\}$ and $i$ being the intensity.  $\mathbf{u}$ is the projection of $\mathbf{p_{L}}$ onto the image plane, which has coordinates $[u,v]^{T}$ w.r.t. the image coordinate system, $\{I\}$. 
Following \cite{barfoot}, the Cartesian coordinates of $\mathbf{p_{L}}$ are first transformed to spherical coordinates $[\theta,\phi,r]^{T}$:
\begin{equation}	
	\begin{aligned}
		& \theta=\arctan(\frac{y_{L}}{x_{L}}),\\
		& \phi=\arctan(\frac{z_{L}}{\sqrt{x_{L}^{2}+y_{L}^2}}),\\
		& r=\sqrt{x_{L}^2+y_{L}^2+z_{L}^2},
	\end{aligned}\label{pro}
\end{equation}
where $\theta$ and $\phi$ denote the azimuth and inclination, respectively. $r$ is the range from $\mathbf{p_{L}}$ to the origin of $\{L\}$. Then, the image coordinates $[u,v]^{T}$ of  $\mathbf{u}$ are given by:
\begin{equation}	
	u = \lceil \frac{\theta}{\Theta_{a}}\rfloor + u_{o},\; v = \lceil \frac{\phi}{\Theta_{i}}\rfloor + v_{o} \label{uv}
\end{equation}
where $\lceil \; \rfloor$ represents rounding a value to the nearest integer. 
$\Theta_{a}$ and $\Theta_{i}$ are the angular resolutions in $u$ (azimuth) and $v$ (inclination) directions, respectively. $u_{o}$ and $v_{o}$ are the offsets: 
\begin{equation}	
	\theta =\Theta_{a}(u - u_{o}), \; \phi =\Theta_{i}(v - v_{o}).  \label{test11}
\end{equation} \par
Assume it is desired that the point with zero-azimuth and zero-inclination to be projected to the center of the image, then the offsets will be: $u_{o}=I_{w}/2$ and $v_{o}=I_{h}/2$, where $I_{w}$ and $I_{h}$ being the image width and height which are determined by the maximum angular width $P_{w}$ and height $P_{h}$ of the point cloud $I_{w} = \lceil \frac{P_{w}}{\Theta_{a}}\rfloor, \; I_{h} = \lceil \frac{P_{h}}{\Theta_{i}}\rfloor$. The pixel $[u,v]^{T}$ is then rendered by a specific color corresponding to the intensity value $i$. Refer to \cite{colormap,PCL} to see how the correspondence between color and intensity value is generated. For each pixel, the range information $r$ is also saved for the sake of the following pose estimation. Thereafter, we step through the point cloud and repeat the above process. The pixels that are not mapped to any points will remain unobserved and are rendered by a unique predefined value. Note that if these pixels are visited later on in the marker detection process, they will not return any 3D points because they represent unobserved regions.

\subsection{Selections of the Angular Resolutions} \label{select}

\begin{figure}[H] 
	\centering
	\includegraphics[width=0.7\linewidth]{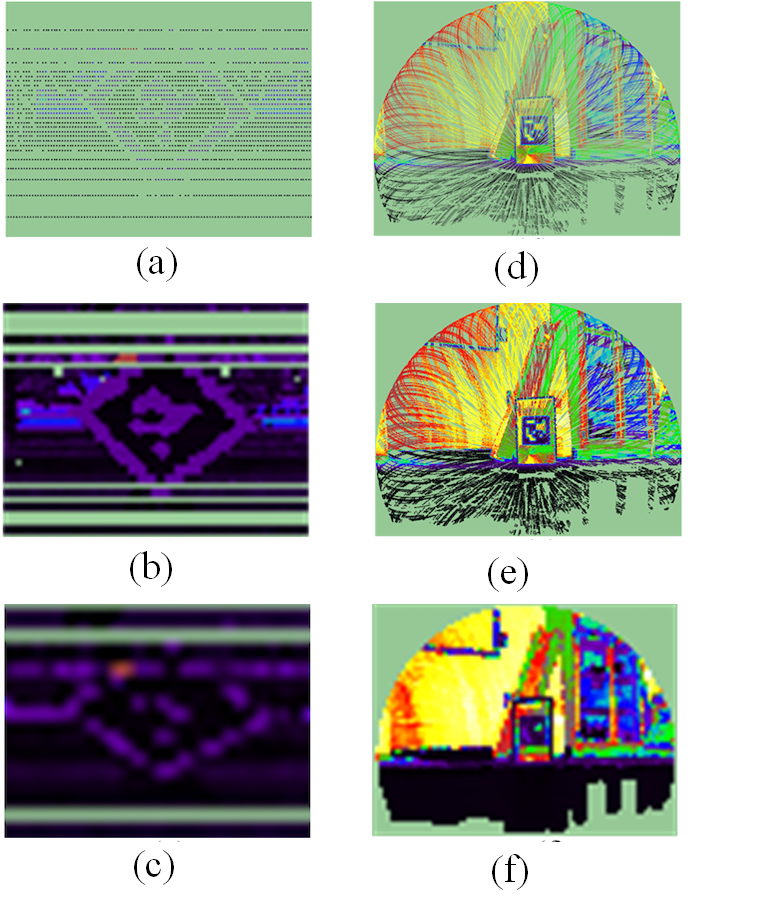}
 \caption{The intensity images generated under different angular resolution settings.
 }
	\label{newreso}
\end{figure}

The selections of $\Theta_{a}$ and $\Theta_{i}$ are crucial as they affect the quality of the intensity image directly. However, the solution to this problem is straightforward. Suppose that the horizontal angular resolution and vertical angular resolution given by the user manual of the employed LiDAR are $\Theta_{h}$ and $\Theta_{v}$. We should set $\Theta_{a}=\Theta_{h}$ and $\Theta_{i}=\Theta_{v}$. Fig. \ref{newreso} illustrates the effect of choosing different values of $\Theta_{a}$ and $\Theta_{i}$ on the intensity image, with the unobserved regions rendered in light green.
\par
Figs. \ref{newreso}(a,b,c) are intensity images generated from the same point cloud given by one LiDAR scan of a Velodyne ULTRA Puck (mechanical LiDAR, $\Theta_{h}=0.4^{\circ}$ and $\Theta_{v}=0.33^{\circ}$). The LiDAR scan is extracted from the dataset provided by \cite{lt}. Moreover, Figs. \ref{newreso}(a,b,c)
are cropped to display.
Figs. \ref{newreso}(d,e,f) are intensity images generated from the same point cloud which is the integration of multiple LiDAR scans of a Livox Mid-40 (solid-state LiDAR, $\Theta_{h}=\Theta_{v}=0.05^{\circ}$). The detailed settings for each subimage are as follows. (a): $\Theta_{a}=0.05^{\circ}$ and $\Theta_{i}=0.05^{\circ}$; raw image size = $7200\times800$. (b): $\Theta_{a}=0.4^{\circ}$ and $\Theta_{i}=0.33^{\circ}$; raw image size = $900\times121$. (c): $\Theta_{a}=0.6^{\circ}$ and $\Theta_{i}=1.0^{\circ}$; raw image size = $600\times40$. (d): $\Theta_{a}=\Theta_{i}=0.025^{\circ}$; raw image size = $1538\times1178$. (e): $\Theta_{a}=\Theta_{i}=0.05^{\circ}$; raw image size = $771\times591$. (f): $\Theta_{a}=\Theta_{i}=0.5^{\circ}$; raw image size = $81\times62$.

In summary, when  $\Theta_{a}<\Theta_{h}$ and $\Theta_{i}<\Theta_{v}$, too many unobserved regions appear around or inside the marker's area (See Fig. \ref{newreso}(a) and (d)). When  $\Theta_{a}>\Theta_{h}$ and $\Theta_{i}>\Theta_{v}$, too many points overlap for the same pixels based on Eq.~(\ref{uv}), such that the marker's pattern could disappear (See Fig. \ref{newreso}(c) and (f)). \par
\begin{figure}[H] 
	\centering
	\includegraphics[width=1.0\linewidth]{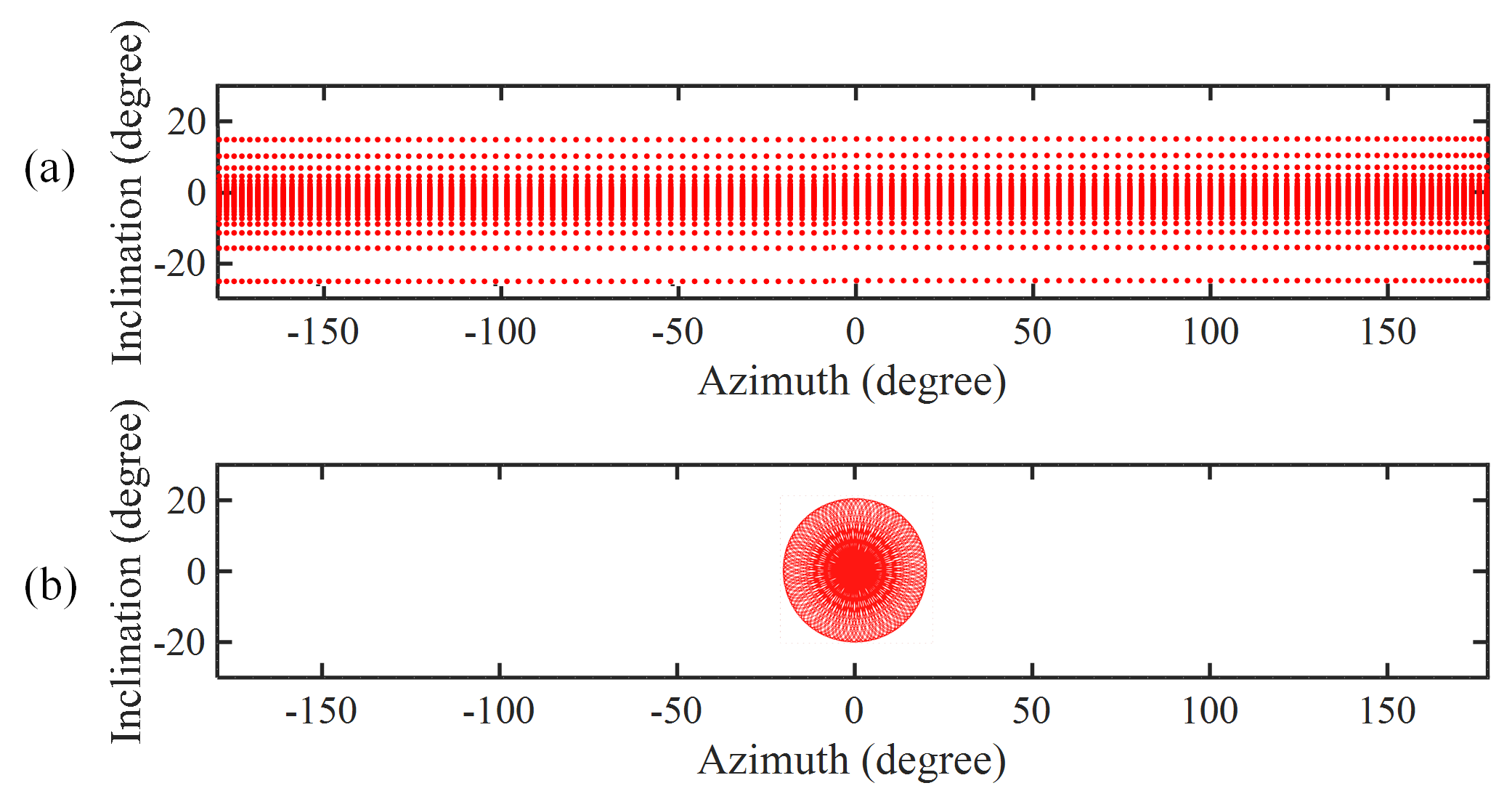}
	\caption{Sampling patterns of the mechanical LiDAR and solid-state LiDAR, with sampling points represented by red scatter plots.
 }
	\label{samplecom}
\end{figure}

Nevertheless, as shown in Fig. \ref{samplecom}, the LiDAR cannot sample perfectly evenly in the inclination/azimuth space. Specifically, Fig. \ref{samplecom}(a) shows the general sampling pattern of a mechanical LiDAR expressed in the azimuth/inclination coordinate system. This is a general schematic that does not correspond to any LiDAR model. Fig. \ref{samplecom}(b) presents the sampling pattern of the Livox Mid-40 LiDAR after one-second integration. Note that the sampling patterns vary when it comes to different solid-state LiDAR models. But generally, the unscanned regions within the valid Field of View (FoV) appear as spots.
Thus, even if the user manual of the employed LiDAR is followed to set $\Theta_{a}=\Theta_{h}$ and $\Theta_{i}=\Theta_{v}$, the unwanted unobserved regions around or inside the marker's area and the points overlapping issue are still inevitable on account the fact that $\Theta_{h}$ and $\Theta_{v}$ are approximate values themselves. Based on the experiments, the points overlapping issue under $\Theta_{a}=\Theta_{h}$ and $\Theta_{i}=\Theta_{v}$ could have an acceptable influence on pose estimation while the unobserved regions impede marker detection.
 \par

\begin{figure}[H] 
	\centering
	\includegraphics[width=0.8\linewidth]{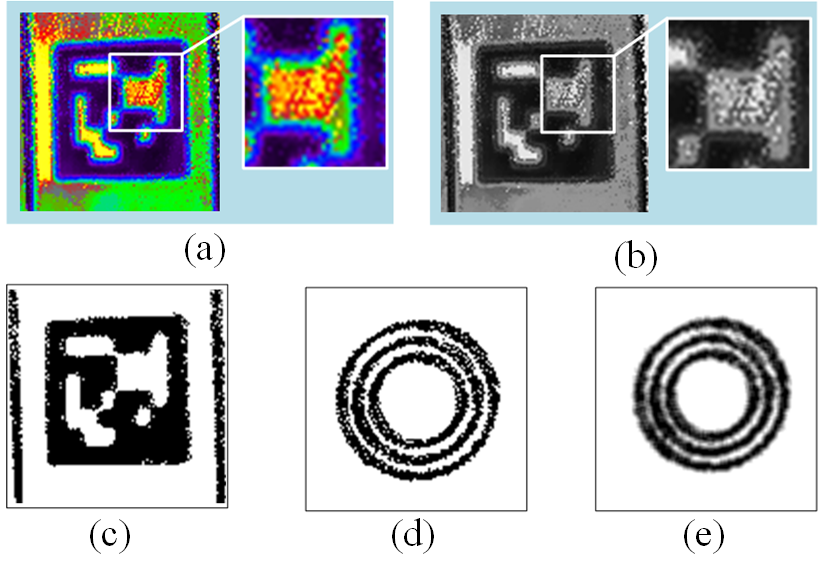}
	\caption{An illustration of image preprocessing in the system.
 }
	\label{dis}
\end{figure}

To remove the image noise caused by unobserved regions, image preprocessing is carried out as shown in Fig. \ref{dis} before inputting the raw intensity image (Fig.~\ref{dis}(a)) into the embedded VFM system. The preprocessing includes grayscale conversion (Fig.~\ref{dis}(b)), naive thresholding (Fig.~\ref{dis}(c)), and optional Gaussian blurring \cite{gb} (Fig.~\ref{dis}(d)). 
In particular, the raw intensity image (Fig.~\ref{dis}(a)) is first converted to a grayscale image (Fig.~\ref{dis}(b)). Then it becomes the binary image (Fig.~\ref{dis}(c)) using naive thresholding. After the naive thresholding, the image noise caused by varying low-intensity values is removed. Gaussian blurring \cite{gb} is recommended when the embedded VFM system, such as CCTag \cite{cctag}, does not contain it. Fig.~\ref{dis}(d) shows the intensity image of a CCTag after the naive thresholding while the detector cannot detect the marker in it. Fig.~\ref{dis}(e) presents the intensity image after applying Gaussian Blur \cite{gb} on Fig.~\ref{dis}(d). Now the marker is detectable.
The methods utilized in the preprocessing are simple but not trivial, which implies that without them the marker detection will fail.
 \par

\subsection{3D Fiducials Estimation} \label{twotwothree}
The preprocessed image introduced in the previous section is then inputted into the embedded VFM system. Thereafter, the VFM system provides the detection information. In this section, we introduce how to project the 2D fiducials given by the detection information back to the 3D space, such that they become 3D fiducials expressed in the LiDAR coordinate system. As for the square markers, the fiducials refer to the four vertices of the quad. While some of the VFM systems adopt the non-square design, the proposed method has no restriction on the marker shape.
\par
As mentioned in Section~\ref{twotwoone}, the range information is stored for each pixel in the intensity image as well. Thus, a 2D pixel with range information ($\mathbf{{u}^{r}}=[u,v,r]^{T}$) can be projected back to the 3D Cartesian coordinate system by solving the inverse of Eqs. (\ref{pro}-\ref{uv}) to find the corresponding $[x_{L},y_{L},z_{L}]^{T}$. However, in the real world, it cannot be guaranteed that the fiducials of the marker are exactly scanned by the LiDAR. Namely, the detected 2D features in Fig.~\ref{dis}(c) could correspond to the unobserved regions in the raw intensity image. As a matter of fact, this occurs frequently in our experiments, as well as in the dataset provided by \cite{lt}. Hence, when these detected but unscanned features are checked, there will be no range value $r$ returned and it is not feasible to solve the inverse of Eqs. (\ref{pro}-\ref{uv}).
\par
\begin{figure}[H] 
	\centering
	\includegraphics[width=1.0\linewidth]{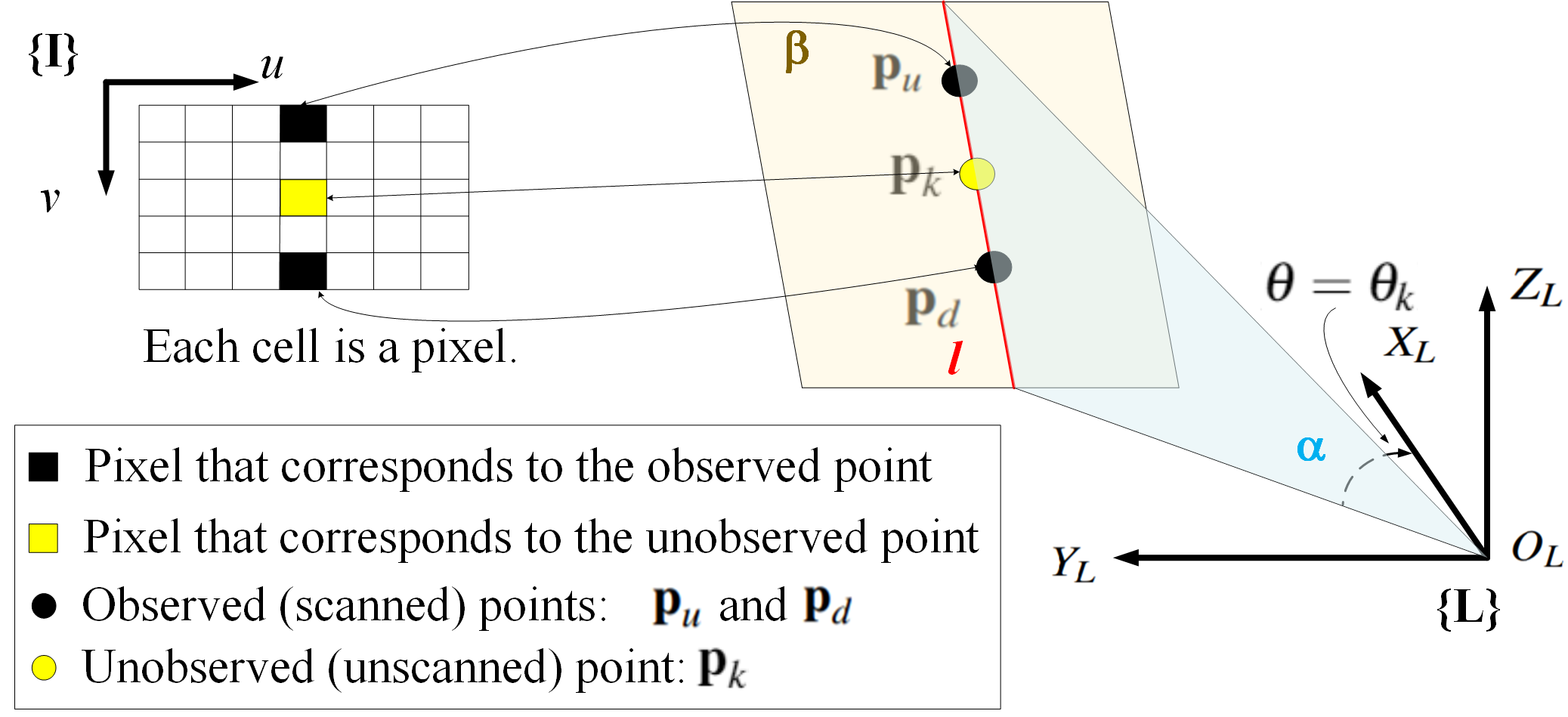}
	\caption{An illustration of the algorithm to estimate the 3D coordinates of a detected but unscanned 2D feature point.} \label{collinear}
\end{figure}
\begin{figure}[H] 
	\centering
	\includegraphics[width=0.8\linewidth]{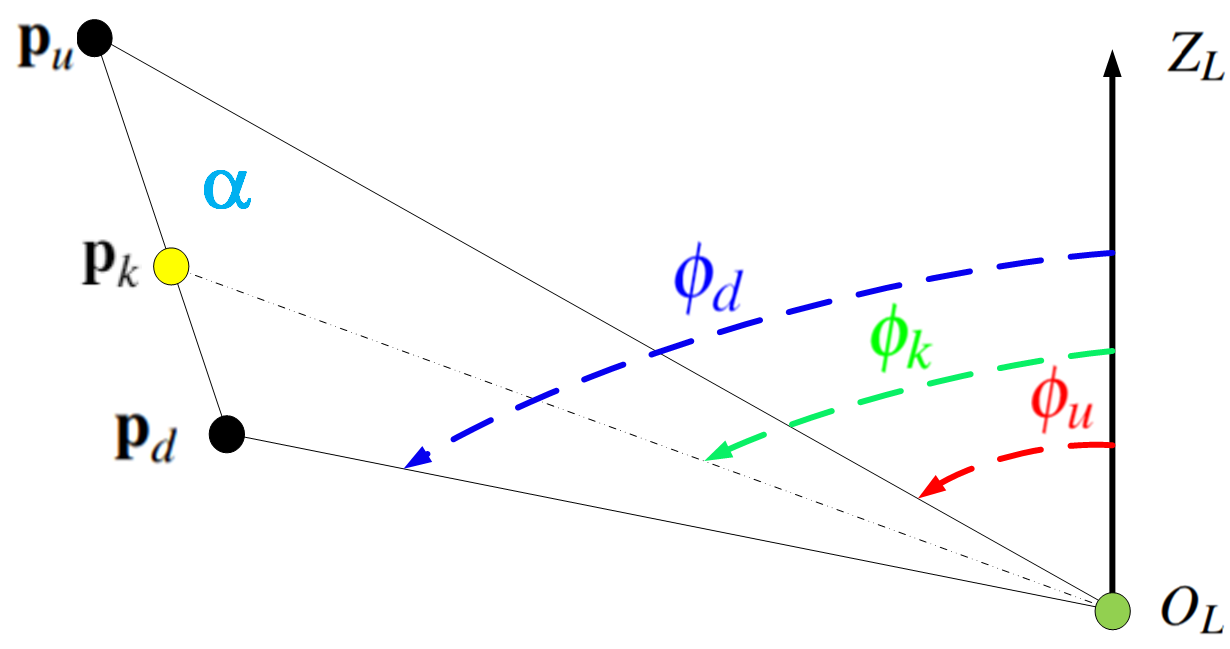}
	\caption{The side view of $\alpha$. $\phi_{d}$, $\phi_{k}$, and $\phi_{u}$ are the inclinations of $\mathbf{{p}_{d}}$, $\mathbf{{p}_{k}}$, and $\mathbf{{p}_{u}}$, respectively. }
	\label{side}
\end{figure}

To resolve this problem, we propose an algorithm as illustrated in Fig.~\ref{collinear}. The algorithm is based on the fact that the markers are planar. Suppose that there is a 2D feature point $\mathbf{{u}_{k}}=[u_{k},v_{k}]^{T}$ that is detected but corresponds to an unobserved region in the raw intensity image. The azimuth $\theta_{k}$ of $\mathbf{{u}_{k}}$ is determined by Eq. (\ref{uv}) through $\theta_{k}=\Theta_{a}(u_{k}-u_{0})$. Define the unknown 3D point corresponding to $\mathbf{{u}_{k}}$ as $\mathbf{{p}_{k}}=[x_{k},y_{k},z_{k}]^{T}$ (the yellow point in Fig.~\ref{collinear}).  Hereafter, suppose that in the same column as $\mathbf{{u}_{k}}$, there is a pair of observed pixels that are symmetric about $\mathbf{{u}_{k}}$, and define their corresponding points as $\mathbf{{p}_{u}}=[x_{u},y_{u},z_{u}]^{T}$ and $\mathbf{{p}_{d}}=[x_{d},y_{d},z_{d}]^{T}$ (the black point in Fig.~\ref{collinear}). As illustrated in Eq. (\ref{uv}), pixels in the same column approximately share the same azimuth. 
Thus, $\mathbf{{p}_{u}}$, $\mathbf{{p}_{k}}$, and $\mathbf{{p}_{d}}$ are on the same plane, $\alpha$, which is specified by fixing the azimuth ($\theta=\theta_{k}$). 
After that, define the plane where the marker is located as $\beta$. Obviously, the intersection of $\alpha$ and $\beta$ is a straight line $\textbf{\textit{l}}$. Under the assumption that $\mathbf{{p}_{u}}$ and $\mathbf{{p}_{d}}$  are on the marker plane $\beta$, it can be shown that  $\mathbf{{p}_{u}}$, $\mathbf{{p}_{k}}$, and $\mathbf{{p}_{d}}$ are collinear and they all fall on $\textbf{\textit{l}}$. To introduce the algorithm more clearly, Fig.~\ref{side} shows the side view of plane $\alpha$.

 \par

Undoubtedly, $\mathbf{{O}_{L}}\mathbf{{p}_{k}}$ is the angle bisector of $\angle\mathbf{{p}_{u}}\mathbf{{O}_{L}}\mathbf{{p}_{d}}$ owing to $\phi_{d}-\phi_{k}=\phi_{k}-\phi_{u}$. Hence, in the light of the angle bisector properties, we have 
	$\mathbf{{p}_{k}}\mathbf{{p}_{d}}/\mathbf{{p}_{u}}\mathbf{{p}_{k}}=\mathbf{{O}_{L}}\mathbf{{p}_{d}}/\mathbf{{O}_{L}}\mathbf{{p}_{u}}$. Note that $\mathbf{{O}_{L}}\mathbf{{p}_{d}}$ and $\mathbf{{O}_{L}}\mathbf{{p}_{u}}$ are the ranges of $\mathbf{{p}_{d}}$ and $\mathbf{{p}_{u}}$ which can be obtained if they are scanned. Thus, the unknown 3D coordinates of $\mathbf{{p}_{k}}$ are estimated by $\mathbf{{p}_{k}}=\mathbf{{M}_{1}}\mathbf{{p}_{d}}+\mathbf{{M}_{2}}\mathbf{{p}_{u}}$, where $\mathbf{{M}_{1}}=\mathrm{diag}(\frac{\mu}{1+\mu} ,\frac{\mu}{1+\mu} ,\frac{\mu}{1+\mu} ) $ and $\mathbf{{M}_{2}}=\mathrm{diag} ( \frac{1}{1+\mu}, \frac{1}{1+\mu}, \frac{1}{1+\mu})$ with $\mu$ being the ratio $\mathbf{{O}_{L}}\mathbf{{p}_{d}}/\mathbf{{O}_{L}}\mathbf{{p}_{u}}$. So far, the 2D fiducials in $\{I\}$, observed and unobserved by the LiDAR, are projected back to ${\{L\}}$.

\section{LiDAR Pose Estimation} \label{3.3}
The aim of LiDAR pose estimation is to seek the Euclidean transformation, $\mathbf{T}=[\mathbf{R}| \mathbf{t}]$, from the world (inertia) coordinate system $\{G\}$ to the LiDAR coordinate system $\{L\}$.  $\mathbf{R}$ is a $3\times3$ orthogonal matrix that represents the rotation. $\mathbf{t}\in\mathbb{R}^{3}$ is the translation vector. Suppose that $\mathbf{f} \in \mathbb{R}^{3}$ are the 3D coordinates of a feature point, the operation of $\mathbf{T}$ on $\mathbf{f} \in \mathbb{R}^{3}$ is $\mathbf{T} \cdot \mathbf{f} = \mathbf{R}\mathbf{f}+\mathbf{t}$ (Refer to Section \ref{threedimen} if needed). LiDAR pose estimation can be resolved through optimally aligning two point sets while in real-world applications, such as SLAM and perception, the point correspondences between the two point sets are unknown, such that the correspondences are also needed to be optimally and iteratively searched \cite{nicp}. However, as seen in the following, with the help of the fiducial marker system, the search for correspondences can be skipped in LiDAR pose estimation, which is a vital benefit brought by using the fiducial marker system.

Thus far two sets of 3D points are obtained. 1) $n$ fiducials w.r.t. $\{L\}$, denoted by $\mathcal{{P}_{L}} = \{ \mathbf{{f}_{1}}, \ \cdots, \ \mathbf{{f}_{n}}\}$, which are given by the 3D fiducials estimation introduced in Section~\ref{twotwothree}; 2) $n$ fiducials w.r.t. $\{G\}$, denoted by $\mathcal{{P}_{W}}= \{ \mathbf{{f}_{1}}^{\prime}, \ \cdots, \ \mathbf{{f}_{n}}^{\prime} \}$, which are predefined. Furthermore, the points in $\mathcal{{P}_{L}} $ and $\mathcal{{P}_{W}}$ are matched based on the ID number and vertex index given by the marker detection. Hence, the LiDAR pose estimation can be transformed into finding $[\mathbf{R}|\mathbf{t}]$ that optimally align $\mathcal{{P}_{L}} $ and $\mathcal{{P}_{W}}$. This is inherently a least square problem:
\begin{equation}	
	\mathbf{R}^{*},\  \mathbf{t}^{*}=\underset{\mathbf{R},\  \mathbf{t}}{\arg \min } \sum_{j=1}^{n}\left\|\mathbf{f}_{j}-\left(\mathbf{R} \mathbf{f}_{j}^{\prime}+\mathbf{t}\right)\right\|^{2}. \label{least}
\end{equation}
	
Considering that the point-correspondence between $\mathcal{{P}_{L}} $ and $\mathcal{{P}_{W}}$ is quite reliable thanks to the embedded VFM system, $\mathbf{R}^{*}$ and $\mathbf{t}^{*}$ can be calculated in closed form by the Singular Value Decomposition (SVD) method \cite{barfoot,icp}. The centroids of $\mathcal{{P}_{L}} = \{ \mathbf{{f}_{1}}, \ \cdots, \ \mathbf{{f}_{n}}\}$ and $\mathcal{{P}_{W}}= \{ \mathbf{{f}_{1}}^{\prime}, \ \cdots, \ \mathbf{{f}_{n}}^{\prime} \}$ are defined as follows:
\begin{equation}
\begin{aligned}
&\mathbf{f}=\frac{1}{\mathrm{n}} \sum_{\mathrm{j}=1}^{\mathrm{n}}\left(\mathbf{f}_{j}\right), \\
&\mathbf{f}^{\prime}=\frac{1}{\mathrm{n}} \sum_{\mathrm{j}=1}^{\mathrm{n}}\left(\mathbf{f}_{j}^{\prime}\right). \label{centroid}
\end{aligned}
\end{equation}
\par
Then, Eq. (\ref{least}) can be reorganized as:
\begin{equation}
\begin{aligned}
	 &\sum_{j=1}^{n}\left\|\mathbf{f}_{j}-\left(\mathbf{R} \mathbf{f}_{j}^{\prime}+\mathbf{t}\right)\right\|^{2} \\ 
  = & \sum_{j=1}^{n}\left\| \mathbf{f}_{j} -\mathbf{R} \mathbf{f}_{j}^{\prime}-\mathbf{t} - \mathbf{f} +\mathbf{R}\mathbf{f}^{\prime} + \mathbf{f} - \mathbf{R}\mathbf{f}^{\prime} \right\|^{2} \\
  = & \sum_{j=1}^{n}\left\| \left(\mathbf{f}_{j} - \mathbf{f} -\mathbf{R}\left(\mathbf{f}_{j}^{\prime}-\mathbf{f}^{\prime}\right) \right) + \left(\mathbf{f}- \mathbf{R}\mathbf{f}^{\prime}-\mathbf{t}\right)
\right\|^{2} \\
  = & \sum_{j=1}^{n}\left(
  \left\| \mathbf{f}_{j} - \mathbf{f} -\mathbf{R}\left(\mathbf{f}_{j}^{\prime}-\mathbf{f}^{\prime}\right) \right\|^{2} +  \left\|\mathbf{f}- \mathbf{R}\mathbf{f}^{\prime}-\mathbf{t} \right\|^{2} + 2 \left(\mathbf{f}_{j} - \mathbf{f} -\mathbf{R}\left(\mathbf{f}_{j}^{\prime}-\mathbf{f}^{\prime}\right) \right)\left(\mathbf{f}- \mathbf{R}\mathbf{f}^{\prime}-\mathbf{t}\right) 
 \right). \label{eq3.6}
  \end{aligned}
\end{equation}
\par
Following the definitions of $\mathbf{f}$ and $\mathbf{f}^{\prime}$ shown in Eq. (\ref{centroid}), we have:
\begin{equation}	
\sum_{j=1}^{n}\left(\mathbf{f}_{j} - \mathbf{f} -\mathbf{R}\left(\mathbf{f}_{j}^{\prime}-\mathbf{f}^{\prime}\right) \right)\left(\mathbf{f}- \mathbf{R}\mathbf{f}^{\prime}-\mathbf{t}\right) =0.
\end{equation} \par
Namely, the last term of Eq. (\ref{eq3.6}) is zero. Thus, based on the derivation given in Eq. (\ref{eq3.6}), the problem introduced in Eq. (\ref{least}) is transformed into the following form:
\begin{equation}	
	\mathbf{R}^{*},\  \mathbf{t}^{*}=\underset{\mathbf{R},\  \mathbf{t}}{\arg \min } \sum_{j=1}^{n}\left(
  \left\| \mathbf{f}_{j} - \mathbf{f} -\mathbf{R}\left(\mathbf{f}_{j}^{\prime}-\mathbf{f}^{\prime}\right) \right\|^{2} +  \left\|\mathbf{f}- \mathbf{R}\mathbf{f}^{\prime}-\mathbf{t} \right\|^{2}\right) \label{newproblem}
\end{equation}\par
As seen in Eq. (\ref{newproblem}), the rotation ($\mathbf{R}$) and translation ($\mathbf{t}$) have been decoupled. In particular, the first term only involves $\mathbf{R}$, so $\mathbf{R}$ can be computed first and then substituted into the second term to obtain $\mathbf{t}$. Define the centroid-aligned coordinates as:
\begin{equation}
\begin{aligned}
&\mathbf{q}_{j} = \mathbf{f}_{j} - \mathbf{f} \\
&\mathbf{q}^{\prime}_{j} = \mathbf{f}_{j}^{\prime}-\mathbf{f}^{\prime}
\end{aligned}
\end{equation}
\par
By substituting the centroid-aligned coordinates into the first term of Eq. (\ref{newproblem}), the computation of rotation is described as follows:
\begin{equation}	
	\mathbf{R}^{*}=\underset{\mathbf{R}}{\arg \min } \sum_{j=1}^{n}\left\|\mathbf{q}_{j}-\mathbf{R} \mathbf{q}_{j}^{\prime}\right\|^{2}. \label{problemr}
\end{equation} \par
Expanding Eq. (\ref{problemr}) yields:
\begin{equation}
\begin{aligned}
    &\sum_{j=1}^{n}\left\|\mathbf{q}_{j}-\mathbf{R} \mathbf{q}_{j}^{\prime}\right\|^{2} \\ 
=&\sum_{j=1}^{n}\left(\mathbf{q}_{j}^{T}\mathbf{q}_{j}+\mathbf{q}_{j}^{\prime}\mathbf{R}^{T}\mathbf{R}\mathbf{q}_{j}-2\mathbf{q}_{j}^{T}\mathbf{R}\mathbf{q}_{j}^{\prime}\right). \label{eq3.11}
 \end{aligned}
\end{equation} \par
Note that, since $\mathbf{R}\in SO(3)$, $\mathbf{R}^{T}\mathbf{R}= \mathbf{I}$. Hence, only the third term in Eq. (\ref{eq3.11}) involves 
$\mathbf{R}$. As a result, the problem described in Eq. (\ref{eq3.11}) is transformed into:
\begin{equation}
\begin{aligned}
    &\sum_{j=1}^{n}-\mathbf{q}_{j}^{T}\mathbf{R}\mathbf{q}_{j}^{\prime}\\
    =&\sum_{j=1}^{n}-Trace\left(\mathbf{R}\mathbf{q}_{j}^{\prime}\mathbf{q}_{j}^{T}\right)\\
    =&-Trace\left(\mathbf{R}\sum_{j=1}^{n}\mathbf{q}_{j}^{\prime}\mathbf{q}_{j}^{T} \right).
 \end{aligned}
\end{equation} \par
Let
\begin{equation}
\mathbf{H}=\sum_{j=1}^{n}\mathbf{q}_{j}^{\prime}\mathbf{q}_{j}^{T}.
\end{equation} \par
The computation of $\mathbf{R}$ is transformed from Eq. (\ref{problemr}) to:
\begin{equation}
\underset{\mathbf{R}}{\min } \ \ -Trace\left(\mathbf{R}\mathbf{H} \right) \Leftrightarrow \underset{\mathbf{R}}{\max } \ \ Trace\left(\mathbf{R}\mathbf{H} \right) \label{eq3.14}
\end{equation}\par
\noindent\textbf{\textit{Lemma}} For any positive definite matrix $\mathbf{A}\mathbf{A}^{T}$, and any orthonormal matrix $\mathbf{B}$
\begin{equation}
Trace\left(\mathbf{A}\mathbf{A}^{T} \right) \geq Trace\left(\mathbf{B}\mathbf{A}\mathbf{A}^{T} \right).
\end{equation}\par
The detailed proof of this Lemma is provided in \cite{icp}. Find the SVD of $\mathbf{H}$:
\begin{equation}
\mathbf{H} = \mathbf{U}\mathbf{\Lambda}\mathbf{V}^{T},
\end{equation}
where $\mathbf{U}$ and $\mathbf{V}$ are $3\times3$ orthonormal matrices, and $\mathbf{\Lambda}$ is a $3\times3$
diagonal matrix with non-negative elements.  Define an orthonormal matrix as follows:
\begin{equation}
\mathbf{X} = \mathbf{V}\mathbf{U}^{T}.
\end{equation} \par
Then, the following equation is obtained:
\begin{equation}
\begin{aligned}
\mathbf{X}\mathbf{H} =& \mathbf{V}\mathbf{U}^{T}\mathbf{U}\mathbf{\Lambda}\mathbf{V}^{T} \\
=& \mathbf{V}\mathbf{\Lambda}\mathbf{V}^{T},
\end{aligned}
\end{equation}
which is symmetrical and positive definite. Consequently, based on the
Lemma, for any $3\times3$ orthonormal matrix $\mathbf{B}$,
\begin{equation}
Trace\left(\mathbf{X}\mathbf{H} \right) \geq Trace\left(\mathbf{B}\mathbf{X}\mathbf{H} \right).
\end{equation}\par
Namely, among all $3\times3$ orthonormal matrices, $\mathbf{X}$ maximizes $Trace\left( \mathbf{R}\mathbf{H}\right)$. Therefore,
\begin{equation}
\mathbf{R}^{*} = \mathbf{X} = \mathbf{V}\mathbf{U}^{T}
\end{equation}
is the solution to the problem described in Eq. (\ref{eq3.14}). Finally, $\mathbf{t}^{*}$ can be computed by substituting $\mathbf{R}^{*}$ into the second term of Eq. (\ref{newproblem}). In this research, the points in $\mathcal{{P}_{W}}$ are coplanar but not collinear. According to \cite{icp}, the solution of Eq. (\ref{least}) is unique. This is the reason why the proposed system is free from the multi-solution problem, which is unlike the VFM systems troubled by the rotation ambiguity problem \cite{munoz2019,ippe,yibo}. This is a superiority of the proposed system over the planar VFM systems \cite{ap3,aruco}.
\par
Note that the LiDAR pose is specified by the definitions of $\{G\}$ and $\{L\}$. It is a common method to define $\{G\}$ by predefining the vertices \cite{ap3}. However, the predefined vertices are optional in both the AprilTag system \cite{ap3} and the proposed system. Without the predefined vertices, the definition of  $\{G\}$ is missing. Consequently, the AprilTag system \cite{ap3} will only output the image coordinates of the vertices in the image plane and the proposed system only outputs the 3D features w.r.t. $\{L\}$. In the implementation, the predefinition of vertices is customizable since we want to leave the authority of defining $\{G\}$ to the users as the AprilTag system \cite{ap3} does.

\section{Experimental Validation} \label{3.4}

\subsection{Experimental Setup}
\begin{figure}[H] 
	\centering
	\includegraphics[width=1.0\linewidth]{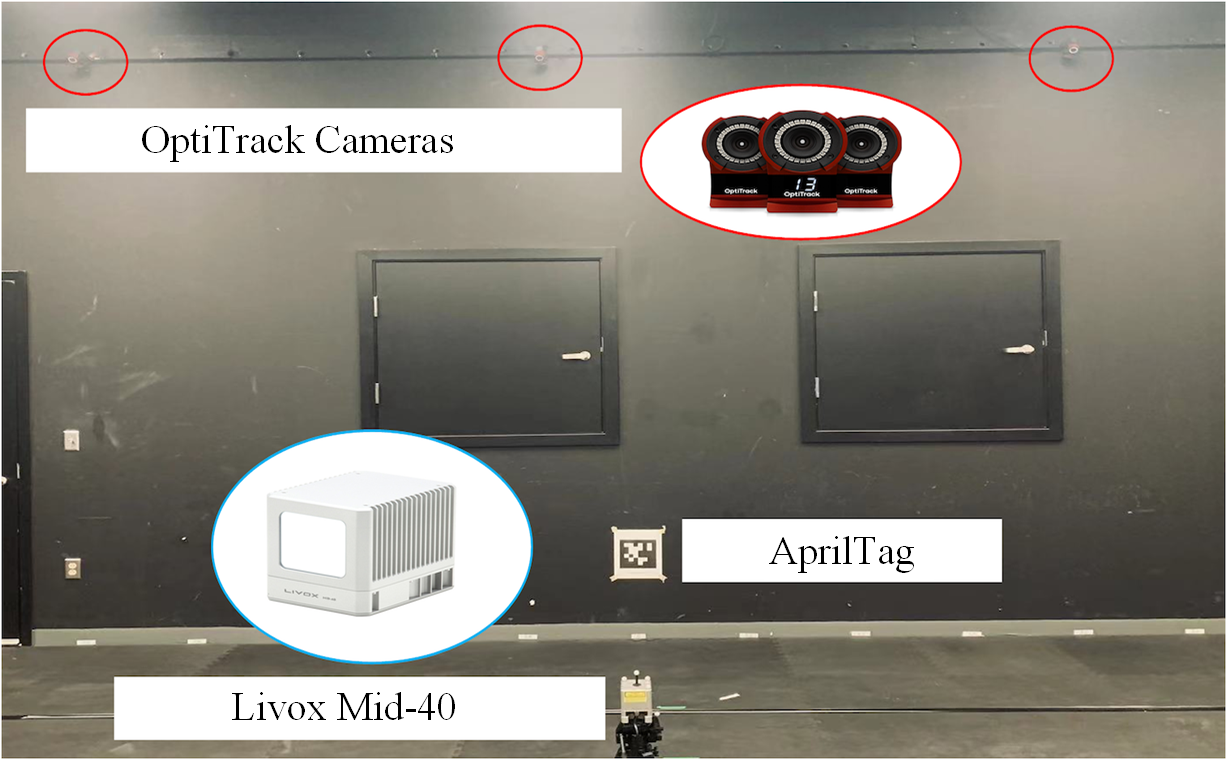}
	\caption{An illustration of the experimental setup.}
	\label{setup}
\end{figure}

To qualitatively evaluate IFM, two LiDAR models—Livox Mid-40 (solid-state LiDAR) and VLP-16 (mechanical LiDAR)—are employed, and patterns from three popular VFM systems, AprilTag \cite{ap3}, CCTag \cite{cctag}, and ArUco \cite{aruco}, are tested. To further verify the pose estimation accuracy of the IFM system, we compare the pose estimation result given by our system with the ground truth provided by the OptiTrack Motion Capture (MoCap) system (See Fig.~\ref{setup}). The MoCap system is composed of 16 OptiTrack cameras and provides the 6-DOF pose information of the predefined rigid body at 100 Hz. The low-cost solid-state LiDAR, Livox Mid-40, is employed to scan a marker pasted on the wall. The marker (ID = 0), with the size of 17.2 cm$\times$17.2 cm, belongs to the \textit{tag36h11} family of AprilTag 3 \cite{ap3} and is printed on letter-sized paper. The location of the marker's center is at (3.620, 0.00, 0.485) m w.r.t. $\{G\}$. Moreover, the rosbag provided by LiDARTag \cite{lt} is utilized to quantitatively evaluate IFM using the Velodyne ULTRA Puck LiDAR (mechanical LiDAR) with the AprilTag marker.
\par

\subsection{Qualitative Evaluation}
\begin{figure}[H] 
	\centering
	\includegraphics[width=1.0\linewidth]{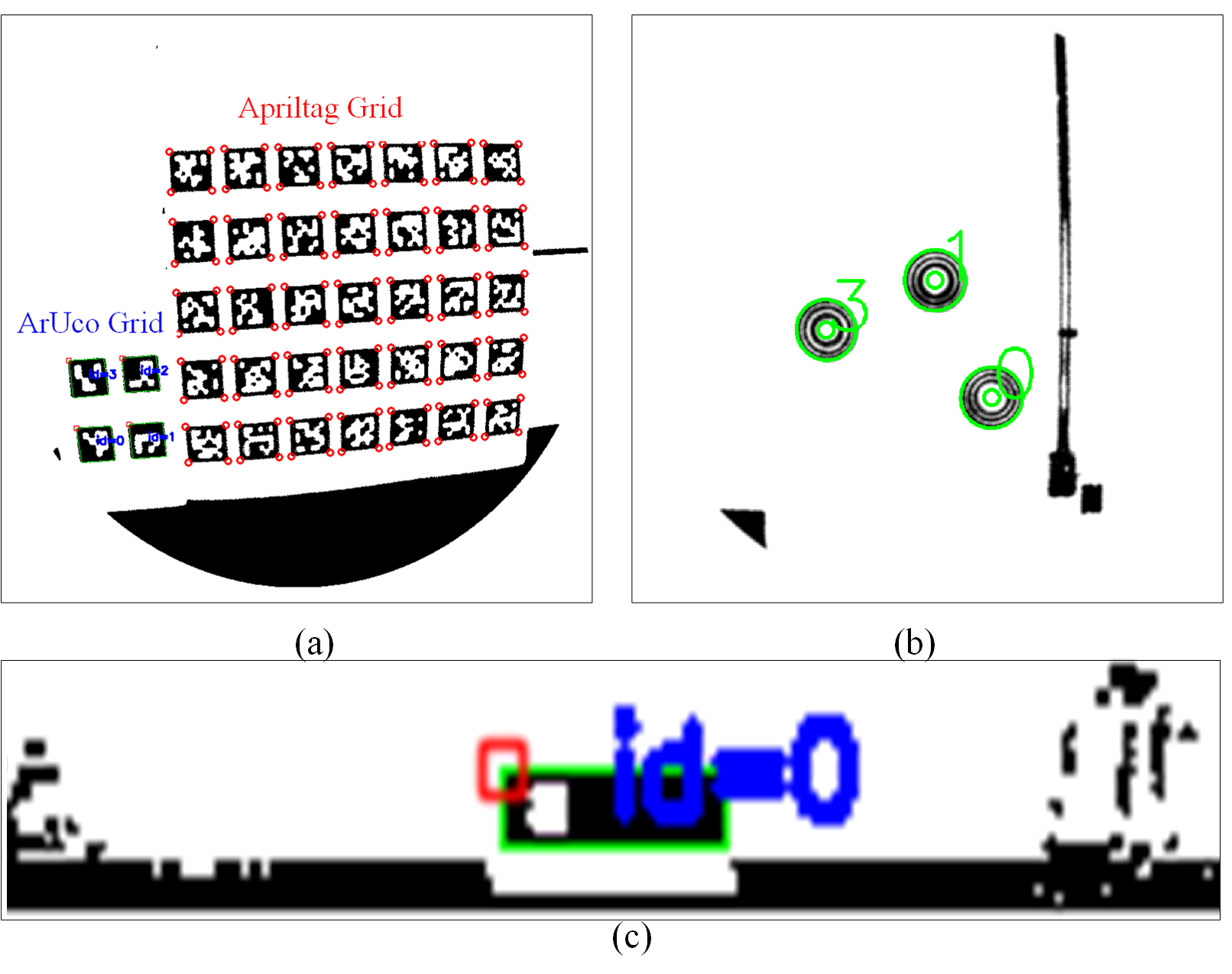}
	\caption{Marker detection results on the preprocessed intensity images.}
	\label{result}
\end{figure}

To qualitatively demonstrate the flexibility of the proposed IFM system, Fig.~\ref{result} is presented. 
Specifically, Fig. \ref{result}(a) corresponds to the scenario shown in Fig.~\ref{flex}(a): a Livox Mid-40 is scanning an AprilTag grid and an ArUco grid. All the markers, 35 AprilTags (family: tag36h11, marker size: 17.2 cm$\times$17.2 cm), and 4 ArUcos (family: 4$\times$4, marker size: 16 cm$\times$16 cm), are detected. Fig. \ref{result}(b) corresponds to Fig.~\ref{flex}(b): a Livox Mid-40 is scanning three CCTag (family: 3 rings, marker radius: 7.2 cm) attached to the wall. All the markers are detected. Fig. \ref{result}(c) corresponds to Fig.~\ref{flex}(c): a VLP-16 is scanning an ArUco (family: original, maker size: 40 cm$\times$40 cm). The marker is detected.
\par
As seen in Fig. \ref{result}, the usage of the IFM system is as convenient as the VFM systems. In particular, the user can place the letter-size markers \cite{ap3,aruco} densely to compose a marker grid, as shown in Fig.~\ref{result}(a), as well as attach some non-square markers \cite{cctag} to the wall freely, as shown in Fig.~\ref{result}(b). In summary, there is no spatial restriction on marker placement. It should be noted that if the angular resolution of the LiDAR is relatively large, large marker size and simpler marker pattern are recommended for the sake of the intensity image quality. For instance, the vertical angular resolution of the VLP-16 is 1.33$^{\circ}$, thus a 40 cm$\times$40 cm original ArUco \cite{aruco} is adopted, as shown in Fig.~\ref{result}(c).

\subsection{Quantitative Evaluation}
\begin{figure}[H] 
	\centering
	\includegraphics[width=1.0\linewidth]{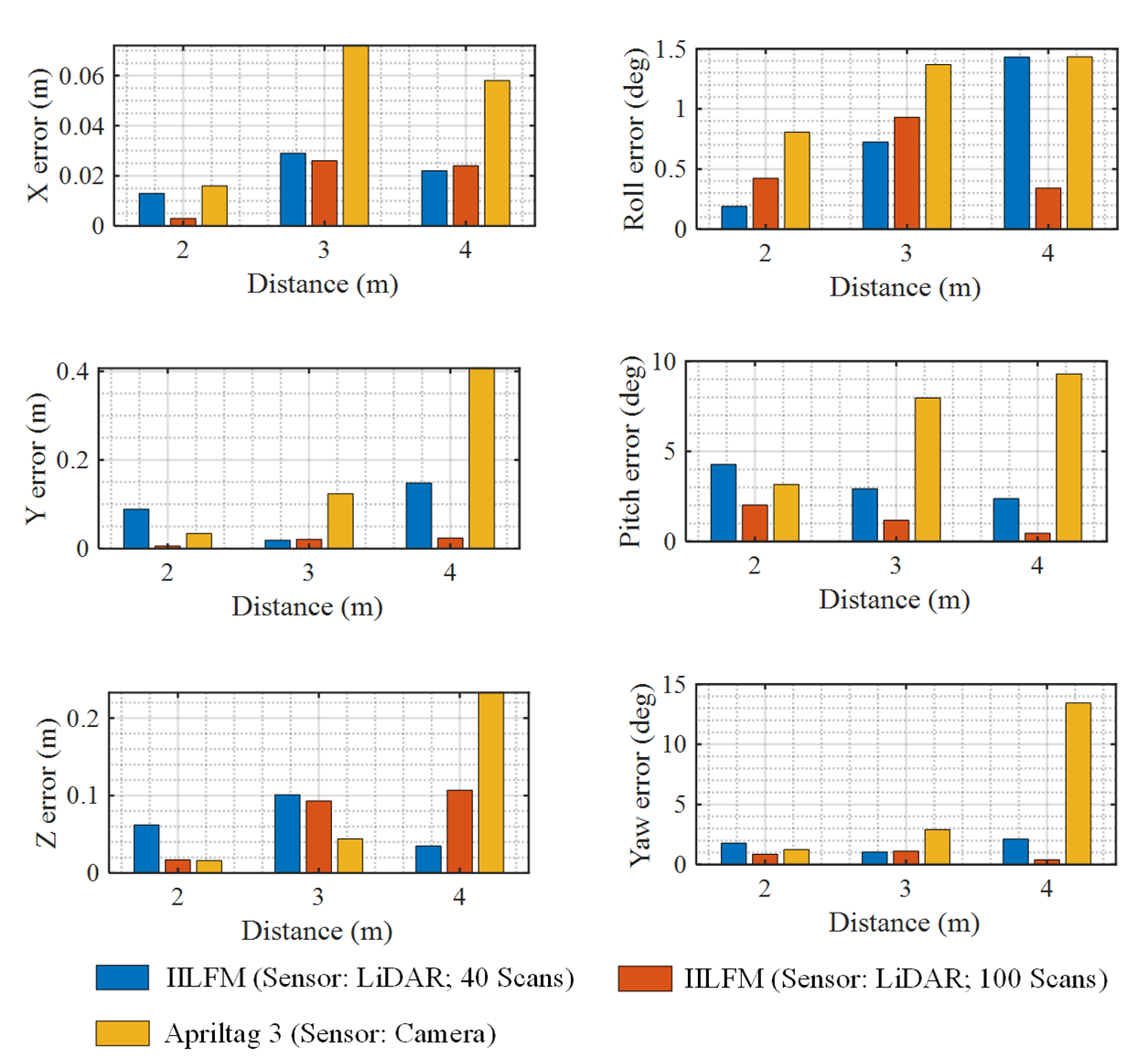}
	\caption{Pose estimation accuracy of the IFM system and the AprilTag 3 system at different distances.}
	\label{bar1}
\end{figure}

As a reminder, the experimental setup is shown in Fig. \ref{setup}. Firstly, the orientation of the LiDAR (Livox Mid-40) is fixed, with the roll, pitch, and yaw angles approximately being zeros. Only the distance from the marker's plane to the LiDAR's $O_{L}Y_{L}\mbox{-}O_{L}Z_{L}$ plane is changed. Then, to compare the conventional VFM system with the proposed system, we test AprilTag 3 \cite{ap3} with a camera (Omnivision OV7251) under the same experimental setup. To intuitively demonstrate the changing trend of accuracy w.r.t. the distance, we present the histogram of errors in Fig.~\ref{bar1}. In particular, the error refers to the absolute difference between the measurement and the ground truth. 
A detailed table containing all the measurements and ground truth from Fig.~\ref{bar1} is available in Table \ref{imtab3}, where the vanilla IFM refers to the approach introduced in this section.
Three issues are illustrated in Fig.~\ref{bar1}: (1) When more LiDAR scans are utilized, the pose estimation accuracy is slightly boosted. The reason behind it is that more LiDAR scans indicate a higher coverage percentage in the FoV \cite{loam}, which implies better intensity image quality. (2) The IFM system shows comparable accuracy as the VFM system. (3) Unlike the VFM system \cite{wang,olson,ap3}, the pose estimation accuracy does not degrade evidently as the distance increases.
\par
Thereafter, the distance from the marker to the LiDAR is set as 2 meters and only the rotation of the LiDAR is adjusted (See Table~\ref{tab2} where the number of scans = 40). 
\begin{table}[H]
	\caption{Pose estimation accuracy of the IFM system with different Euler angles.}
	\begin{center}
		\begin{tabular}{c|c|c|c|c}
			\hline\hline
			Setup & Term & Ground Truth &\textbf{IFM}&Error\\ \hline
			\multirow{6}{*}{ pitch $\approx -15^{\circ}$} &x (m) &1.629   & 1.661   & -0.032  \\  \cline{2-5} 
			&y (m) &-0.066  & -0.011  & -0.055  \\  \cline{2-5} 
			&z (m) &0.618   & 0.699   & -0.081  \\  \cline{2-5} 
			&roll (deg) &-2.673  & -4.936  & 2.263  \\  \cline{2-5} 
			&pitch (deg) &\textbf{-15.060} & -21.145 & 6.085 \\  \cline{2-5} 
			&yaw (deg) &-0.171  & 0.546   & -0.717 \\  \hline
			\multirow{6}{*}{ pitch $\approx 15^{\circ}$} &x (m) &1.684   & 1.622   & 0.062    \\  \cline{2-5} 
			&y (m) &-0.063  & -0.088  & 0.026  \\  \cline{2-5} 
			&z (m) &0.590   & 0.660   & -0.070  \\  \cline{2-5} 
			&roll (deg) &3.657   & 0.900   & 2.757 \\  \cline{2-5} 
			&pitch (deg) &\textbf{14.849}  & 10.961  & 3.888 \\  \cline{2-5} 
			&yaw (deg) &1.136   & -1.921  & 3.057 \\  \hline
			\multirow{6}{*}{ yaw $\approx -15^{\circ}$} &x (m) &1.638   & 1.612   & 0.027   \\  \cline{2-5} 
			&y (m) &-0.618  & -0.237  & -0.381 \\  \cline{2-5} 
			&z (m) &0.610   & 0.585   & 0.025  \\  \cline{2-5} 
			&roll (deg) &-0.009  & -0.792  & 0.783 \\  \cline{2-5} 
			&pitch (deg) &-0.662  & -2.088  & 1.426\\  \cline{2-5} 
			&yaw (deg) &\textbf{-15.387} & -17.937 & 2.550 \\  \hline
			\multirow{6}{*}{ yaw $\approx 15^{\circ}$} &x (m) &1.627   & 1.632   & -0.005  \\  \cline{2-5} 
			&y (m) &-0.194  & -0.146  & -0.048  \\  \cline{2-5} 
			&z (m) &0.609   & 0.553   & 0.057   \\  \cline{2-5} 
			&roll (deg) &0.490   & -1.994  & 2.484 \\  \cline{2-5} 
			&pitch (deg) &-0.388  & -0.359  & -0.028 \\  \cline{2-5} 
			&yaw (deg) &\textbf{14.924}  & 10.678  & 4.246 \\  \hline\hline
		\end{tabular}
		\label{tab2}
	\end{center}
\end{table}
By comparing the errors in Fig.~\ref{bar1} and Table~\ref{tab2}, it is seen that when the plane of the LiDAR is angled towards the marker plane, the pose estimation performance of the IFM system is as good as that when the LiDAR is perpendicular to the marker.
\par
Although the results in Fig.~\ref{bar1} and Table~\ref{tab2} might be inferior to other high-accuracy solutions, such as total stations and prisms \cite{station}, the merit of the proposed framework lies in its low cost, flexibility, and convenience.
\par
To validate the pose estimation accuracy of our system on the mechanical LiDAR as well as to compare the proposed system with the state-of-the-art LFM system, LiDARTag \cite{lt}, which is also the only existing fiducial marker system for the LiDAR as far as we know, Table~\ref{tab3} is presented. Specifically, we use the rosbag (ccw\_10m.bag) provided by \cite{lt} as the benchmark on which we conduct the comparison. ccw\_10m.bag records the raw data of a 32-Beam Velodyne ULTRA Puck LiDAR scanning a 1.22 m$\times$1.22 m AprilTag (tag16h6), from a distance of 10 meters while the relative angle between the LiDAR plane and the marker is around 45$^{\circ}$. Only the vertices estimation is compared on account that \cite{lt} solely provides the ground truth of the vertices and the vertices estimation is a follow-up process after the pose estimation in the LiDARTag system \cite{lt,lt2}. 
\par
Table~\ref{tab3} illustrates that the IFM system is slightly inferior to LiDARTag in terms of accuracy. This is mainly because the LiDARTag system estimates the pose by finding the transmission that projects points inside the marker cluster into a predefined template (bounding box) at the origin of LiDAR \cite{lt,lt2}, while our system adopts the vertices to compute the pose directly just as the AprilTag \cite{ap3} and ArUco \cite{aruco} systems do. \par
\begin{table}[H]
	\caption{Comparison of the IFM system and LiDARTag.}
	\begin{center}
		\begin{tabular}{c|c|c|c|c|c}
			\hline\hline
				System & Vertex & x (m) &y (m)& z (m) & Error (m)\\ \hline
				\multirow{4}{*}{Ground Truth } &1 &9.739 & 0.758 & -0.161 & --  \\  \cline{2-6} 
				&2 &9.940 & 0.072 & -0.732 & -- \\  \cline{2-6} 
				&3 &10.272 & -0.414 & -0.032 & -- \\  \cline{2-6} 
				\cite{lt} &4 &10.072 & 0.271 & 0.539 & --\\  \hline
				\multirow{4}{*}{LiDARTag } &1 & 9.736 & 0.762 & -0.174 & 0.015\\  \cline{2-6} 
				&2 &9.944 & 0.066 & -0.730 & 0.009 \\  \cline{2-6} 
				&3 &10.271 & -0.405 & -0.016 & 0.019 \\  \cline{2-6} 
				\cite{lt} &4 & 10.063 & 0.291 & 0.539 & 0.022\\  \hline
				\multirow{4}{*}{IFM } &1 &9.728 & 0.766 & -0.184 & 0.013\\  \cline{2-6} 
				&2 & 9.963 & 0.052 & -0.705 & 0.033 \\  \cline{2-6} 
				&3 &10.307 & -0.432 & -0.016 & 0.044 \\  \cline{2-6} 
				&4 &10.045 & 0.263 & 0.564 & 0.041 \\  \hline\hline
		\end{tabular}
		\label{tab3}
	\end{center}
\end{table}
Our system utilizes fewer correspondences compared to \cite{lt,lt2}, and thus the ranging noise on a single point could have more effects on the pose estimation of our system. It should be noted that the intention of proposing the LFM system is to improve the real-world applications, such as SLAM \cite{munoz2019}, multi-sensor calibration \cite{lt2}, and AR \cite{ar}. As shown in Fig. \ref{bar1}, the proposed system outperforms the AprilTag system \cite{ap3}, which is already widely used in the above-mentioned applications. Moreover, the VFM systems \cite{ap3,aruco} use only four vertices of a marker (or a limited number of points if the marker is non-square). Note that it is definitely feasible to acquire more feature points from the inner coding area of the marker, however, this will increase the complexity of the VFM systems. Hence, following a simple but effective design concept, we also use only four vertices of a marker in our proposed system. If higher accuracy is needed, it is feasible to use the proposed marker detection information to find the point clustering of the marker in the point cloud and then input the point clustering into the pose estimation block of LiDARTag. However, as mentioned previously, the superiority of the IFM system is the flexibility and extensibility. For example, marker detection is infeasible for the LiDARTag system \cite{lt} in the scenarios shown in Fig.~\ref{flex} and Fig.~\ref{setup} since the marker placement does not satisfy the requirement of LiDARTag, and in addition, the current version of LiDARTag does not support any non-square marker, such as CCTag \cite{cctag}.

\subsection{Computational Time Analysis}
A computational time analysis is conducted on a desktop with Intel Xeon W-1290P Central Processing Unit (CPU). The LiDARTag \cite{lt} runs at around 100 Hz on it. The time consumption of the marker detection of our system, which includes intensity image generation, preprocessing, 2D features detection, and computation of 3D features, mainly depends on the size of the intensity image and the embedded VFM system. Suppose that the embedded VFM system is AprilTag 3 \cite{ap3}, the marker detection takes around 7 ms (approx 143 Hz) in the case shown in Fig.~\ref{result}(c) where $\Theta_{a}=0.3^{\circ}$, $\Theta_{i}=1.33^{\circ}$, and intensity image size = $1201\times27$. For the case shown in  Fig.~\ref{result}(a), the marker detection takes around 25 ms (approx 40 Hz), where intensity  $\Theta_{a}=\Theta_{i}=0.05^{\circ}$ and image size = $771\times591$. The time consumption of the following pose estimation process is around 1.8 $\mu$\textit{s} as the closed-form solution can be obtained directly through SVD \cite{icp} (Refer to Section~\ref{3.3}).
\subsection{Limitations Analysis} \label{twofourfour}
There are some limitations to the application of LFM systems. First, the distance from the object (marker) to the LiDAR must exceed the minimum detectable range. This limitation is caused by the hardware attributes and it affects all the LiDAR applications not only the LFM systems. Secondly, to utilize the solid-state LiDAR, it is required to wait for the growth of the scanned area inside the FoV to obtain a relatively dense point cloud. Again this is a limitation caused by the hardware attributes. In contrast, the mechanical LiDAR only requires one LiDAR scan to work, however, a larger and simpler marker is recommended if the angular resolution is large.  Finally, the false positives are occasionally found in the experiments while the issue of wrong ID detection (not exactly the same as the false positive but similar) is also reported in LiDARTag \cite{lt}. For the proposed system, the false positive rate is mainly determined by the embedded VFM system, thus novel VFM systems with lower false positive rates, such as \cite{ap3,aruco}, are preferred. Moreover, due to the adoption of 3D-to-2D spherical projection, the vanilla IFM exhibits two limitations:
\par
\begin{figure}[t] 
	\centering
\includegraphics[width=1.0\linewidth]{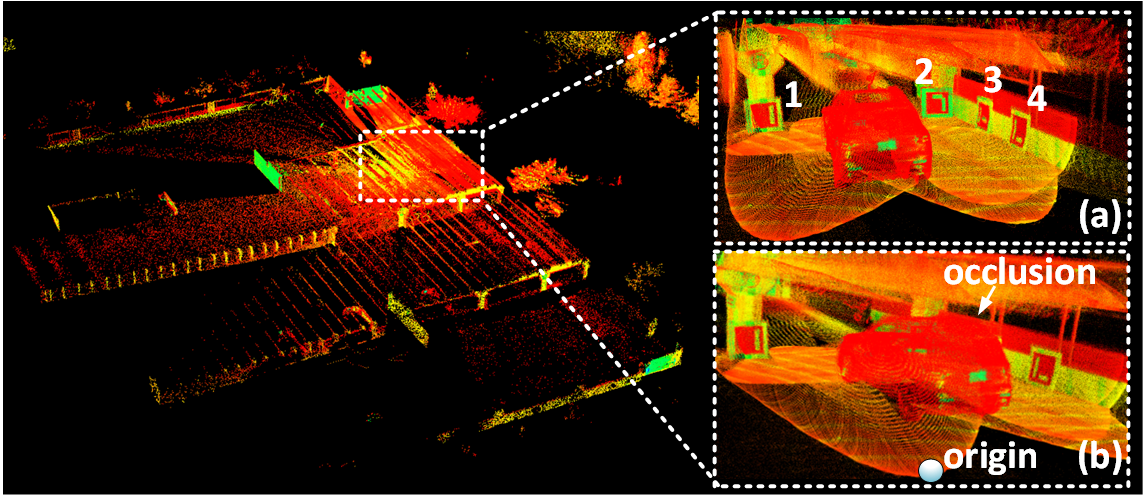}
	\caption{An illustration of the limitation of spherical projection for 3D maps. This map is constructed using Traj LO \cite{traj}.}
	\label{occ}
\end{figure}

(1) IFM can only detect fiducials in a single-view point cloud and is not applicable to a 3D LiDAR map, as spherical projection only applies to a single-view point cloud \cite{rangenet}. Take the 3D map shown in Fig. \ref{occ} as an example. Unless we manually adjust the perspective to view the map as shown in Fig. \ref{occ}(a), the four tags are not simultaneously visible due to occlusion, as seen in Fig. \ref{occ}(b).
\par
(2) As the distance between the tag and the LiDAR increases, the tag's projection size decreases until it becomes too small to be detected.
 \par
The next section addresses these two limitations caused by the 3D-to-2D spherical projection. Moreover, the vanilla IFM employs a constant threshold to process intensity images (See Fig. \ref{dis}(c)), which is sufficient for static scenes with trivial or no viewpoint changes. However, the marker detection of the vanilla IFM is not robust when the viewpoints change in a wide range, which hinders its application to in-the-wild multiview point cloud registration. An adaptive threshold marker detection method is developed to address this problem in Section \ref{5.2}.

\chapter[Improvements to Vanilla IFM Localization]{Improvements to Vanilla IFM \\Localization
\label{improve}}
\section{Overview} \label{4.1}
\begin{figure}[H] 
	\centering
\includegraphics[width=0.8\linewidth]{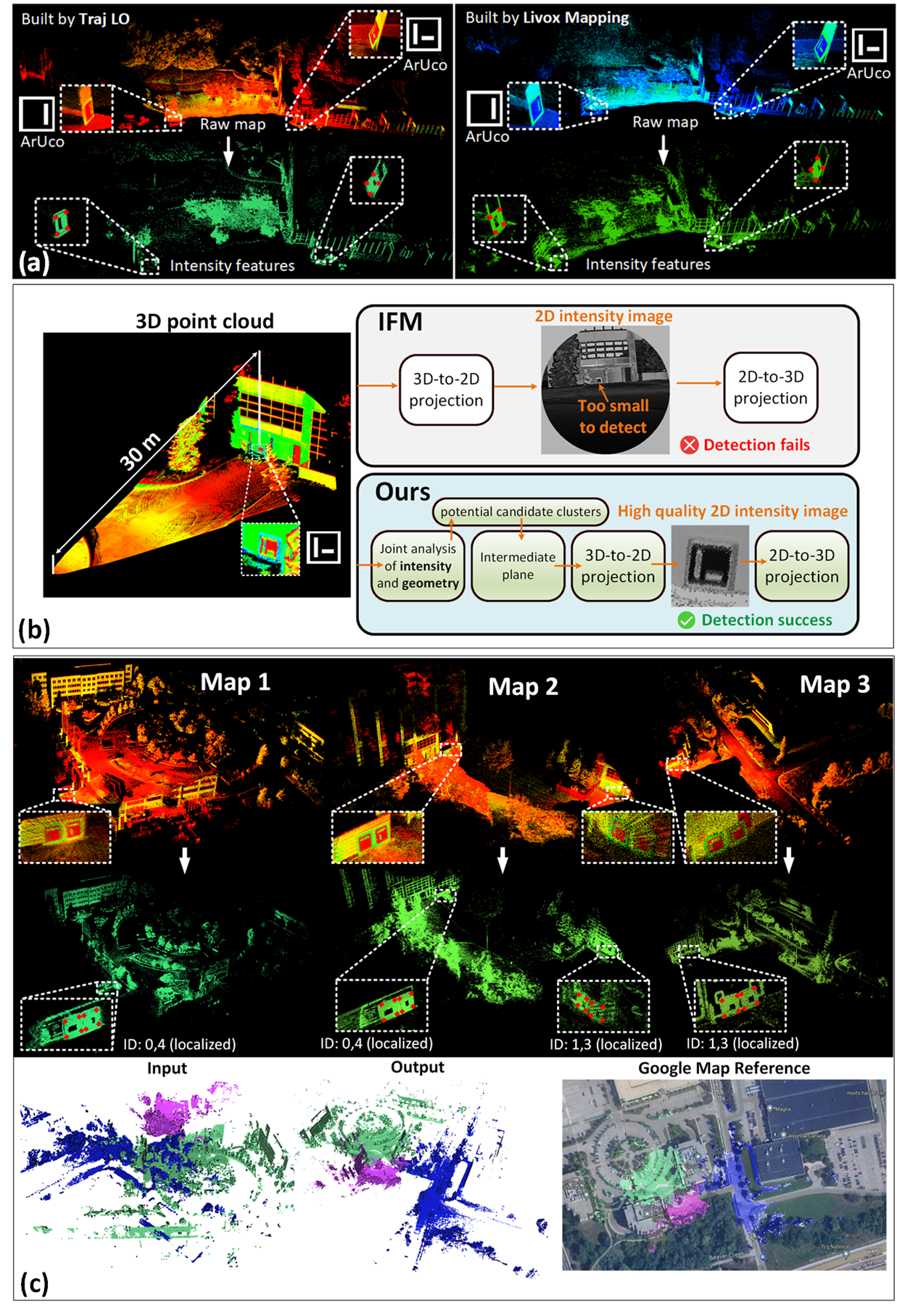}
	\caption{An overview of the improvements to the vanilla IFM.}
	\label{mov}
\end{figure}

This chapter introduces an algorithm that addresses the two limitations of the vanilla IFM, caused by the adoption of 3D-to-2D projection, as introduced in Section \ref{twofourfour}. The overview of the improvements to the vanilla IFM is shown in Fig. \ref{mov}. First, as seen in Fig. \ref{mov}(a), the proposed algorithm extends thin-sheet LFM localization from single-view point clouds to 3D LiDAR maps. One might consider such a solution to localize 3D fiducials in the map: incorporating marker detection into the front end of SLAM and detecting fiducials during SLAM. For example, \cite{map3} applies this solution by integrating AprilTag \cite{ap3} detection into a visual SLAM method (Vins-mono \cite{vins}). The feasibility of this solution is not denied. However, the concern is that it involves low-level modifications to the front end of the SLAM method. Therefore, it is not generalizable, as different LiDAR-based SLAM frameworks \cite{traj,sdk} have various front-end pipelines. Consequently, every time the baseline SLAM method changes, the modification must be made again. In contrast, as shown in Fig. \ref{mov}(a), the proposed algorithm is a generalizable method that can be applied to maps built using different methods. 
Second, compared to the vanilla IFM, which directly projects the 3D point cloud onto an image plane, as shown in Fig. \ref{mov}(b), the proposed method jointly analyzes the point cloud from both intensity and geometry perspectives to extract potential candidate clusters. By introducing an intermediate plane to project these clusters, higher-quality intensity images are obtained. As a result, the proposed method improves the range over which markers can be detected. In this example, the distance is 30 meters, and the marker size is 69.2 cm $\times$ 69.2 cm, whereas the proposed method can localize the marker, but the vanilla IFM cannot.
Third, as seen in Fig. \ref{mov}(c), the extension of LFM localization to 3D maps enables downstream tasks such as 3D map merging. These low-overlap 3D maps are merged using marker localization results from the proposed method and the algorithm in Section \ref{multi}.
\section{Joint Analysis of Point Clouds from Intensity and Geometry Perspectives} \label{4.2}
\subsection{Downsampling Based on 3D Intensity Gradients} \label{ic} 
\begin{figure}[ht] 
	\centering
\includegraphics[width=1.0\linewidth]{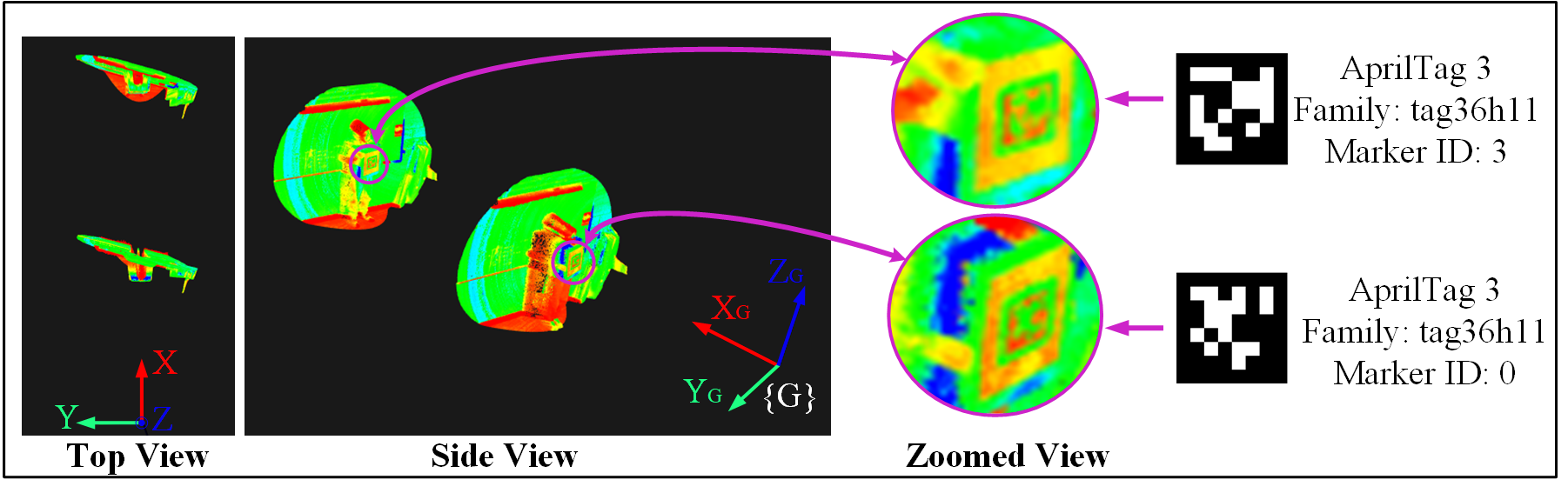}
	\caption{The example used to explain the design purpose and result of each step.}
	\label{sub}
\end{figure}

A synthesis point cloud (see Fig. \ref{sub}) with a simple scene is utilized to explicitly illustrate the design purpose and results of each operation in our pipeline. The two presenters are holding two different AprilTags \cite{ap3}. As illustrated in the top view, observing along the X-axis of the global coordinate system, the back subpoint cloud is totally blocked by the front one. Thus, although this is not a 3D LiDAR map, it possesses the most important feature of a 3D LiDAR map in this research: the spherical projection (Eq. (\ref{pro})) is not applicable. Note that Fig. \ref{sub} is only used to explicitly present the results and explain design purposes due to its simplicity. The ultimate objective is to localize the fiducial markers on a 3D map, which represents a larger and more complex scene. 
\par
The LFMs are sheets of thin paper that are not spatially distinguishable from the attached planes. Thus, it is infeasible to adopt previous geometric features-based 3D object detection methods \cite{single, rangenet} to find the LFMs. Therefore, a new feature extraction solution is developed to address this problem. 
The analysis is given as follows. A fiducial marker is composed of a black-and-white pattern and, as a result, presents as a high-intensity contrast object in the view of a LiDAR (see the zoomed views of Fig. \ref{mov}, Fig. \ref{occ}, and Fig. \ref{sub}). This indicates that the point cloud can first be analyzed from the intensity perspective. \par
In particular, downsampling is conducted on the raw point cloud based on the 3D intensity gradients. The intensity is taken as a function $I(\mathbf{p})$ of the 3D coordinates $\mathbf{p} = [x,y,z]^{T}$. Suppose that the given 3D point/location is $\mathbf{{p}_{0}}=[x_{0},y_{0},z_{0}]^{T}$, the point set composed of the neighbouring $n$ points around $\mathbf{{p}_{0}}$ is defined as $\mathcal{P}_{I}=\{\mathbf{{p}_{1}},\cdots,\mathbf{{p}_{n}}\}$. 
In practice, the following equation is used to approximate $I(\mathbf{p})$:
\begin{equation}
\hat{I}(\mathbf{p}) = \mathbf{C}^{T}\mathbf{p} + b, \label{eq4.1}
\end{equation}
where $\mathbf{C} \in \mathbb{R}^{3 \times 1}$ is the coefficient vector and $b\in \mathbb{R}$ is the intercept. Since the intensity values of points in $\mathcal{P}_{I}$ are known, $\mathbf{C}$ can be estimated by solving the following model fit (least square) problem:
\begin{equation}	
\underset{\mathbf{C}^{*}, b^{*}}{\arg \min } \sum_{i=1}^{n}\left\|\hat{I}(\mathbf{p}_{i})-I(\mathbf{p}_{i})\right\|^{2}\label{least2}.
\end{equation} \par
Coefficient regression \cite{barfoot} is performed to solve this problem. Let the design matrix be: 
\begin{equation}	
		\begin{aligned}
			& \mathbf{D} =\left[\begin{array}{cccc}  1 &\Delta x_{1} & \Delta y_{1} & \Delta z_{1} \\ 
  1 &\Delta x_{2} & \Delta y_{2} & \Delta z_{2} \\  
 \cdot   & \cdot  & \cdot & \cdot \\ 
 \cdot   & \cdot  & \cdot & \cdot \\  
 \cdot   & \cdot  & \cdot & \cdot \\  
  1 &\Delta x_{n} & \Delta y_{n} & \Delta z_{n} \\  
   \end{array}\right],
		\end{aligned}
	\end{equation} 
where $\mathbf{D}\in \mathbb{R}^{n \times 4}$. $\Delta x_{i}=(x_i-x_{0})$, $\Delta y_{i}=(y_i-y_{0})$, $\Delta z_{i}=(z_i-z_{0})$. The first column is the constant term, representing the intercept. Centering is applied to the design matrix, as the focus is on the local gradients of intensity value at a given 3D position within $\mathcal{P}_{I}$. Since the goal is to fit the intensity value as a function of 3D positions, the response variable vector would be:
\begin{equation}	
\mathbf{I}_{in} =\left[ \Delta I_{1}, \Delta I_{2}, \cdots, \Delta I_{n} \right]^{T},
\end{equation}
where $\Delta I_{i}=(I_i-\bar I)$ with $\bar I \in \mathbb{R}$ being the mean of the intensity values of the points in $\mathcal{P}_{I}$. Again, centering is performed as the focus is on the local gradients. Suppose that the regression coefficient vector $\mathbf{E}\in \mathbb{R}^{4 \times 1}$ is defined as:
\begin{equation}	
\mathbf{E} = \left[\begin{array}{c}  \mathbf{C}^{*}  \\ 
b^{*}
   \end{array}\right].
\end{equation} \par
Following \cite{barfoot}, the regression coefficient vector can be calculated by:
\begin{equation}	
\mathbf{E} =(\mathbf{D}^{T}\mathbf{D})^{-1}\mathbf{D}^{T}\mathbf{I}_{in}.
\end{equation} \par
$\hat{I}(\mathbf{x})$ is obtained by substituting the elements of $\mathbf{E}$ into Eq. (\ref{eq4.1}). Denote the 3D intensity gradients at the given 3D position as 
$\nabla I \in \mathbb{R}^{1 \times 3}$:
\begin{equation}
\begin{aligned}
 \nabla I &= \frac{\partial \hat{I}(\mathbf{p})}{\partial \mathbf{p}}\\
 &=\frac{\partial \mathbf{C}^{* T}\mathbf{p}}{\partial \mathbf{p}} + \frac{\partial b^{*}}{\partial \mathbf{p}}\\
 &=\mathbf{C}^{* T}
 \end{aligned}
\end{equation} \par

\begin{figure}[H] 
	\centering
\includegraphics[width=1.0\linewidth]{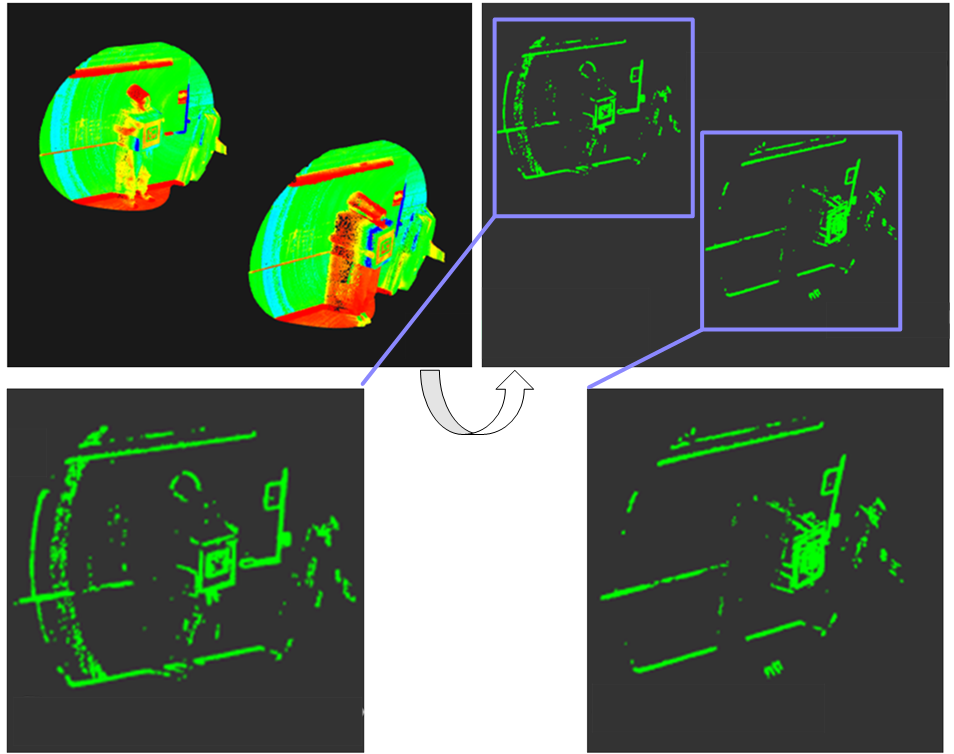}
	\caption{The effect of applying downsampling from the intensity perspective.}
	\label{step1}
\end{figure}

The direction of $\nabla I$  of a given point indicates the direction where the intensity has the fastest decline and the norm, $|\nabla I|$, implies the rate of descent. 
Inspired by LOAM \cite{loam}, which selects a point as a feature if its geometric curvature is larger than a threshold, a 3D point is preserved if its $|\nabla I|$ is larger than a threshold in the downsampling procedure. The result following downsampling execution is depicted in Fig. \ref{step1}. As seen, the majority of unnecessary points are filtered out.\par
\par

\subsection{Spatial Distribution Analysis of Downsampling Result} \label{clu}
\begin{figure}[ht] 
	\centering
\includegraphics[width=0.8\linewidth]{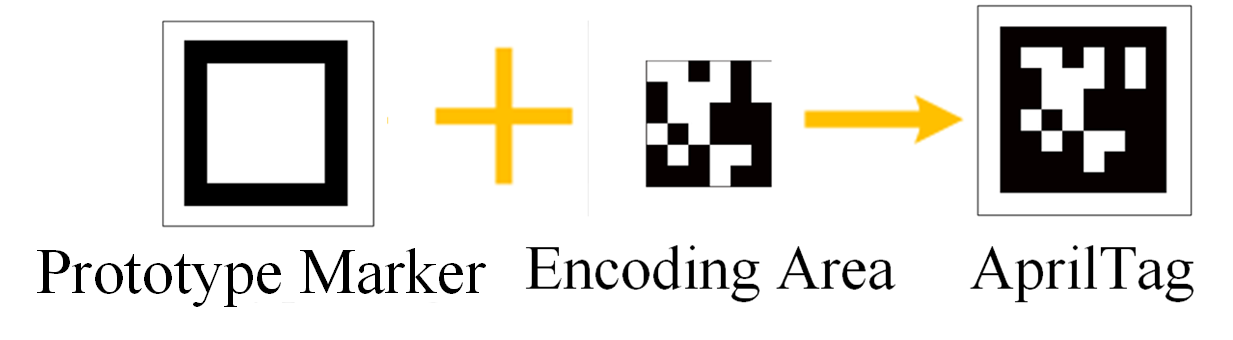}
	\caption{A diagram to illustrate the design of a typical square fiducial marker \cite{ap3}. }
	\label{td}
\end{figure}

The downsampling preserves the points belonging to the outlines of all objects with high intensity-contrast. In this section, the point cloud is analyzed from a geometric perspective.
The foundation of doing so is the fact that the points belonging to the fiducial markers will be isolated from those of the other objects after the downsampling (See the zoomed view of  Fig. \ref{step1}). This is due to the design of the marker's pattern as shown in Fig. \ref{td}. 
\par
\begin{figure}[H] 
	\centering
\includegraphics[width=1.0\linewidth]{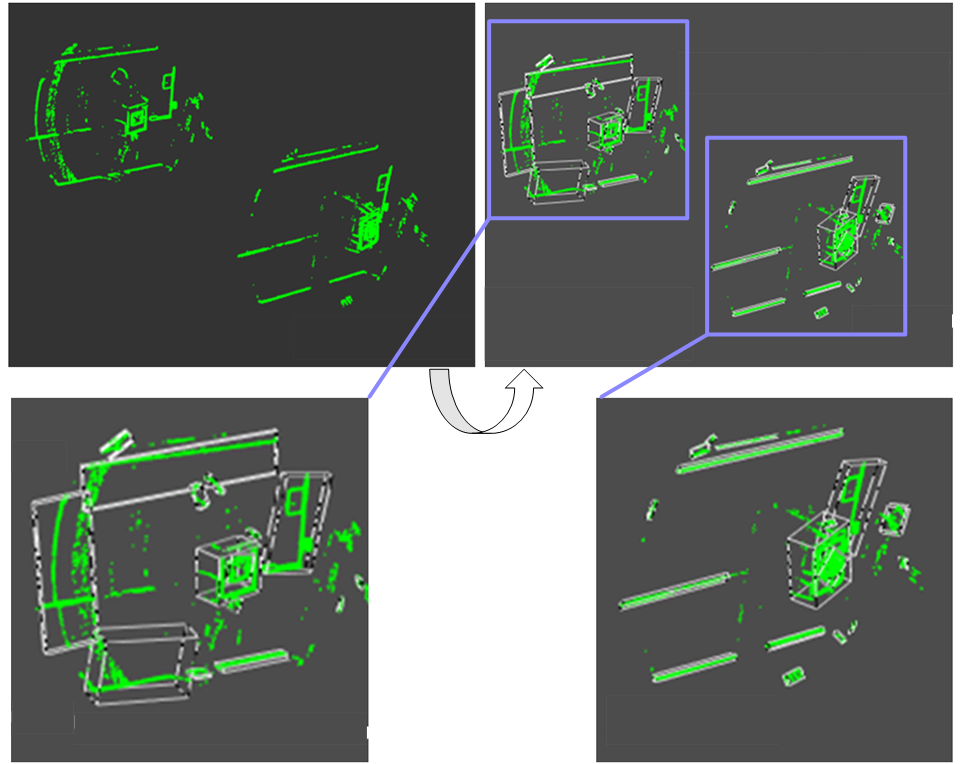}
	\caption{The effect of clustering on the downsampling result.}
	\label{step2}
\end{figure}

In particular, a square fiducial marker is a combination of the prototype marker (a black frame inside a white frame) and the encoding area. The white regions of the prototype marker naturally have higher intensity values than the black regions, and thus, the prototype marker after the downsampling is rendered as a square double-ring that isolates the points inside the coding area from the environment. The isolation makes the points belonging to the markers spatially distinguishable from those of the other objects. Thus, we employ the method introduced in \cite{rusu} to further segment the downsampling result into clusters. Each cluster is represented by an Oriented Bounding Box (OBB) in Fig. \ref{step2}. 
\par

\subsection{Filtering Out Unwanted Clusters} 
As depicted in Fig. \ref{step2}, there are many OBBs after clustering. In this section, the geometric characteristics of each OBB are analyzed to verify if it has the potential to be a fiducial marker. Specifically, the bounding box of a cluster needs to satisfy two criteria to be recognized as a valid candidate. \par
\noindent\textbf{Criterion 1.} The first criterion is subject to the marker size: 
\begin{equation}	
		\sqrt{2a^{2}+t_{M}^{2}}\leq L_{OBB} \leq \sqrt{4a^{2}+t_{M}^{2}},
		\label{first}
	\end{equation} 
where $L_{OBB}=\sqrt{l^{2}+w^{2}+h^{2}}$ is the cuboid diagonal of the OBB with $l$, $w$, and $h$ ($h \leq t_{M}$) being the length, width, and height, respectively. $a$ denotes the side length of the marker. $t_{M}$ is the trifling thickness of the marker. \par
\begin{figure}[ht] 
	\centering
\includegraphics[width=0.7\linewidth]{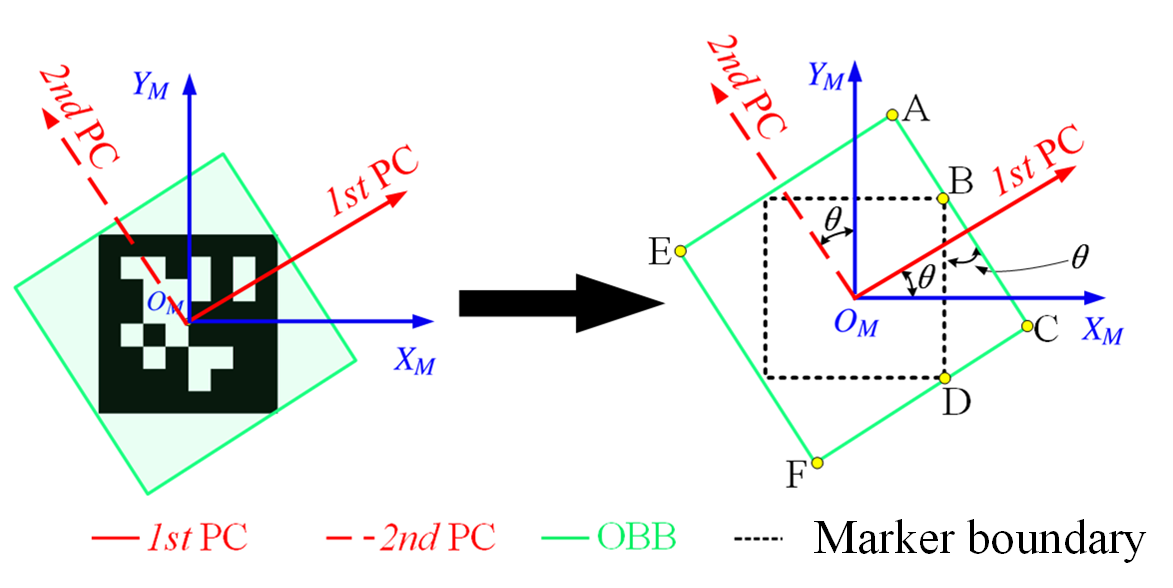}
	\caption{A diagram of the possible OBB size for a given marker.}
	\label{proof}
\end{figure}

\noindent\textbf{Explanation of Criterion 1.} The trivial height is neglected for now, and the possible size range of the OBB is considered in 2D space. Fig. \ref{proof} shows the 2D geometric relations.
 \par
Principal Component Analysis (PCA) is performed on the distribution of points belonging to the marker.
Suppose that $O_{M}-X_{M}-Y_{M}$ is the marker coordinate system and $\theta$ is the angle between the first Primary Component (PC) and the x-axis. On account of the fact that the first PC and second PC are perpendicular to each other \cite{bounding}, the angle between the second PC and the y-axis is also $\theta$. In addition, $\angle DBC=\theta$. Thus, the side length of the OBB, expressed as $a (\mathrm{cos}\theta +\mathrm{sin}\theta)$, falls within the range $[a, \sqrt{2}a]$, given that $\theta \in [0, 2\pi]$. Furthermore, the area of the 2D OBB, $S_{OBB}$, is in the following range:
	\begin{equation}	
		a^{2}\leq S_{OBB} \leq 2a^{2}.
		\label{range}
	\end{equation} \par
Returning to 3D space, the marker's thickness, $\delta$, is also taken into account. The first criterion becomes Eq. (\ref{first}).
\par
\noindent\textbf{Criterion 2.} The second criterion is shown in Eq. (\ref{second}): 
	\begin{equation}	
		 1/1.5 \leq l/w \leq 1.5.
		\label{second}
	\end{equation} \par

\begin{figure}[H] 
	\centering
\includegraphics[width=1.0\linewidth]{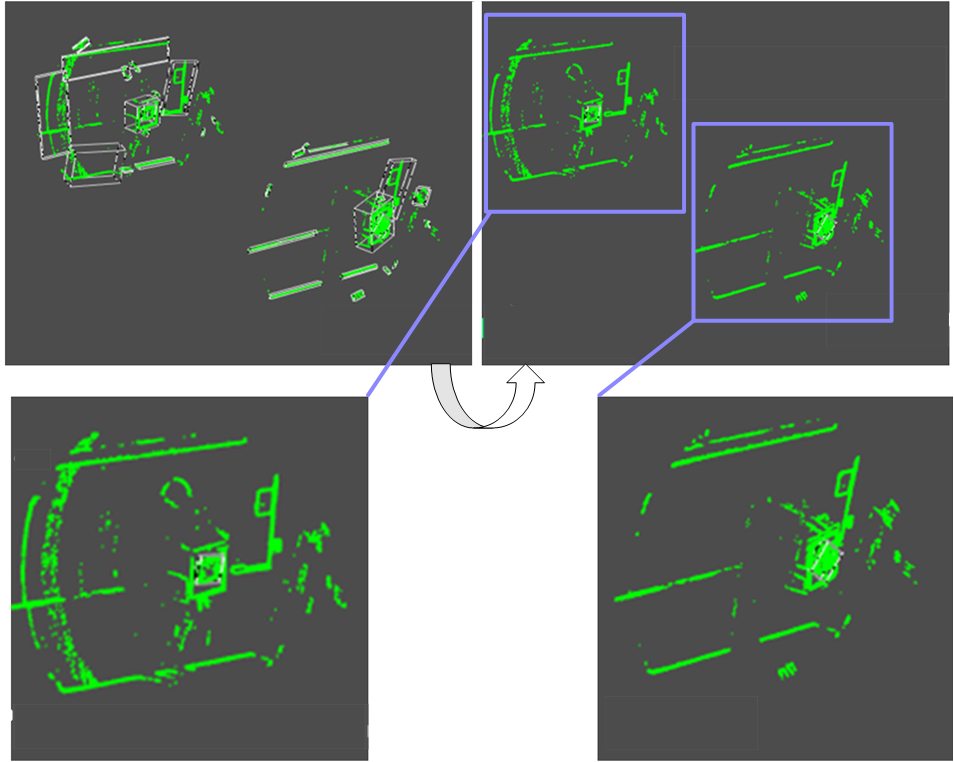}
	\caption{The effect of filtering out the unwanted clusters.}
	\label{step3}
\end{figure}

\noindent\textbf{Explanation of Criterion 2.} This criterion is based on the fact that the shape of the marker is square and the OBB projection on the plane of length and width is also square. Ideally, $l \approx w$. Whereas in the real world, the marker cannot be perfectly scanned by LiDAR, and LiDAR also has ranging noise. 
As a result, the shape of the OBB on the length and width plane could be distorted. Hence, the requirement on $l \approx w$ is relaxed as shown in Eq. (\ref{second}). It is presented in Fig. \ref{step3} that the unwanted OBBs are filtered out after the operation introduced in this section.
 \par

\section{Marker Localization via Intermediate Plane} \label{4.3}
After analyzing the point cloud from the 3D intensity gradients and 3D geometric perspectives, locations of objects with high intensity-contrast and shapes/sizes similar to the fiducial marker are obtained. 
These locations (OBBs) are then inspected in the raw point cloud. When extracting points falling into the OBBs, a buffer is adopted to extend the OBBs, preserving more regions around an OBB in case it does not completely cover the fiducial marker. This indicates that, for each OBB, the pose (position and orientation), $\mathbf{T}_{OBB}$ \footnote{$\mathbf{T}_{OBB}$ denotes the transmission from the world coordinate system $\{G\}$ to the OBB frame. The OBB frame refers to a coordinate system whose X, Y, and Z axes are parallel to the length, width, and height of the OBB. }, is kept while the size is enlarged by multiplying the length, width, and height with an amplification factor, $t_{b}$. The recommended value of $t_{b}$ is twice the marker's side length based on our experiment. \par
\subsection{Motivation for Adopting the Intermediate Plane}
\begin{figure}[H] 
	\centering
\includegraphics[width=1.0\linewidth]{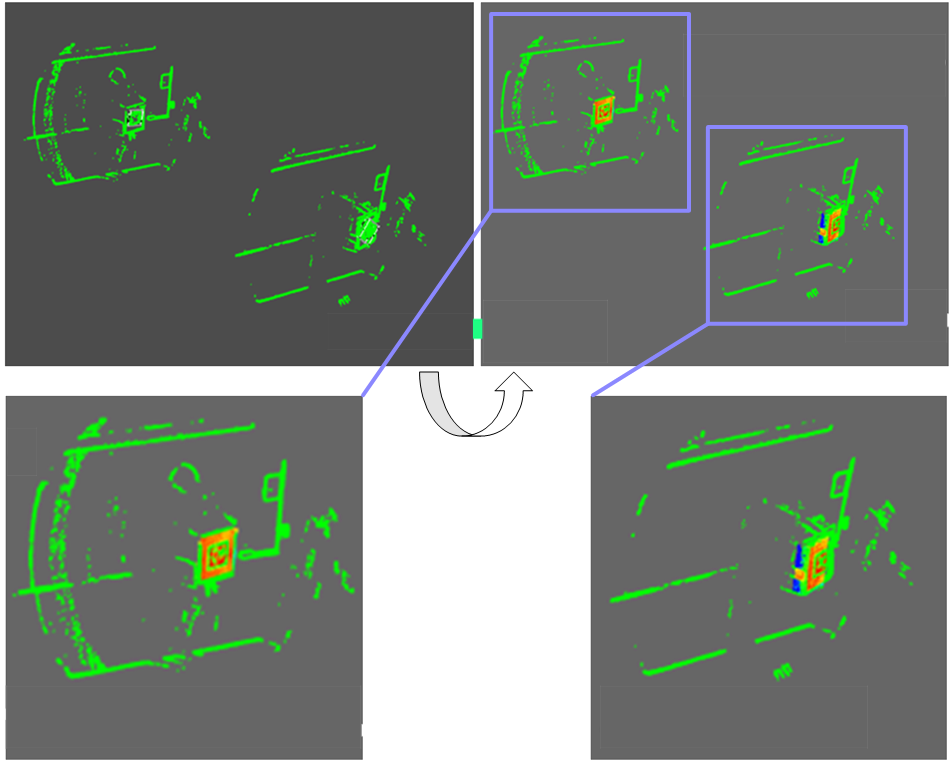}
	\caption{The result of extracting points falling into the preserved OBBs from the raw point cloud.}
	\label{step4}
\end{figure}

\begin{figure}[H] 
	\centering
\includegraphics[width=1.0\linewidth]{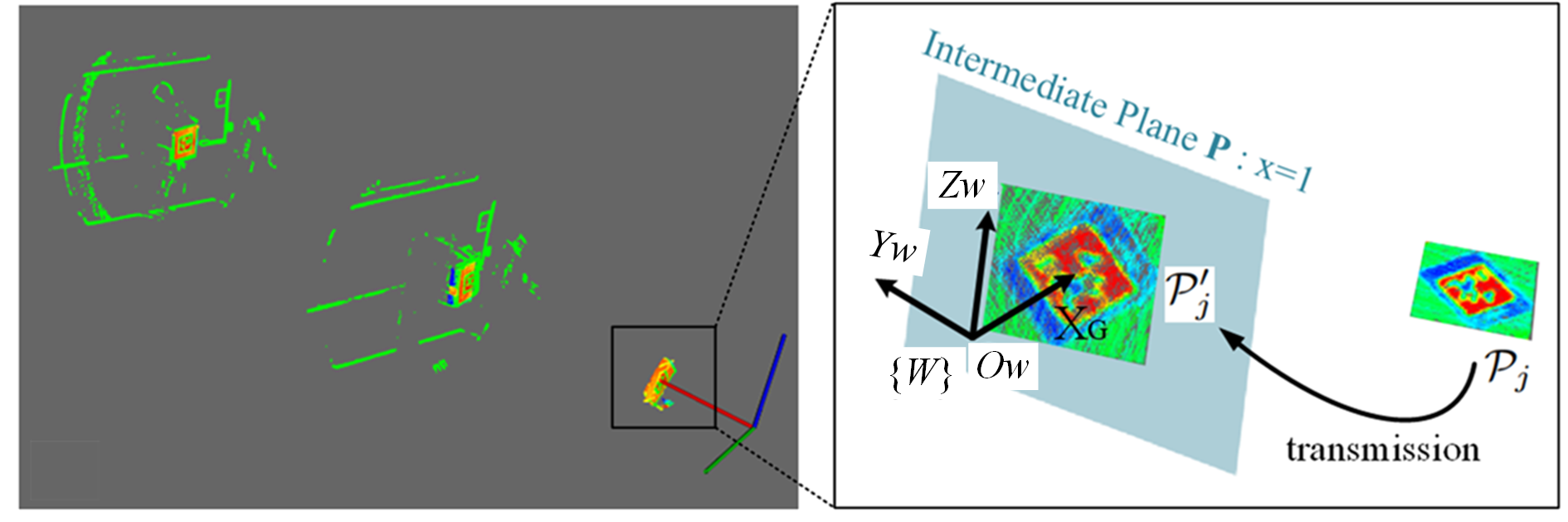}
	\caption{An illustration of how the intermediate plane helps solve the occlusion issue.}
	\label{inter1}
\end{figure}

\begin{figure}[H] 
	\centering
\includegraphics[width=1.0\linewidth]{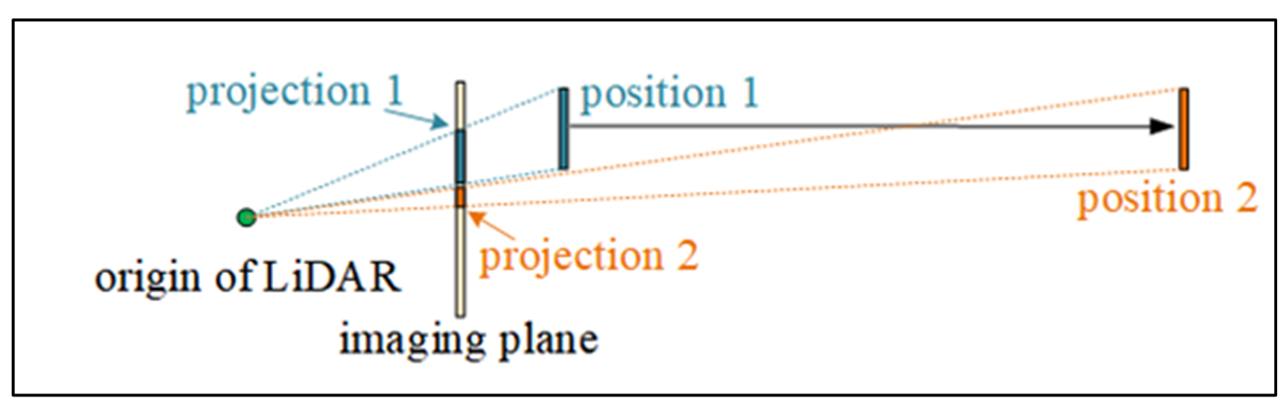}
	\caption{An illustration of how the intermediate plane helps solve the occlusion issue.}
	\label{inter2}
\end{figure}

An intermediate plane-based method is proposed to determine if an OBB contains a fiducial marker. There are two reasons for adopting the intermediate plane. First, as seen in Fig. \ref{step4}, although the points belonging to the candidate markers are extracted from the raw point cloud, unfortunately, the spherical projection (Eq. (\ref{pro})) cannot be applied in this case, as occlusion still exists when observing the point cloud from the origin. (Review Fig. \ref{sub} if needed). 
In contrast, as depicted in Fig. \ref{inter1}, the adoption of the intermediate plane resolves this problem by transferring the clusters onto the intermediate plane one by one, thereby addressing the occlusion issue.
Second, as shown in Fig. \ref{mov}(b), when the marker is far from the LiDAR, directly applying the spherical projection will result in a projection that is too small to be detected. However, as presented in Fig. \ref{inter2}, the introduction of the intermediate plane addresses this challenge. Specifically, transferring the point cloud of a fiducial marker onto the intermediate plane physically reduces the distance between the point cluster and the LiDAR, thereby resulting in a projection of better quality.
\subsection{Utilization of the Intermediate Plane}
\begin{figure}[H] 
	\centering
\includegraphics[width=1.0\linewidth]{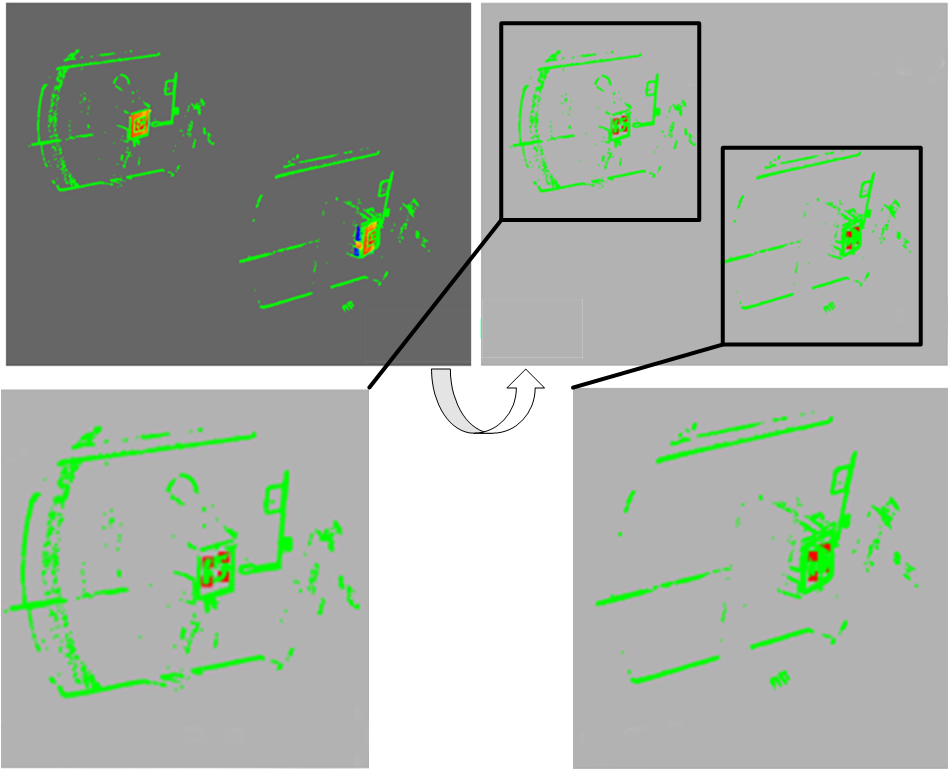}
	\caption{The fiducial marker localization result after applying the intermediate plane method. The vertices of the localized fiducial marker are rendered red.}
	\label{step5}
\end{figure}

Given that the pose of each OBB ($\mathbf{T}_{OBB}$) is known, the perspective for observing these candidate clusters can be adjusted using $\mathbf{T}_{OBB}$. 
In particular, the viewpoint is fixed at the origin, and the points extracted from each OBB are transferred one by one to the intermediate plane (denoted by $\mathbf{P}$), as shown in Fig. \ref{inter1}.
The detailed transmission process is as follows. Define the point set of the $j$-th OBB as $\mathcal{P}_{j}$ and a point of $\mathcal{P}_{j}$ as $\mathbf{{p}_{j}}\in \mathbb{R}^{3}$. $\mathbf{{p}_{j}}$ is first transmitted to the origin of the world coordinate system $\{G\}$ through the inverse of $\mathbf{T}_{OBB}$:
	\begin{equation}	
			\mathbf{{p}_{G}}  = \mathbf{T}_{OBB}^{-1} \cdot \mathbf{{p}_{j}}= \mathbf{R}^{-1} \cdot \mathbf{{p}_{j}}-\mathbf{R}^{-1}\mathbf{t}, 
		\label{trans}
	\end{equation}
where $\mathbf{{p}_{G}}$ is the point transmitted to the origin of $\{G\}$. $\mathbf{T}_{OBB} = [\mathbf{R}|\mathbf{t}]$ where $\mathbf{R}$ is the $3 \times 3$ orthogonal rotation matrix and $\mathbf{t}$ is the $3 \times 1$ translation vector. After transmitting all the points in $\mathcal{P}_{j}$ using Eq. (\ref{trans}), a new point set $\mathcal{P}^{G}_{j}$ is obtained. Then, all points belonging to $\mathcal{P}^{G}_{j}$ are transferred to the intermediate plane $\mathbf{P}$:
	\begin{equation}	
			\textbf{p}^{\prime}_{j} = \mathbf{T}_{in} \cdot \textbf{p}_{G}=\mathbf{R}_{in} \cdot \textbf{p}_{G}+\mathbf{t}_{in}, \\
		\label{trans2}
	\end{equation}
where $\textbf{p}^{\prime}_{j}$ is the point transmitted to the intermediate plane. $\mathbf{T}_{in} = [\mathbf{R}_{in}|\mathbf{t}_{in}]$ where $\mathbf{R}_{in} = \mathbf{I}_{3 \times 3}$ is an identity matrix and $\mathbf{t}_{in} = [1 \mathrm{m}, 0, 0]^{T} $ since the plane equation is $x=1 \mathrm{m}$. Define the point set on the intermediate plane as $\mathcal{P}^{\prime}_{j}$. Given that $\mathcal{P}^{\prime}_{j}$ is a point cloud with no occlusions, marker localization in it is straightforward using the vanilla IFM. IFM returns the marker ID and the locations of the vertices labeled by index if the marker is found within the candidate OBB. Then, the locations of vertices can be transferred back to the original positions in the 3D map using the reverse processes of Eqs. (\ref{trans}-\ref{trans2}), and the poses of markers (from $\{G\}$ to the marker coordinate system) are obtained by solving SVD. The final marker localization result is presented in Fig. \ref{step5}. 
\par
\begin{figure}[H] 
	\centering
\includegraphics[width=0.4\linewidth]{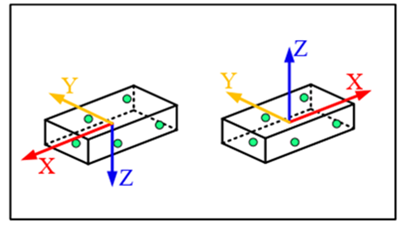}
	\caption{An illustration of the potential rotational ambiguity issue of $\mathbf{T}_{OBB}$.}
	\label{add}
\end{figure}

Note that as depicted in Fig. \ref{add}, the rotation of $\mathbf{T}_{OBB}$ may be ambiguous, potentially causing a flipped intensity image due to the symmetry of the 3D OBB. In practice, both the raw intensity image and the flipped intensity image are checked, as a pattern in the coding library will not have its flipped version present \cite{ap3,aruco}.

\section{3D LiDAR Map Construction} \label{new4.4}
This section briefly introduces how the 3D LiDAR maps tested in Section \ref{4.4} are constructed. In particular, two methods—Livox Mapping \cite{sdk} and Traj LO \cite{traj}—are utilized in this dissertation for constructing large-scale 3D maps.
\subsection{Livox Mapping}
Livox Mapping \cite{sdk} is the official implementation of LiDAR-based SLAM released by Livox, which is built on LOAM \cite{loam}. Suppose that the set of sampled data of the current scan is:
\begin{equation}
\mathcal{P}_{loam} = \{ \{x_{1},y_{1},z_{1},I_{1}\}, \{x_{2},y_{1},z_{2},I_{2}\}, \cdots, \{x_{n},y_{n},z_{n},I_{n}\} \}. \label{loamdata}
\end{equation} \par
To eliminate motion blur effects, the raw data is preprocessed by applying motion compensation. This involves either piecewise processing, where a frame is divided into subframes and processed in parallel to reduce the time interval, or linear interpolation, which modifies 3D point positions using the constant velocity assumption. Then, corner and plane feature points are extracted based on the curvature and normal of each point:
\begin{equation}
\begin{aligned}
&\mathcal{C}_{loam} = \{ \mathbf{p}_{i} \in \mathcal{P}_{loam} | \ \ \kappa(\mathbf{p}_{i})> t_{cor} \}, \\
&\mathcal{S}_{loam} = \{ \mathbf{p}_{i} \in \mathcal{P}_{loam} | \ \ \mathbf{N}(\mathbf{p}_{i})< t_{sur} \},
\end{aligned}
\end{equation} 
where $\mathcal{C}_{loam}$ and $\mathcal{S}_{loam}$ are the sets of corner features and surface features, respectively. $\kappa(\mathbf{p}_{i})$ and $\mathbf{N}(\mathbf{p}_{i})$ are the functions to calculate the curvature and normal of $\mathbf{p}_{i}$, respectively. $t_{cor}$ and $t_{sur}$ are the predefined thresholds. Then, voxel grid filtering is applied to downsample the feature points. The downsampled features are matched and registered with the features in the previous scan to estimate the relative pose between the current scan and the previous scan. Moreover, to eliminate the effect of dynamic objects, the features with too large residuals are removed. Finally, the preserved points are accumulated using the estimated poses to construct the 3D map.
\subsection{Traj LO}
Traj LO \cite{traj} is the state-of-the-art pure LiDAR-based SLAM method. The set of sampled data of the current scan is:
\begin{equation}
\mathcal{P}_{Traj} = \{ \{x_{1},y_{1},z_{1},I_{1},t_{1}\}, \{x_{2},y_{1},z_{2},I_{2},t_{2}\}, \cdots, \{x_{n},y_{n},z_{n},I_{n},t_{n}\} \},
\end{equation}
where $t_{i}$ is the sampling time of $\mathbf{p}_{i}$. As seen, compared to Eq. (\ref{loamdata}), Traj LO also utilizes the temporal information. First, both piecewise processing and linear interpolation are performed for motion compensation. In particular, the current scan (time window) is divided into $K$ equidistant segments of small resolution. Thus, the time interval becomes tiny by choosing a large value of $K$, enhancing the constant velocity assumption and allowing the method to handle rapid motion effectively. Traj LO proposes that the LiDAR trajectory (poses) is a function of time. Specifically, the LiDAR trajectory at the current time window, denoted by $\mathbf{T}(t)$, consists of $K$ piecewise functions: 
\begin{equation}
\mathbf{T}(t) = \{ \tau(t_{1}), \tau(t_{2}), \cdots, \tau(t_{K})  \},
\end{equation}
where $\tau(t_{i})$ is the function representing the LiDAR trajectory at the $i$-th time segment. Traj LO applies point-to-plane ICP for registration without selecting feature. $\mathbf{T}(t)$ is estimated by jointly minimizing the following energy function:
\begin{equation}
\underset{\mathbf{T}(t)^{*}}{\arg \min } \left( \mathcal{E}_{reg} + \mathcal{E}_{kine} + \mathcal{E}_{marg} \right),
\end{equation}
where $\mathcal{E}_{reg}$, $\mathcal{E}_{kine}$, and $\mathcal{E}_{marg}$ represent the registration energy, kinematics energy, and marginalization energy, respectively. Since each 3D point is associated with a sampling timestamp, its location in the world coordinate system can be determined using $\mathbf{T}(t)$. The 3D map is constructed by transforming the observed 3D points into the world coordinate system.

\section{Experimental Validation} \label{4.4}
\subsection{Extending LFM Localization to 3D Maps:\\ Tackling the First Limitation} \label{321}
As a reminder, the first limitation of the vanilla IFM is its restriction to single-view point clouds, rather than 3D maps. This section demonstrates how the proposed method addresses this limitation.
In particular, as presented in Table \ref{imtab1}, the proposed method is tested on eight 3D maps built from rosbags collected using a Livox MID 40 LiDAR \cite{sdk}. There are 2 to 4 ArUco markers, measuring 69.2 cm × 69.2 cm, in each map. The number of localized markers out of the total in each map is also included in Table \ref{imtab1}. The error in Table \ref{imtab1} is the average Euclidean distance between the estimated fiducial locations and the ground truth, which is obtained manually using CloudCompare \cite{cloudcompare}, a 3D annotation tool. 
\begin{table}[H]
\caption{Quantitative evaluation of marker localization on 3D maps. }
	\centering
 
	\begin{center}
		\begin{tabular}{c|c|c}
			\hline \hline
				Scene & Localized/Total & Error (m) \\ \hline
    Fig. \ref{mov}(a)--Traj LO \cite{traj} & 2/2 & 0.016 \\ \hline
   Fig. \ref{mov}(a)--Livox Mapping \cite{sdk} & 2/2 & 0.013 \\ 
 \hline
   Fig. \ref{map2}-Traj LO \cite{traj} & 3/3 & 0.015  \\ \hline
   Fig. \ref{map2}-Livox Mapping \cite{sdk} & \textbf{2/3} & 0.020  \\ \hline 
Fig. \ref{mov}(c)--Map 1 & 2/2 & 0.026 \\ \hline
Fig. \ref{mov}(c)--Map 2 & 4/4 & 0.021 \\ \hline
Fig. \ref{mov}(c)--Map 3 & 2/2 & 0.017 \\
\hline
 Fig. \ref{map3} & 4/4 & 0.013
 \\ \hline \hline
			
		\end{tabular}
	\end{center}
		\label{imtab1}
\end{table} \par
\begin{figure}[H] 
	\centering
\includegraphics[width=1.0\linewidth]{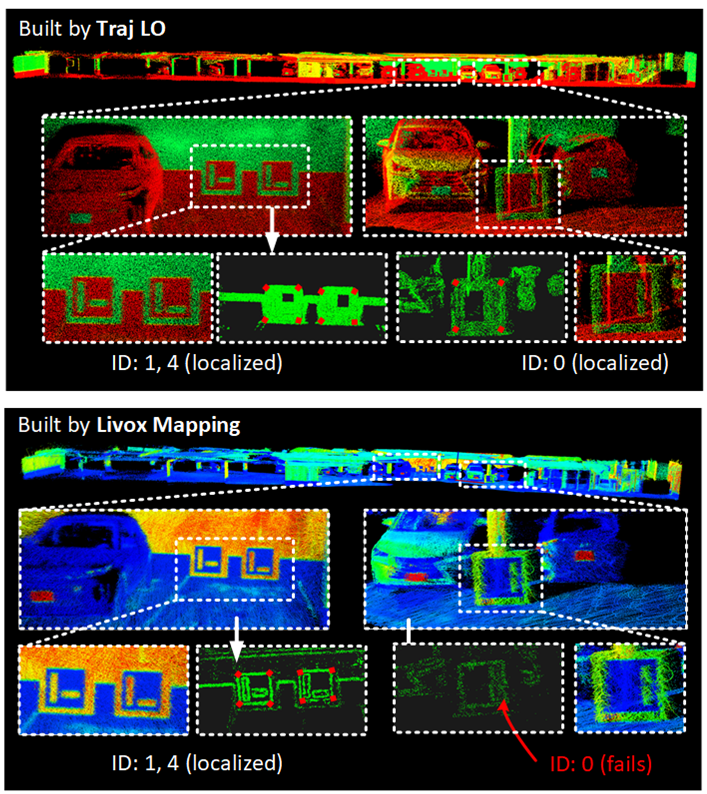}
	\caption{The LiDAR scans the indoor parking lot from right to left, moving rapidly when scanning marker ID 0.}
	\label{map2}
\end{figure}
\begin{figure}[H] 
	\centering
\includegraphics[width=1.0\linewidth]{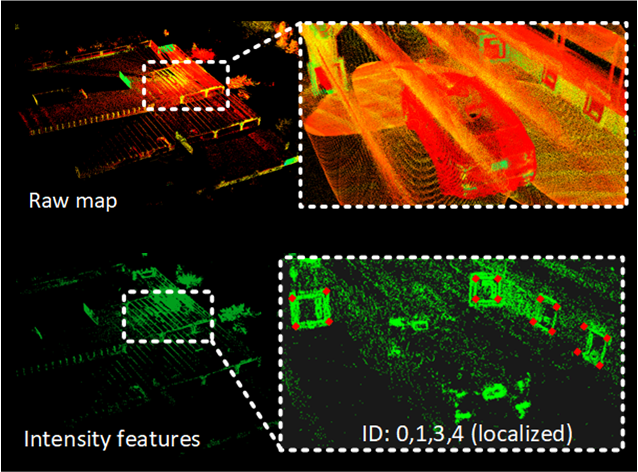}
	\caption{Traj LO \cite{traj} is utilized to create this 3D map.}
	\label{map3}
\end{figure}

The results in Fig. \ref{mov}(a) (outdoor office park, two maps) and Fig. \ref{map2} (indoor small parking lot, two maps) demonstrate the generalizability of the proposed algorithm to maps constructed using various methods \cite{traj,sdk}. 
Each scene uses the same rosbag as input for different SLAM methods, including Traj LO \cite{traj} and Livox Mapping \cite{sdk}, to build 3D maps. As shown, the proposed method successfully localizes all markers in these maps except for one marker in the map built by Livox Mapping in Fig. \ref{map2}. This is because Traj LO, as the state-of-the-art pure LiDAR-based SLAM method, is more robust to rapid motion \cite{traj} than Livox Mapping, and therefore produces maps with less motion blur. For this reason, all the remaining 3D maps in this section are built using Traj LO \cite{traj}. Thereafter, four additional scenes are presented, as shown in Fig. \ref{mov}(c) (maps 1-3) and Fig. \ref{map3}. As seen, all the fiducial markers are successfully localized. 

\subsection{Extending the Valid LFM Localization Range:\\ Addressing the Second Limitation} \label{322}
\begin{figure}[ht] 
	\centering
\includegraphics[width=1.0\linewidth]{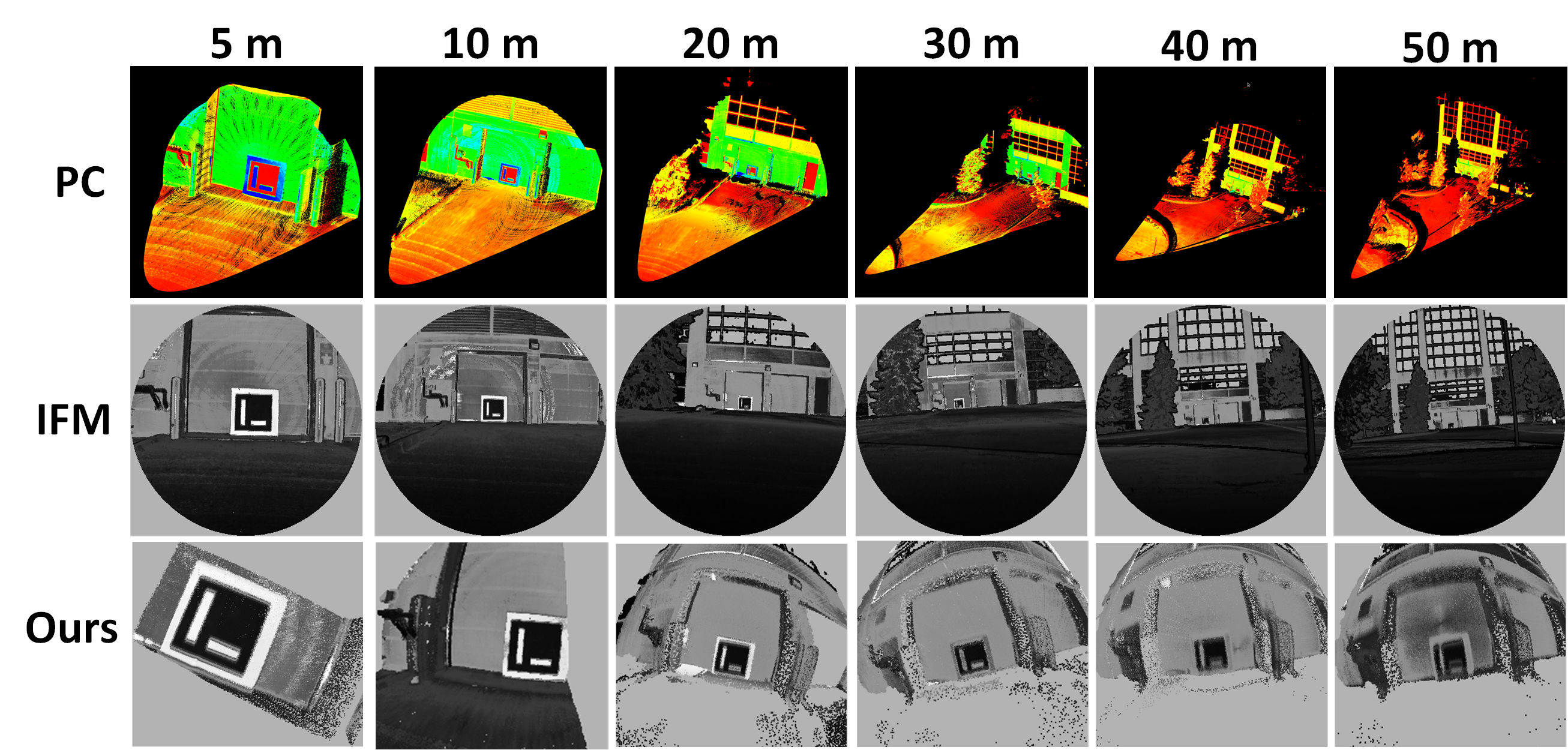}
	\caption{Comparison of the intensity images generated by the vanilla IFM and the proposed method at different distances.}
	\label{imgs}
\end{figure}

The second limitation of the vanilla IFM is that the projection of the marker is too small to be detected when the distance between the marker and LiDAR is far. This section reports experimental results that demonstrate the improvements achieved by the proposed method in this regard. Specifically, as depicted in Fig. \ref{imgs}, we fix the position of a 69.2 cm $\times$ 69.2 cm ArUco marker and move the LiDAR to increase the relative distance. The intensity images obtained by the vanilla IFM and the proposed method at different distances are also presented in Fig. \ref{imgs}. 

The quantitative evaluation results are presented in Table \ref{imtab2}. As seen, the vanilla IFM fails to localize the marker when the distance reaches 20 meters, whereas the proposed method remains valid even when the distance extends to 50 meters. The intensity images shown in Fig. \ref{imgs} illustrate the reason: as the distance increases, the projection of the marker becomes smaller and smaller using the vanilla IFM. In contrast, the proposed method extracts the point cluster belonging to the marker and transforms it onto the intermediate plane, thereby maintaining a high-quality projection where the marker remains detectable. However, it was observed that at a distance of 50 meters, although the marker is localized, the method returns a false ID number. Upon observing the intensity image at 50 meters, we believe this occurs because the captured points become sparser and noisier as the distance increases, leading to poorer quality of the point cloud for the marker. \par

\begin{table}[H]
\caption{Quantitative evaluation of marker localization at various distances. }
	\centering
	\begin{center}
		\begin{tabular}{c|c|c|c|c|c|c}
			\hline \hline
				Method/Distance (m) & 5 & 10 & 20 & 30 & 40 & 50 \\ \hline
Vanilla IFM Error (m) & 0.018 & 0.033 & fails & fails & fails & fails \\ \hline
			Ours Error (m) & 0.009 & 0.016 & 0.019 & 0.023 & 0.031 & 0.037 (false ID)  \\ \hline \hline
		\end{tabular}
  \end{center}
		\label{imtab2}
\end{table}
\subsection{Marker Localization Accuracy Improvement}
In this section, the same experimental setup shown in Fig. \ref{setup} is used to demonstrate that the proposed method boosts pose estimation accuracy compared to the vanilla IFM. In particular, a 16.4 cm $\times$ 16.4 cm AprilTag \cite{ap3} is placed in front of the LiDAR, and the ground truth pose is provided by a motion capture system. In addition to the vanilla IFM, a comparison is also made with the widely-used AprilTag 3 \cite{ap3}. When testing AprilTag 3, the adopted sensor is a camera. As shown in Table \ref{imtab3}, both the accuracy of the vanilla IFM and AprilTag 3 degrades as the distance from the sensor to the marker increases, but the degradation in accuracy is less severe for IFM compared to AprilTag 3. The accuracy of the proposed approach is slightly better than the vanilla IFM at a distance of 2m. However, unlike the vanilla IFM, the proposed approach maintains decent accuracy as the distance increases. The reason is illustrated in Fig. \ref{imgs}: the proposed method generates higher-quality intensity images thanks to cluster extraction and the adoption of the intermediate plane.




\begin{table}[H]
\begin{center}
\caption{Comparison of the proposed approach, the vanilla IFM, and AprilTag3 \cite{ap3} with respect to pose estimation accuracy.}
\begin{tabular}{c|c|c|c|c}
\hline \hline
 Distance& Method & X error (m) & Y error (m) &Z error (m) \\
\cline{1-5} 
\multirow{3}{*}{ 2m} & AprilTag3 \cite{ap3} & 0.016 & 0.034 &0.016 \\ \cline{2-5} 
& Vanilla IFM & 0.003 & 0.006 &0.017 \\ \cline{2-5} 
& \textbf{Ours} &\textbf{ 0.002} &\textbf{ 0.005}  &\textbf{0.011} \\ \cline{1-5}

\multirow{3}{*}{3m} & AprilTag3 \cite{ap3} & 0.058 & 0.124 &0.044 \\ \cline{2-5} 
& Vanilla IFM & 0.026  & 0.021 &0.093 \\ \cline{2-5} 
& \textbf{Ours} &\textbf{ 0.006} &\textbf{ 0.009} &\textbf{0.015} \\ \cline{1-5}

\multirow{3}{*}{ 4m} & AprilTag3 \cite{ap3} & 0.072 & 0.407 &0.233 \\ \cline{2-5} 
& Vanilla IFM & 0.024 & 0.024 &0.107 \\ \cline{2-5} 
& \textbf{Ours} &\textbf{ 0.008} & \textbf{0.014} &\textbf{0.016} \\ \cline{1-5}

Distance& Method & Roll error (deg)&  Pitch error (deg) & Yaw error (deg) \\ \hline
\multirow{3}{*}{ 2m} & AprilTag3 \cite{ap3} & 0.807&  3.166 &  1.244 \\ \cline{2-5} 
& Vanilla IFM & 0.423&  0.457 &  0.399 \\ \cline{2-5} 
& \textbf{Ours} &\textbf{ 0.315}&  \textbf{0.305} & \textbf{ 0.391} \\ \cline{1-5}

\multirow{3}{*}{ 3m} & AprilTag3 \cite{ap3} & 1.369&  7.963 & 2.904 \\ \cline{2-5} 
& Vanilla IFM & 0.930&  1.182 &  0.859 \\ \cline{2-5} 
& \textbf{Ours} &\textbf{ 0.343}& \textbf{ 0.322} & \textbf{ 0.455} \\ \cline{1-5}

\multirow{3}{*}{ 4m} & AprilTag3 \cite{ap3} & 1.433&  9.292 &  13.343 \\ \cline{2-5} 
& Vanilla IFM & 0.342&  2.020 &  1.111 \\ \cline{2-5} 
& \textbf{Ours} &\textbf{ 0.302}&  \textbf{0.389} &\textbf{ 0.478} \\ 
\hline \hline
\end{tabular}
\label{imtab3}

\end{center}
\end{table}

\subsection{Application and Discussion} 
The localized fiducials on the 3D maps are beneficial for downstream tasks, such as map merging. Specifically, as presented in Fig. \ref{mov}(c), multiple large-scale 3D maps built by LiDAR-based SLAM can be merged using the method introduced in Section \ref{multi}, utilizing the fiducials provided by the proposed method on 3D maps.\par
As illustrated by the experiments conducted in Sections \ref{321} and \ref{322}, the intention behind improving the vanilla IFM is to handle large-scale outdoor scenes and cases where the distance between the LiDAR and the marker is significant. Nevertheless, the improvements come at the cost of increased computational time. Thus, unless faced with challenging large-scale scenarios, the vanilla IFM is capable of handling regular small-scale scenes and is preferred for its simplicity and efficiency.

\chapter[Exploiting LFMs for Mapping and Localization]{Exploiting LFMs for Mapping and \\Localization
 \label{multi}}
 \section{Overview} \label{5.1}
\begin{figure}[H] 
	\centering
\includegraphics[width=0.8\linewidth]{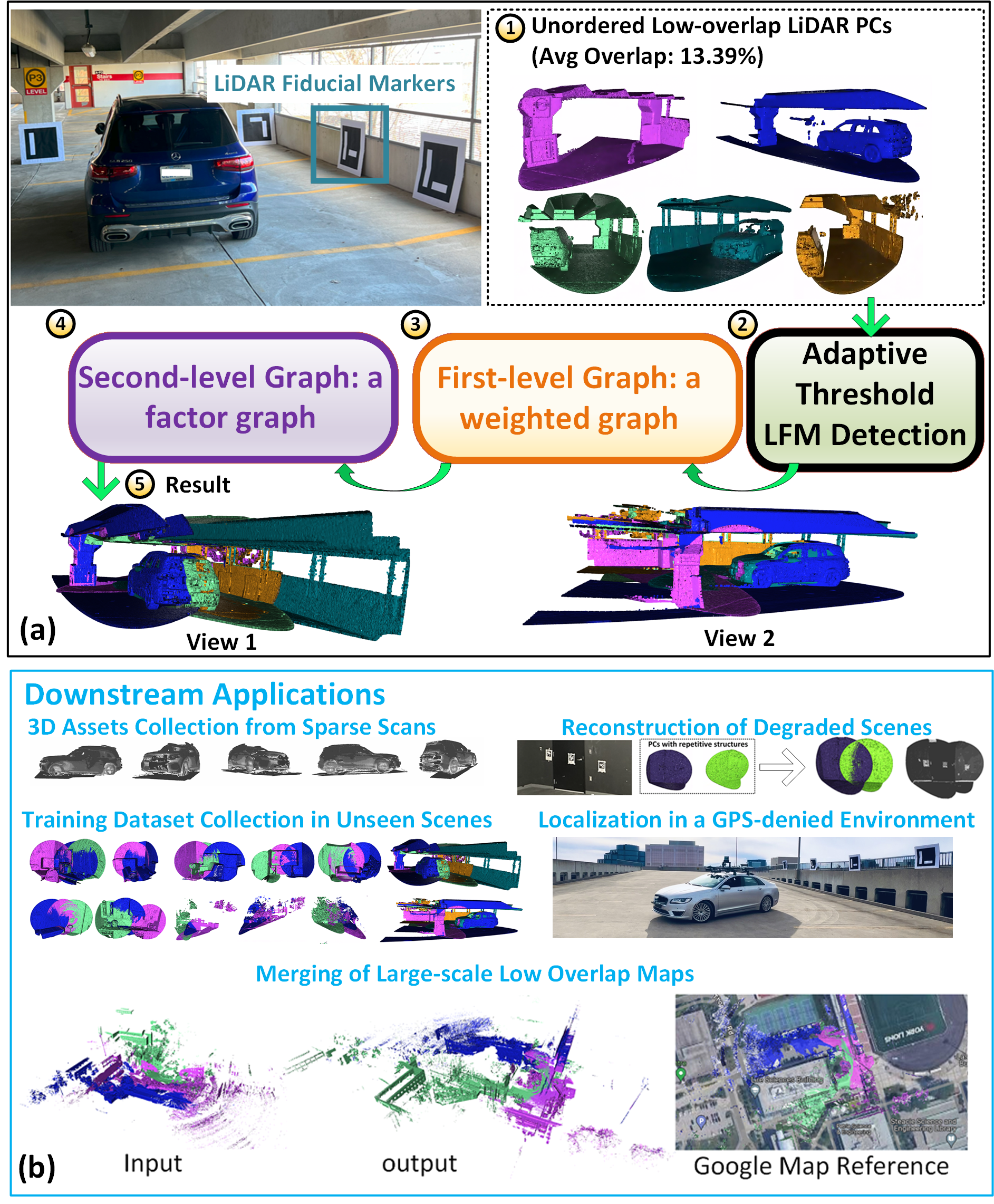}
	\caption{An overview of the proposed framework for mapping and localization using LFMs.}
	\label{mov2}
\end{figure}

This chapter introduces a framework for mapping and localization using LFMs. Specifically, mapping is performed by registering multiview point clouds, while localization is achieved by estimating the relative poses between them. Fig. \ref{mov2} provides an overview of the proposed framework. As shown, the LFMs are thin sheets of board or paper attached to surfaces, with no impact on the environment's geometry, making them suitable for tasks such as training dataset collection for point cloud registration. 
Given an unordered set of low-overlap point clouds, the proposed method efficiently registers them in a global frame. 
Due to the use of a constant threshold for processing intensity images (See Fig. \ref{dis}(c)), the vanilla IFM becomes unstable when the viewpoint varies significantly. To address this issue, an adaptive threshold marker detection method, robust to viewpoint changes, is proposed.
The multiview point cloud registration problem is formulated as a MAP problem, and tackled using two-level graphs.
The first-level graph, constructed as a weighted graph, efficiently processes the unordered point clouds and estimates the initial poses between them. The weights represent the pose estimation error of each marker observation, and the optimal initial poses are obtained by finding the shortest path from an anchor scan to each non-anchor scan.
Using the initial values, the second-level graph, a factor graph, resolves the MAP problem by globally optimizing the poses of point clouds, markers, and marker corner positions. We conduct both qualitative and quantitative experiments to demonstrate the superiority of the proposed method over competitors \cite{mdgd,sghr,se3et,geotransformer,teaser,qingdao,kiss}. Especially, the proposed method is robust to any unseen scenarios with extremely low overlap, making it a convenient, efficient, and low-cost tool for diverse applications that pose significant challenges to existing methods. As shown in Fig. \ref{mov2}, the downstream applications include 3D asset collection from sparse scans, training dataset collection for point cloud registration in unseen scenes, reconstruction of degraded scenes, navigation in GPS-denied environments, and merging of large-scale low overlap 3D maps.
\section{Methodology} \label{new5.2}
\subsection{Adaptive Threshold LFM Detection} \label{5.2}
\begin{figure}[H] 
	\centering
\includegraphics[width=1.0\linewidth]{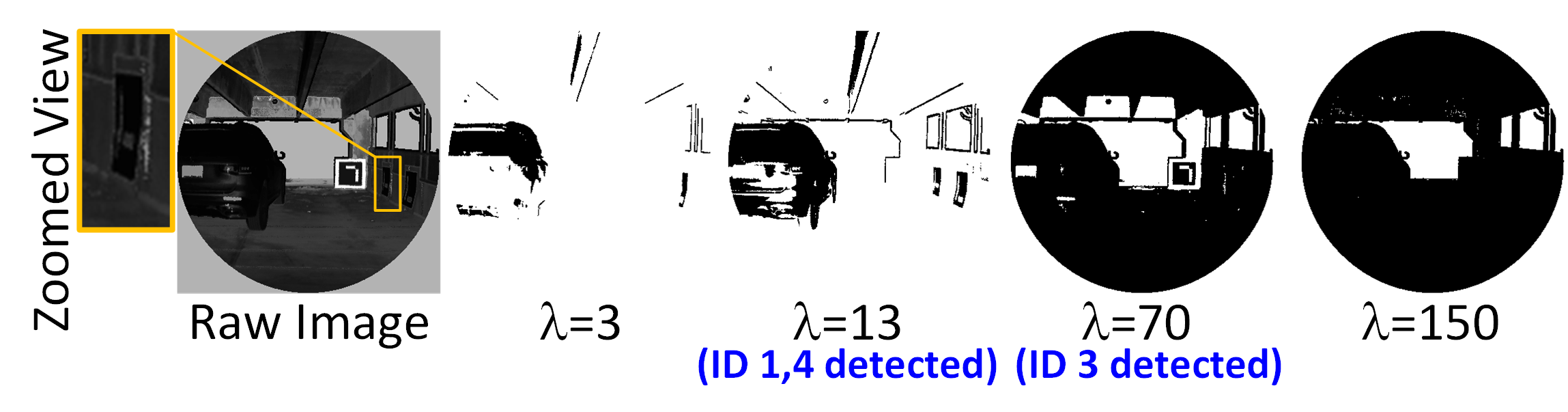}
	\caption{The raw intensity image binarized with different threshold values.}
	\label{threshold}
\end{figure}

In the vanilla IFM method, binarization is applied to the raw intensity image due to the imaging noise (See the zoomed views of Fig. \ref{threshold}). As shown in Fig. \ref{threshold}, the effect is determined by the threshold $\lambda$.
In Section \ref{IFM}, it is found that $\lambda$ can be selected as a constant if the viewpoint does not change drastically. However, with significant changes in the scene, the value of $\lambda$ needs adjustment. 
As an example, consider the case shown in Fig. \ref{threshold}, where three markers are placed. Markers identification (ID) 1 and 4 are detectable when $\lambda$=13, while marker ID 3 is detectable when $\lambda$=70. Thus, $\lambda$=13 and $\lambda$=70 are the optimal thresholds compared to other values for markers ID 1 and 4 and marker ID 3, respectively, denoted by $\lambda^{*}$.
Therefore, unless an appropriate $\lambda$ is carefully selected for each point cloud (viewpoint) and for each marker, the LFM detection will fail.
Unlike the vanilla IFM, which determines $\lambda^{*}$ based on experience or experimentation, an algorithm to automatically seek $\lambda^{*}$ is developed in this section.
\begin{algorithm}[th]
    \KwIn{Raw intensity image, $\mathcal{I}$}
    \KwOut{The optimal threshold, $\lambda^{*}$.}
    
    \textbf{Initialize parameters:} Search scope, $S$. step size, $\delta$. 
    queue for saving detected markers, $Q=[ \ ]$. length of $Q$, $Q_l=0$.
    temporary queue, $Q_{temp}=[\ ]$. length of  $Q_{temp}$,  $Q_{temp,l}=0$. the optimal threshold, $\lambda^{*}=0$. search step, $i=0$.\\
    Define the binarization operation as $\Psi(\mathcal{I},\lambda)$. \\ Define the marker detector operation as $\Gamma(\mathcal{I})$. \\
    \While{$i < S$}{
    $\lambda = \delta\times i$\\
        $\mathcal{I} = \Psi(\mathcal{I},\lambda)$\\
         $\Gamma(\mathcal{I}) \rightarrow Q_{temp}$\\
        \If{$Q_{temp,l} \geq Q_{l}$}{
        \For{marker \textbf{in} $Q_{temp}$}{
        \If{marker \textbf{not in} $Q$}{append \textit{marker} to $Q$}
        }
        $\lambda^{*}=\lambda$}
        $i = i+1$\\     
    }
    Return the image binarized with $\lambda^{*}$ and $Q$.
    \caption{ Search for the optimal threshold, $\lambda^{*}$. } \label{algo1}
\end{algorithm}
Given that the focus is on marker detection rather than image denoising, \textbf{Algorithm \ref{algo1}} is designed to utilize the detection result as feedback for the automatic search of $\lambda^{*}$.
In particular, the core of the algorithm is to maximize the length of a memory queue for saving detected markers (denoted by $Q$) by gradually increasing $\lambda$.
Namely, the aim is to detect as many markers as possible by finding the optimal threshold for each marker in the scene.
After applying \textbf{Algorithm \ref{algo1}}, the 2D fiducials located on the image binarized with $\lambda^{*}$ are projected into 3D fiducials using the subsequent steps of IFM. The proposed adaptive threshold LFM detection algorithm addresses the problem of a constant threshold being inapplicable to all point clouds and markers, especially when LiDAR moves in the wild and scenes undergo significant changes.
\subsection{Problem Formulation} \label{5.3}
In this section, the problem formulation is introduced. Suppose that the marker size is $a$. In the marker coordinate system $\{M\}_{j}$, the 3D coordinates of the four corners of the $j$-th marker, $^{j}\mathbf{p}^{j,1},^{j}\mathbf{p}^{j,2},^{j}\mathbf{p}^{j,3},$ and $^{j}\mathbf{p}^{j,4}$, are $[-a/2,-a/2,0]^{T}, [a/2,-a/2,0]^{T}, [a/2,a/2,0]^{T},$ and $[-a/2,a/2,0]^{T}$, respectively. LFM detection returns the 3D coordinates of the corners expressed in $\{F\}_{i}$ (the local coordinate system of the $i$-th scan $f_{i}$).
Thus, the 6-DOF transmission from $\{M\}_{j}$ to $\{F\}_{i}$, denoted as $\mathbf{T}^{j}_{i}$, can be found by solving the following least square problem:
 \begin{equation}	
	\mathbf{T}^{j,*}_{i}=\underset{\mathbf{T}^{j}_{i}}{\arg \min } \sum_{s=1}^{4}\left\| \mathbf{T}^{j}_{i} \ \cdot \ ^{j}\mathbf{p}^{j,s}-_{i}\mathbf{p}^{j,s}\right\|^{2}.\label{least3}
\end{equation}

The SVD method \cite{barfoot} is employed to compute $\mathbf{T}^{j,*}_{i}$. The set of measurements, including 
\begin{enumerate}[label=(\roman*)]
	 \item the corner positions w.r.t. $\{M\}_{j}$,
	\item the corner positions w.r.t. $\{F\}_{i}$,
	\item the marker poses w.r.t. $\{F\}_{i}$,
 
\end{enumerate}
are denoted as $\mathcal{Z}$. 
To register the multiview point clouds, it is necessary to find a globally consistent pose for each scan so that the point clouds can be transformed into a complete point cloud in the world coordinate system $\{G\}$. 
To this end, the following variables are considered: 
\begin{enumerate}[label=(\roman*)]
	 \item the poses of point clouds w.r.t. $\{G\}$,
	\item the poses of markers w.r.t. $\{G\}$,
	\item the marker corner positions w.r.t. $\{G\}$.
\end{enumerate}
\par
The set of variables is represented by $\Theta$. Finally, a MAP inference problem is formulated: given the measurements, $\mathcal{Z}$, the goal is to find the optimal variable set $\Theta^{*}$ that maximizes the posterior probability $P(\Theta \mid \mathcal{Z})$:
\begin{equation}
\
	\Theta^*=\underset{\Theta}{\arg \max } \ P(\Theta \mid \mathcal{Z}).\label{map}
\end{equation} 
\par
In the following, a framework consisting of two-level graphs is designed to solve this MAP problem. Inspired by the coarse-to-fine pipeline of SGHR \cite{sghr}, a first-level graph is developed to efficiently and exhaustively determine the relative poses among point clouds, while a second-level graph is used to globally optimize the variables.

\subsection{First-level Graph} \label{1level}
As aforementioned, the variables in $\Theta$ to be optimized cannot be directly obtained through the measurements in $\mathcal{Z}$. Moreover, deriving the initial values of the variables requires both efficiency and accuracy, given that the input is an unordered set of low overlap point clouds. 
The first-level graph is designed to address this challenge. First, the computation of the relative pose between two point clouds is considered. Suppose the $j$-th marker appears in the scenes of two scans $f_{i}$ and $f_{m}$, $\mathbf{T}^{j}_{i}$ and $\mathbf{T}^{j}_{m}$ can be calculated using Eq. (\ref{least3}). Consequently, the relative pose between $f_{i}$ and $f_{m}$, denoted as $\mathbf{T}_{i,m}$, is available from
 \begin{equation}	
	\mathbf{T}_{i,m}=(\mathbf{T}^{j}_{i})^{-1} \mathbf{T}^{j}_{m},\label{relative}
\end{equation}
where $(\mathbf{T}^{j}_{i})^{-1}$ indicates the inverse of $\mathbf{T}^{j}_{i}$. $(\mathbf{T}^{j}_{i})^{-1}$ indicates the pose that transforms 3D points from $\{F\}_{i}$ to $\{M\}_{j}$. Although the method introduced in Eq. (\ref{relative}) is straightforward, it cannot be directly applied because the input in this work is an unordered set of point clouds that do not follow a temporal or spatial sequence.
Specifically, for scans (point clouds) with no shared marker observations, their relative pose has to be calculated through pose propagation among other scans.
\par
However, multiple alternative paths could exist. Even for scans that share marker observations, there could be more than one overlapped marker. 
\par
\begin{figure}[H] 
	\centering
\includegraphics[width=0.75\linewidth]{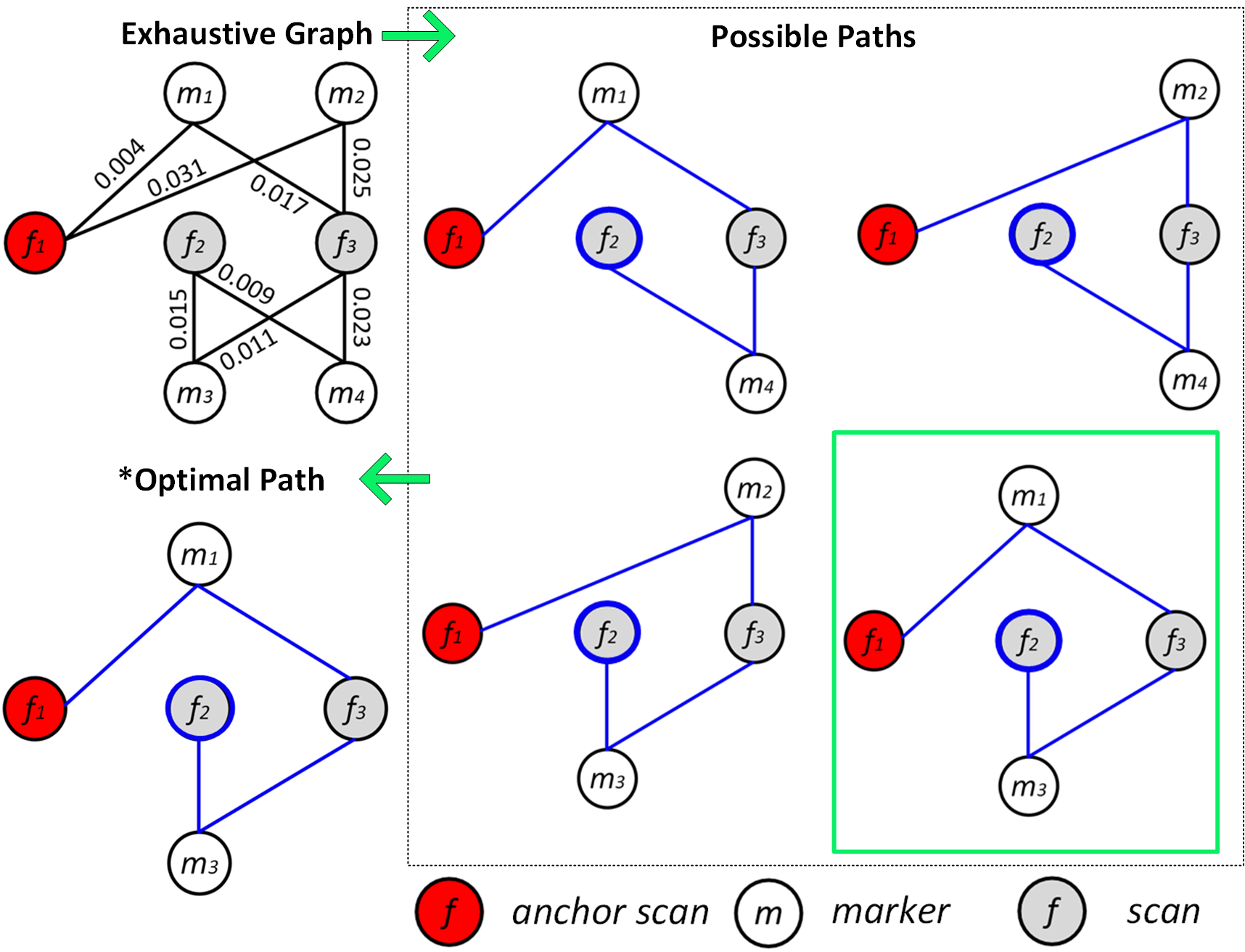}
	\caption{An illustration of the first-level graph.
    After applying the proposed adaptive marker detection to all scans, an exhaustive weighted graph is constructed, with the scans and markers as nodes and the point-to-point errors as edge weights.
    The aim is to derive the relative pose between each non-anchor scan and the anchor scan with optimal accuracy.
    However, for a given non-anchor scan, such as $f_{2}$ in this simple case, there may be multiple possible paths in the exhaustive graph leading to the anchor scan ($f_{1}$).
    Thus, Dijkstra’s algorithm \cite{dij} is employed to find the optimal path with the minimum accumulation of point-to-point errors (weights).}
	\label{firstlevel}
\end{figure}

Hence, to accurately and efficiently estimate relative poses among scans, it is necessary to design an algorithm to determine which scans and markers to apply Eq. (\ref{relative}). Specifically, the objective is to infer the relative poses with the highest possible accuracy using only the necessary low-dimensional information through a simple process. 
Thus, as shown in Fig. \ref{firstlevel}, the first-level graph is constructed in the form of a weighted graph. 

In particular, when processing the point clouds with the proposed adaptive threshold method one by one, if a marker is detected in a point cloud, the corresponding scan node and marker node are added to the graph along with a weighted edge. The edge weight is the pose estimation point-to-point error ${e}_{pp}$:
\begin{equation}	
	{e}_{pp}=\sum_{s=1}^{4}\left\| \mathbf{T}^{j,*}_{i} \ \cdot \ ^{j}\mathbf{p}^{j,s}-_{i}\mathbf{p}^{j,s}\right\|^{2}.\label{epp}
\end{equation}
\par
Eq. (\ref{epp}) indicates the substitution of the result given by SVD back into the right side of Eq. (\ref{least3}). ${e}_{pp} \in \mathbb{R}^{+}$ is employed as the metric to evaluate the quality of pose estimation of the marker in the corresponding point cloud. The first scan is defined as the anchor scan. The local coordinate system of the anchor scan is set as the world coordinate system. 
Namely, $\{F\}_{1}=\{G\}$. Only the relative poses between the anchor scan and each non-anchor scan are considered. Although there could be multiple paths from the given scan to the anchor scan (see Fig. \ref{first}), the information on pose estimation quality has been saved by constructing the first-level graph as a weighted graph. Therefore, Dijkstra’s algorithm \cite{dij} is employed to obtain the shortest path. Then, the relative pose is computed along the shortest path iteratively using Eq. (\ref{relative}). Given that the relative pose computation is achieved with the lowest accumulation of $e_{pp}$, the estimation result achieves the highest possible accuracy. Moreover, the search for the optimal path to propagate poses is based merely on the one-dimensional $e_{pp}$, without incorporating any 6-DOF poses or 3D locations.
\par
In this way, the point cloud poses w.r.t. $\{G\}$ are obtained. Since the marker detection provides the marker poses and the 3D positions of marker corners w.r.t. the local coordinate system of corresponding scan, the initial values of the marker poses w.r.t. $\{G\}$ and the corner positions w.r.t. $\{G\}$ can be derived through the point cloud poses w.r.t. $\{G\}$.

\subsection{Second-level Graph} \label{5.5}
To address the MAP problem introduced in Eq. (\ref{map}), we construct the second-level graph as a factor graph to globally optimize the variables using the initial values obtained from the first-level graph. Specifically, we create the second-level graph in three stages.\par
\begin{figure}[H]
	\centering
	\includegraphics[width=5.5in]{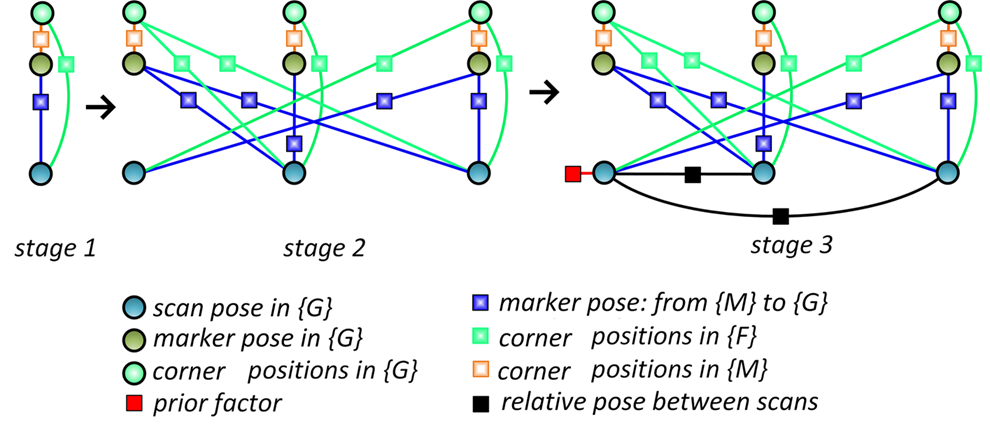}

	\caption{The procedures for formulating the second-level graph. The variable nodes are represented by circles and the factor nodes are represented by squares. In Stage One, when a marker is detected in a scan, six types of nodes are added to the graph, including (1) scan pose in $\{G\}$, (2) marker pose in   $\{G\}$, (3) corner positions in $\{G\}$, (4) marker pose from $\{M\}$ to $\{G\}$, (5) corner positions in $\{F\}$, and (6) corner positions in $\{M\}$, along with their corresponding edges. In Stage Two, all the marker detection results are traversed, and the operation from Stage One is repeated for each detected marker.  In Stage Three, a prior factor connecting the anchor scan is added, along with factors representing the relative poses between the anchor scan and non-anchor scans.} \label{secondgraph}

\end{figure}

\noindent\textbf{Stage One.} Suppose that the $j$-th marker is detected in the $i$-th scan, the following six types of nodes are added to the second graph. The variable nodes include: 
\begin{itemize}
    \item [(1)] Node $\mathbf{T}^{j}$, which refers to the 6-DOF pose of the $j$-th marker \textit{w.r.t.} $\{G\}$;
    \item [(2)] Node $\mathbf{T}_{i}$, which refers to the 6-DOF pose of the $i$-th scan \textit{w.r.t.} $\{G\}$;
    \item [(3)] Nodes $\{ \mathbf{p}^{j,1}, \mathbf{p}^{j,2}, \mathbf{p}^{j,3}, \mathbf{p}^{j,4} \}$, which refer to 3D coordinates of the corners of the $j$-th marker \textit{w.r.t.} $\{G\}$. 
\end{itemize}
\noindent The factor nodes include: 
\begin{itemize}
    \item [(4)] Node $\mathbf{T}^{j}_{i}$, which refers to the measurement of the 6-DOF pose of the $j$-th marker \textit{w.r.t.} $\{F\}_{i}$;
    \item [(5)] Nodes $\{ ^{j}\mathbf{p}^{j,1}, \ ^{j}\mathbf{p}^{j,2}, \ ^{j}\mathbf{p}^{j,3}, \ ^{j}\mathbf{p}^{j,4} \}$, which refer to the measurement of the 3D coordinates of the corners of the $j$-th marker \textit{w.r.t.} $\{M\}_{j}$;

    \item [(6)] Nodes $\{ _{i}\mathbf{p}^{j,1}, \ _{i}\mathbf{p}^{j,2},$\ $ _{i}\mathbf{p}^{j,3},$ $_{i}\mathbf{p}^{j,4} \}$, which refer to 3D coordinates of the corners of the $j$-th marker \textit{w.r.t.} $\{F\}_{i}$.
\end{itemize}
By adding variables representing the 3D coordinates of the corners, we can also optimize the fiducial localization results. The first stage in Fig. \ref{secondgraph} shows the added nodes and edges when a marker is detected in a scan. \par
\noindent\textbf{Stage Two.} Thereafter, we traverse all the marker detection results and conduct the operation from Stage One for each detected marker. 
After processing all the detected markers, the second-level graph becomes the one shown in Stage Two of Fig. \ref{secondgraph}. Since the operation in this stage essentially consists of repetitions of Stage One, the node types are no different from those in the previous stage. \par
\noindent\textbf{Stage Three.} Considering that the local coordinate system of the first scan, $\{F\}_{1}$, is set as the global coordinate system, $\{G\}$, we add a prior factor that connects to the first (anchor) scan node,  $\mathbf{T}^{1}$. Finally, we add factor nodes representing the relative poses between the anchor scan and each non-anchor scan. Up to this point, the factor graph is completed, as shown in Stage Three of Fig. \ref{secondgraph}. \par
Following \cite{isam2,gtsam}, since the factor graph is determined, the posterior probability $P(\Theta \mid \mathcal{Z})$ is factorized as:
\begin{equation}
P(\Theta \mid \mathcal{Z})=\prod_k P^{(k)}(\Theta),
\label{obj}
\end{equation}
where $P^{(k)}$ are the factors in the second graph. We follow the standard pipeline \cite{isam2} to model them as Gaussians:
\begin{equation}
 P^{(k)}(\Theta) \propto \exp \left(-\frac{1}{2}\left\|h_k\left(\Theta\right)\ominus z_k\right\|_{\Sigma_k}^2\right),
\end{equation}
where $h_k\left(\Theta \right)$ is the measurement function and $z_k$ is a measurement.
$\|e\|_{\Sigma}^2 \triangleq e^T \Sigma^{-1} e$ denotes the squared Mahalanobis distance with $\Sigma$ being the covariance matrix. 
Following \cite{tagslam}, if $z_k\in\mathbb{R}^{3\times1}$ is a 3D position, $\ominus$ refers to straight subtraction for elements. 
If $z_k\in SE(3)$ is a 6-DOF pose, $\ominus$ generates 6-dimensional Lie algebra (refer to Section \ref{lie}) coordinates:
\begin{equation}{\resizebox{0.6\hsize}{!}{$\mathbf{T}_{A} \ominus \mathbf{T}_{B} = [[\log(\mathrm{Rot}(\mathbf{T}_{B}^{-1}\mathbf{T}_{A}))]^{T}_{\vee}, \mathrm{Trans}(\mathbf{T}_{B}^{-1}\mathbf{T}_{A})^{T} ]^{T},
$} } \end{equation} 
where for a $\mathbf{T} \in SE(3)$, $\mathrm{Rot}(\mathbf{T})$ denotes the rotation matrix $\mathbf{R} \in SO(3)$ and $\mathrm{Trans}(\mathbf{T})$ denotes translation vector $\mathbf{t \in \mathbb{R}^{3\times1}}$. 
$\log(\cdot)$ represents the matrix logarithm. $\vee$ is the \textit{vee} map operator. The detailed introduction of $\vee$ is provided in Section \ref{lie}.  With the Gaussian modeling, the objective function in Eq. (\ref{obj}) is transformed into a least square problem by applying the negative logarithm:
\begin{equation}
\resizebox{0.7\hsize}{!}{$\arg \min _{\Theta}(-\log \prod_k P^{(k)}(\Theta))= 
\arg \min _{\Theta} \frac{1}{2} \sum_k\left\|h_k\left(\Theta\right)\ominus z_k\right\|_{\Sigma_k}^2.$}
\end{equation} \par

We utilize the Levenberg-Marquardt algorithm \cite{gtsam} to solve this problem. The acquisition of initial values is introduced in Section \ref{1level}. The noise covariance matrices are determined by the quantitative experiments conducted in \cite{iilfm}.
\section{Livox-3DMatch Dataset} \label{dataset}
A common approach \cite{sghr,pre} to evaluating a learning-based point cloud registration model is to train it on 3DMatch \cite{3dmatch} and test it on various benchmarks, including 3DMatch \cite{3dmatch}, ETH \cite{eth}, and ScanNet \cite{scan}. However, the 3DMatch benchmark is mainly constructed from RGB-D camera captures of indoor scenes \cite{3dmatch}. In this dissertation, a new dataset, named Livox-3DMatch, is collected to enrich the training data for learning-based methods. The enrichment contains two key components. 
\par
\begin{figure}[H] 
	\centering
\includegraphics[width=1.0\linewidth]{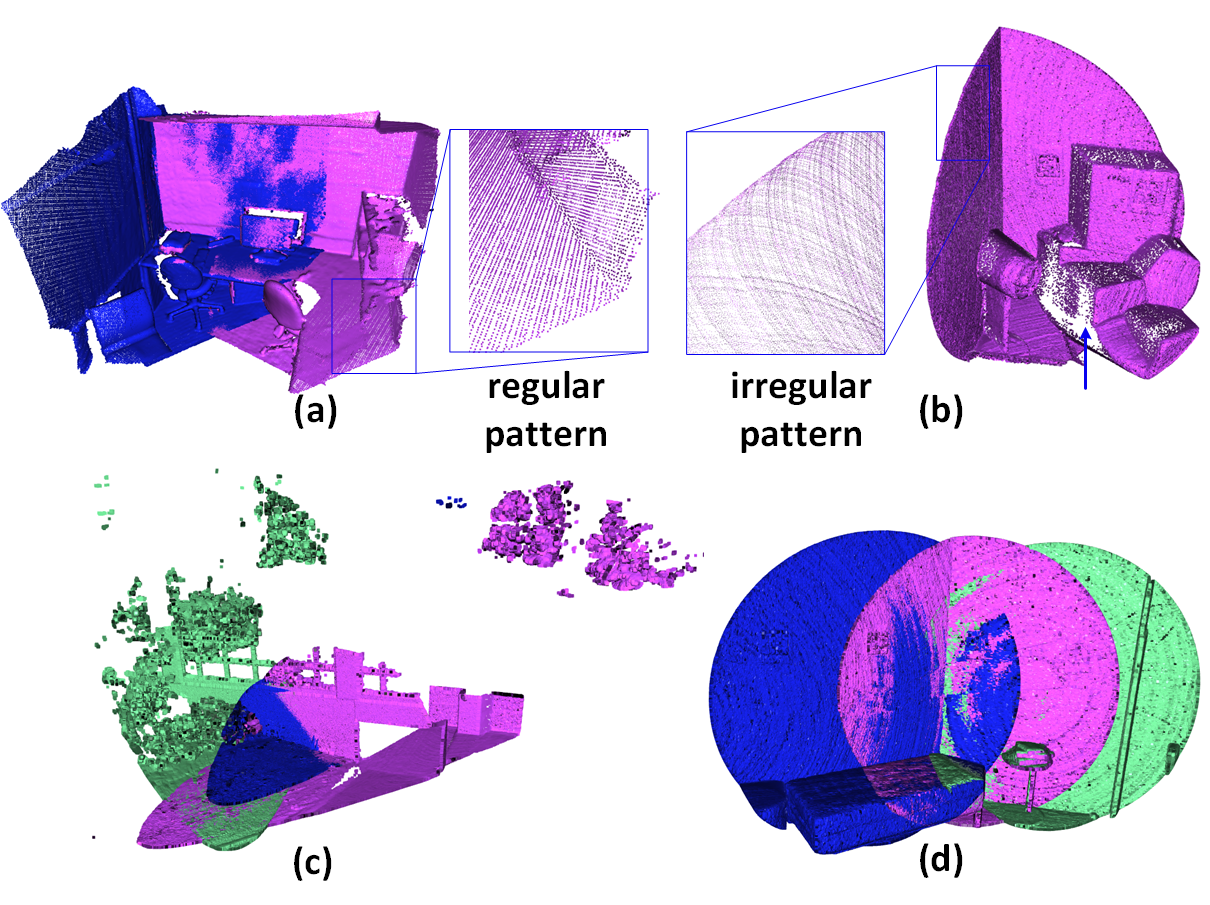}
	\caption{Comparison of 3DMatch and Livox-3DMatch. (a): A random sample from 3DMatch. The point cloud sampled by an RGB-D camera has a regular pattern and less noise. (b): A random sample from Livox-3DMatch. The Livox LiDAR point cloud has an irregular pattern and more noise. (c): An example of an outdoor scene in Livox-3DMatch. (d): A selectively sampled challenging case in which the overlapping regions are mainly planes.}
	\label{livox3d}
\end{figure}

First, the sensor we adopt is a Livox MID-40 LiDAR, which has different sampling patterns compared to the RGB-D camera (See Fig. \ref{livox3d}(a) and (b)). Thus, it adds point clouds with new features to the training data. 
\par
Second, scenes that are absent or rare in 3DMatch are selectively sampled, thereby enriching the scenes in the training data. For example, the valid depth range of an RGB-D camera is usually less than ten meters, making it unsuitable for sampling outdoor scenes. In contrast, some outdoor scenes (See Fig. \ref{livox3d}(c)) are collected for Livox-3DMatch, considering that a LiDAR can sample objects a few hundred meters away. Moreover, some challenging cases (See Fig. \ref{livox3d}(d)) are gathered where the overlapping regions are mainly planes, which are rare in 3DMatch \cite{3dmatch}. A more detailed introduction to the scenes in Livox-3DMatch is given in Sections \ref{test1} and \ref{test2}. In Section \ref{testadd}, we demonstrate that the proposed Livox-3DMatch can boost the performance of the state-of-the-art learning-based methods \cite{mdgd,sghr} on various benchmarks \cite{3dmatch,eth,scan}.

\section{Experimental Validation} \label{5.7}

\subsection{Experimental Setup}
In this section, both the solid-state LiDAR (Livox Mid-40) and mechanical LiDAR (RS-Ruby-128) are employed to validate the proposed framework. In particular, the point clouds are sampled from various indoor and outdoor scenes, including offices, a meeting room, lounges, a kitchen, office buildings, and thickets. In indoor scenes, letter-sized AprilTags are utilized as LFMs. In outdoor scenes, poster-sized ArUcos are employed to provide LiDAR fiducials. For experiments concerning registration accuracy, the ground truth poses are obtained by manually registering through CloudCompare \cite{cloudcompare}, a popular 3D annotation tool. The ground truth for evaluating the quality of 3D asset collection is a high-fidelity 3D model acquired from a 3D assets website. For experiments involving localization accuracy, the ground truth trajectory is provided by the MoCap system for the indoor scene and by Real-Time Kinematic (RTK) for the outdoor scene.

\subsection{Evaluation of the Adaptive Marker Detection Method} \label{test0}
\begin{figure}[ht] 
	\centering
\includegraphics[width=1.0\linewidth]{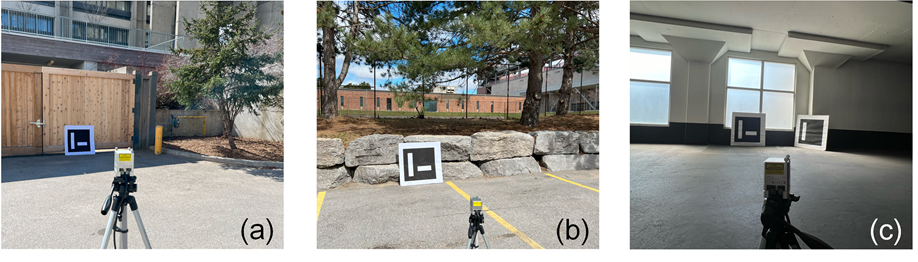}
	\caption{Setup for testing the adaptive threshold marker detection algorithm: (a) between groups of buildings, (b) in an open outdoor area, and (c) in a large indoor parking lot.}
	\label{threshold2}
\end{figure}

\noindent\textbf{Data.}
The Livox MID-40 LiDAR and 69.2 cm $\times$ 69.2 cm ArUco marker(s) are placed in three different scenes, as shown in Fig. \ref{threshold2}. The three scenes are: between groups of buildings, in open outdoor areas, and in large indoor parking lots. In each scene, we collect point clouds and test the proposed adaptive threshold method at different relative positions.
 \par
\noindent\textbf{Results and Analysis.}
Table \ref{tabthreshold} presents the results, where x and y indicate the relative position of the LiDAR in the marker (ID 4) coordinate system. Specifically, in scene Fig. \ref{threshold2}(c), there are two markers (ID 4 and ID 1), so the column for $\lambda^{*}$ contains two values. As seen in the table, $\lambda^{*}$ varies significantly as the relative position changes, demonstrating the necessity of the proposed adaptive marker detection method. In addition, the results shown in Fig. \ref{threshold2}(c) also illustrate the necessity of the memory queue design, considering that a single $\lambda^{*}$ might not be applicable to all markers in the same scene. Please note that the intention of reporting the values of the optimal threshold in Table \ref{tabthreshold} is to demonstrate that a constant threshold, as adopted in \cite{iilfm}, is not applicable to this task. However, the major concern of the adaptive threshold marker detection algorithm is to detect all markers in a scene. Namely, the marker detection results saved in the memory queue are the most important output rather than the thresholds.

\begin{table}[H]
\caption{Demonstration of the necessity of the proposed adaptive marker detection algorithm}
\begin{center}

\resizebox{1.0\columnwidth}{!}{

\begin{tabular}{c|c|c|c | c|c|c|c| c|c|c|c}
\hline\hline
Scene & x (m) & y (m) & $\lambda^{*}$ &Scene & x (m) & y (m) & $\lambda^{*}$ &Scene & x (m) & y (m) & $\lambda^{*}$ \\ \hline
 \multirow{21}{*}{\shortstack{Fig. \ref{threshold2}(a)\\Between\\groups of\\buildings}} &0 &4   & 36  & \multirow{21}{*}{\shortstack{Fig. \ref{threshold2}(b)\\In an open\\outdoor\\area}} &0 &4   & 7 & \multirow{21}{*}{\shortstack{Fig. \ref{threshold2}(c)\\In a large\\indoor\\parking lot}} &4 &4   & 20/6   \\  \cline{2-4} \cline{6-8} \cline{10-12} 
 &4 &4  & 9 &  &4 &4  & 8 &  &4 &4  & 8/8 \\  \cline{2-4} \cline{6-8} \cline{10-12} 
&-4 &4   & 6 & &-4 &4  & 6 &  &-4 &4  & 6/8   \\  \cline{2-4} \cline{6-8} \cline{10-12} 
 &0 &5  & 20 &  &0 &5  & 9 &  &0 &5  & 15/10 \\  \cline{2-4} \cline{6-8} \cline{10-12} 
  &5 &5 & 8 & &5 &5  & 8&  &5 &5  & 8/12\\  \cline{2-4} \cline{6-8} \cline{10-12} 
   &-5 &5  & 7&  &-5 &5  & 9 &  &-5 &5  & 7/13 \\  \cline{2-4} \cline{6-8} \cline{10-12} 
    &0 &6  & 19 &  &0 &6  & 9 &  &0 &6  & 8/13\\  \cline{2-4} \cline{6-8} \cline{10-12} 
  &6 &6 & 10& &6 &6  & 6 &  &6 &6  & 11/11\\  \cline{2-4} \cline{6-8} \cline{10-12} 
   &-6 &6  & 11 & &-6 &6  & 8 &  &-6 &6  & 11/10 \\  \cline{2-4} \cline{6-8} \cline{10-12} 
     &0 &7  & 17 &&0 &7  & 9 &  &0 &7  & 19/20 \\  \cline{2-4} \cline{6-8} \cline{10-12} 
  &7 &7 & 8 &&7 &7  & 8 &  &7 &7  & 15/14\\  \cline{2-4} \cline{6-8} \cline{10-12} 
   &-7 &7  & 9  &&-7 &7  & 9&  &-7 &7  & 28/16  \\  \cline{2-4} \cline{6-8} \cline{10-12} 
   &0 &8  & 15  &&0 &8  & 13&  &0 &8  & 10/16 \\  \cline{2-4} \cline{6-8} \cline{10-12} 
  &8 &8 & 9 &&8 &8  & 11 &  &8 &8  & 11/17\\  \cline{2-4} \cline{6-8} \cline{10-12} 
   &-8 &8  & 10  &&-8 &8  & 15&  &-8 &8  & 12/15 \\  \cline{2-4} \cline{6-8} \cline{10-12} 
    &0 &9  & 17  &&0 &9  & 10 &  &0 &9  & 13/11\\  \cline{2-4} \cline{6-8} \cline{10-12} 
  &9 &9 & 8 &&9 &9  & 12 &  &9 &9  & 20/26\\  \cline{2-4} \cline{6-8} \cline{10-12} 
   &-9 &9  & 9  &&-9 &9  & 12  &  &-9 &9  & 16/12\\ \cline{2-4} \cline{6-8} \cline{10-12} 
   &0 &10  & 14  &&0 &10  & 13&  &0 &10  & 24/28\\  \cline{2-4} \cline{6-8} \cline{10-12} 
  &10 &10 & 11 &&10 &10  &10 &  &10 &10  & 10/19\\  \cline{2-4} \cline{6-8} \cline{10-12} 
   &-10 &10  & 12   &&-10 &10  & 15 &  &-10 &10  & 15/19 \\ \cline{2-4} \cline{6-8} \cline{10-12} 
 \hline\hline
\end{tabular}

}
\label{tabthreshold}
\end{center}
\end{table}

\subsection{Evaluation of Point Cloud Registration Accuracy} \label{test1}
\begin{figure}[H] 
	\centering
\includegraphics[width=1.0\linewidth]{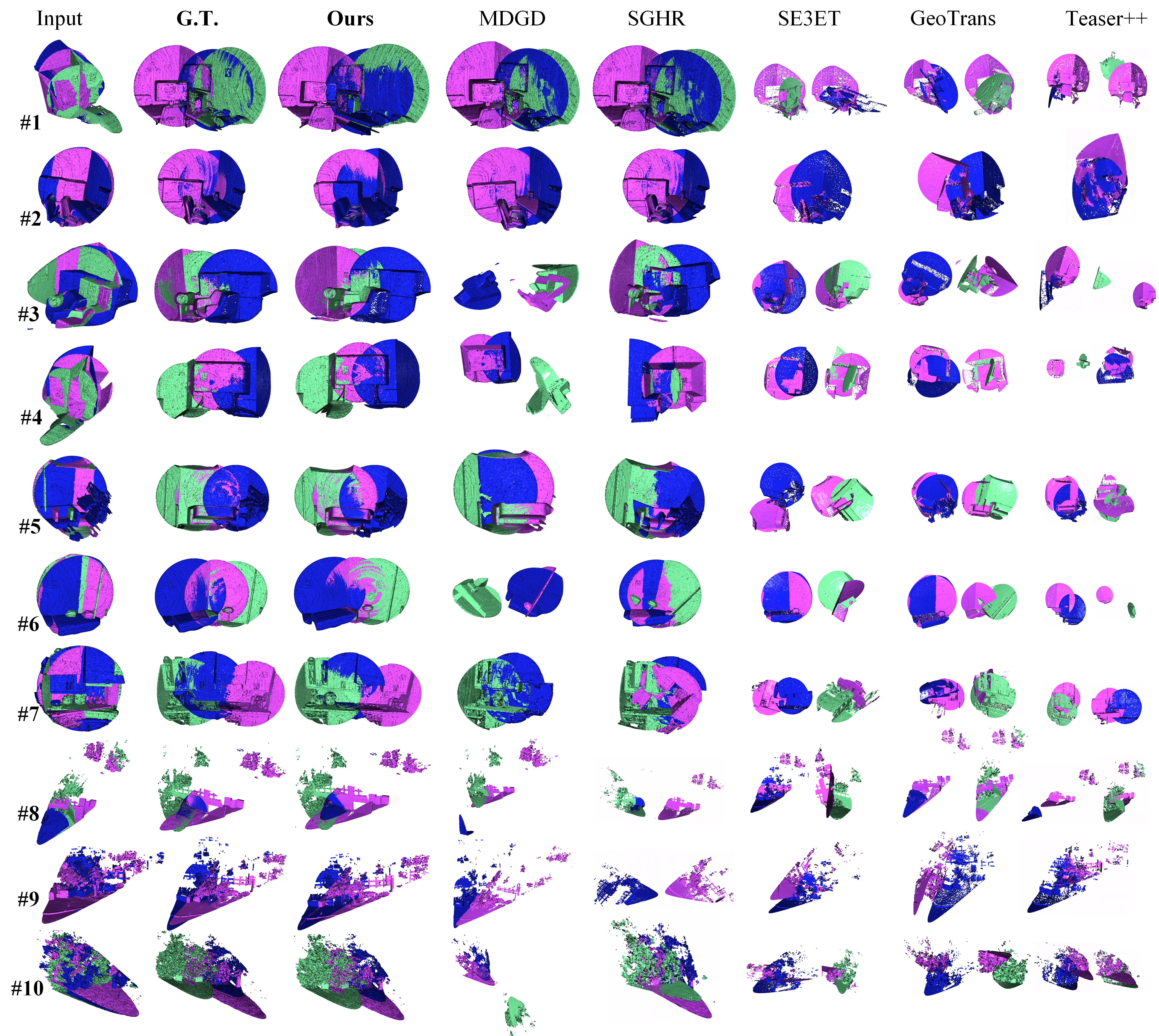}
	\caption{A comparison with SOTA methods, including MDGD \cite{mdgd}, SGHR \cite{sghr}, SE3ET \cite{se3et}, GeoTrans \cite{geotransformer}, and Teaser++ \cite{teaser}, on ten scenes. The scenes include the office (1-3), the meeting room (4), the lounge (5,6), the kitchen (7), the office building (8,9), and the thicket (10). Each scene consists of three scans, except scenes 2 and 9, which are composed of two scans.}
	\label{newbench}
\end{figure}

\noindent\textbf{Data.} Given that the existing point cloud registration benchmarks \cite{3dmatch,eth,scan} lack fiducial markers in the scenes, a new dataset is constructed with the Livox MID-40, as shown in Fig. \ref{newbench}. As illustrated in the caption of Fig. \ref{newbench}, the newly collected dataset covers various indoor and outdoor scenes. Indoors, multiple 16.4 cm × 16.4 cm AprilTags \cite{ap3} are positioned in the environment. Outdoors, multiple 69.2 cm × 69.2 cm ArUcos \cite{aruco} are placed in the environment. Note that since the LFMs in this work are thin sheets of objects attached to other surfaces, they are almost invisible in the point clouds. This is infeasible if LiDARTags \cite{lt} or calibration boards \cite{cal,cal2,a4} are adopted to provide fiducials. These scenes are challenging due to low overlap, and the overlap rate \cite{pre} of each scene is also presented in Table \ref{tabnew}. The ground truth poses between scans are obtained manually using CloudCompare \cite{cloudcompare}. Note that manually aligning point clouds is a labor-intensive and time-consuming process, highlighting the benefits of developing an automatic tool like the proposed framework. The inference time of the methods is compared using an AMD Ryzen 7 5800X CPU.\\
\noindent\textbf{Competitors and Metrics.}
Competitors include the latest state-of-the-art (SOTA) multiview point cloud registration methods, MDGD \cite{mdgd} (RA-L$^{\prime}$24), SGHR \cite{sghr} (CVPR$^{\prime}$23), and the SOTA pairwise methods, SE3ET \cite{se3et} (RA-L$^{\prime}$24), GeoTrans \cite{geotransformer} (TPAMI$^{\prime}$23), and Teaser++ \cite{teaser} (T-RO$^{\prime}$20). All the competitors are learning-based methods, except for Teaser++, which is a geometry-based method. For all pairwise methods, we manually select pairs of point clouds that have overlap as the inputs.
The root-mean-square errors (RMSEs) \cite{shuo} are employed as the metric: 

\begin{equation}
\begin{aligned}
&\operatorname{RMSE}_{T}=\sqrt{\sum_{n=0}^{N_s} e_{T, n}^2 /(N_s+1)}, 
\\
& \operatorname{RMSE}_{R}=\sqrt{\sum_{n=0}^{N_s} e_{R, n}^2 /(N_s+1)},
\end{aligned} \label{rmse}
\end{equation}
where $e_{T, n}^2$ and $e_{R, n}^2$ represent the squared Euclidean distances between the ground truth and the estimates of translation and rotation, respectively. $N_s$ denotes the number of samples. We also compare the inference time of the methods with an AMD Ryzen 7 5800X CPU.
\par
\noindent\textbf{Results and Analysis.}
The qualitative and quantitative results are presented in Fig. \ref{newbench} and Table \ref{tabnew}, respectively. 
The pairwise methods \cite{se3et,geotransformer,teaser}, although provided manually selected pairs of point clouds with overlap, struggle in all scenes.
This is because they are tailored for extracting geometric features either in a learning-based manner \cite{se3et,geotransformer} or through conventional geometry \cite{teaser}. Thus, when the overlap ratio between point clouds is low and the overlapped regions lack sufficient geometric features, these methods struggle to find features and utilize them to align point clouds. 
In contrast, the multiview point cloud registration methods \cite{mdgd,sghr} show some successful cases. In particular, SGHR \cite{sghr} successfully registers the point clouds in scenes 1 to 3 and one point cloud pair in scene 4. MDGD \cite{mdgd}, built upon the framework of SGHR \cite{sghr} but developing a new matching distance-based overlap estimation module, aligns scenes 1 and 2 with decent accuracy and also successfully registers one pair of point clouds in scenes 4, 5, 7, and 10. The reason is that, in addition to learning pairwise registration, the multiview point cloud registration methods \cite{mdgd,sghr} also learn to further optimize or refine the pose graph constructed using pairwise registration results.
Despite this, they fail in other scenes. Reviewing their success cases, it is found that these indoor scenes are similar to those in their training dataset \cite{3dmatch}. While failure cases, such as scenes 4 to 10, are rare or lacking in the training dataset. This indicates that their generalization to unseen scenarios is limited due to the out-of-distribution problem. The proposed method achieves the best performance. Specifically, our method does not require the overlapping regions to have explicit geometric features, thanks to the thin-sheet format of our LiDAR fiducial markers. Moreover, by virtue of the proposed adaptive marker detection algorithm, our method can robustly align any novel indoor or outdoor scenes with thin-sheet markers.
The proposed method also yields the best efficiency as it focuses on registering point clouds through LiDAR fiducial markers rather than analyzing the entire point clouds. The process of our method takes several dozen seconds, while other methods need several minutes on the AMD Ryzen 7 5800X CPU.
\begin{table*}[ht]
\caption{Quantitative comparison with MDGD \cite{mdgd}, SGHR \cite{sghr}, SE3ET \cite{se3et}, GeoTrans \cite{geotransformer}, and Teaser++ \cite{teaser} on our dataset.}
\centering
	\resizebox{1.0\columnwidth}{!}{
\begin{tabular}{c|c|c|c|c|c|c}
\hline\hline

\multicolumn{2}{c|}{Scene $\#$ } & 1 (Office)  &2 (Office) & 3 (Office)  & 4 (Meeting Room) &5 (Lounge)   \\ 

\multicolumn{2}{c|}{Avg/Min Overlap Rate (\%)} & 27.17/14.29  &46.88/46.88 &44.66/22.26 & 25.80/1.40 &43.82/15.48  \\ 
\hline
\multirow{3}{*}{Teaser++ \cite{teaser}} &$\mathrm{RMSE}_{R}$ (rad) & 1.504  & 1.795 & 2.394 & 2.277 & 2.493 \\
&$\mathrm{RMSE}_{T}$ (m) & 1.724 & 1.890 & 2.075 & 2.252& 1.911 \\
(T-RO$^{\prime}$20)& Time (s) & 313.1 & 298.3 & 336.4 & 322.3& 388.9  \\
\hline
\multirow{3}{*}{GeoTrans \cite{geotransformer}} &$\mathrm{RMSE}_{R}$ (rad) & 1.331  & 1.451 & 1.935 & 1.632 & 1.532 \\
&$\mathrm{RMSE}_{T}$ (m) & 1.502 & 1.671 & 1.912 & 1.552& 1.325 \\
(TPAMI$^{\prime}$23)& Time (s) & 91.8 & 87.3 & 95.1 & 93.9& 105.3  \\
\hline
\multirow{3}{*}{SE3ET \cite{se3et}} &$\mathrm{RMSE}_{R}$ (rad) & 0.925  & 0.834 & 1.776 & 1.358 & 1.381 \\
&$\mathrm{RMSE}_{T}$ (m) & 0.996 & 1.113 & 1.736 & 1.273& 1.207 \\
(RA-L$^{\prime}$24)& Time (s) & 116.8 & 94.6 & 99.5 & 95.7& 115.7 \\
\hline
\multirow{3}{*}{SGHR \cite{sghr}} & $\mathrm{RMSE}_{R}$ (rad)  & 0.152  & \textbf{0.060}  & 1.198 & 1.825 & 1.311   \\
& $\mathrm{RMSE}_{T}$ (m) & 0.020 & 0.010 & 0.094 & 0.156& 0.231 \\
(CVPR$^{\prime}$23)& Time (s) & 846.1 & 785.0 & 886.5 & 825.3& 851.7 \\
\hline
\multirow{3}{*}{MDGD \cite{mdgd}} & $\mathrm{RMSE}_{R}$ (rad)  & \textbf{0.032}  & 0.062  & 1.457 & 0.933 & 0.679   \\
& $\mathrm{RMSE}_{T}$ (m) & \textbf{0.015} & \textbf{0.009} & 0.352 & 1.035& 0.113  \\
(RA-L$^{\prime}$24)& Time (s) & 626.1 & 573.3 & 645.5 & 619.2& 612.0 \\
\hline
\multirow{3}{*}{\textbf{Ours}} & $\mathrm{RMSE}_{R}$ (rad) & 0.036 & 0.068 & \textbf{0.089} & \textbf{0.065}& \textbf{0.088} \\
& $\mathrm{RMSE}_{T}$ (m) &   0.017 &  0.011 &  \textbf{0.028} &  \textbf{0.031}& \textbf{ 0.032}  \\
        
& Time (s) & \textbf{31.1}   & \textbf{21.2}  & \textbf{35.7}   & \textbf{32.6 }& \textbf{36.2}  \\
\hline
\multicolumn{2}{c|}{Scene $\#$ } & 6 (Lounge)&  7 (Kitchen)& 8 (Building)&  9 (Building) & 10 (Thicket) \\ 

\multicolumn{2}{c|}{Avg/Min Overlap Rate (\%)} & 50.66/27.24&  31.31/4.19 & 19.06/12.20 &  44.65/44.65 & 12.16/5.42 \\ 
\hline
\multirow{3}{*}{Teaser++ \cite{teaser}} &$\mathrm{RMSE}_{R}$ (rad) &  1.946& 1.632&  1.818 & 1.779 & 1.876 \\
&$\mathrm{RMSE}_{T}$ (m) & 1.733& 2.089&  1.861 & 1.848 & 1.891 \\
(T-RO$^{\prime}$20)& Time (s) & 419.8& 441.5&  397.1 & 327.2 & 401.2 \\
\hline
\multirow{3}{*}{GeoTrans \cite{geotransformer}} &$\mathrm{RMSE}_{R}$ (rad) & 1.471& 1.198&  1.355 & 1.377 & 1.526 \\
&$\mathrm{RMSE}_{T}$ (m) &1.235& 1.132&  1.730 & 1.554 & 1.531 \\
(TPAMI$^{\prime}$23)& Time (s) & 122.3& 127.3&  110.5 & 91.1 & 121.4 \\
\hline
\multirow{3}{*}{SE3ET \cite{se3et}} &$\mathrm{RMSE}_{R}$ (rad) & 1.352& 1.077&  1.156 & 1.232 & 1.376 \\
&$\mathrm{RMSE}_{T}$ (m) & 1.130& 1.095&  1.469 & 1.413 & 1.392 \\
(RA-L$^{\prime}$24)& Time (s) & 136.0& 139.3&  123.5 & 97.3 & 132.0 \\
\hline
\multirow{3}{*}{SGHR \cite{sghr}} & $\mathrm{RMSE}_{R}$ (rad)  &  1.696& 3.42 &  2.792 & 3.034 &2.810 \\
& $\mathrm{RMSE}_{T}$ (m) & 0.274&  0.949&  2.171 & 0.686 & 1.787 \\
(CVPR$^{\prime}$23)& Time (s) & 979.3& 983.6&  909.9 & 794.6 &950.4 \\
\hline
\multirow{3}{*}{MDGD \cite{mdgd}} & $\mathrm{RMSE}_{R}$ (rad)  & 1.715& 0.856 &  0.722 & 0.651 &0.779 \\
& $\mathrm{RMSE}_{T}$ (m) & 0.422&  0.337&  1.553 & 0.686 & 0.939 \\
(RA-L$^{\prime}$24)& Time (s) & 726.4 & 730.5&  689.6 & 590.6 &717.4 \\
\hline
\multirow{3}{*}{\textbf{Ours}} & $\mathrm{RMSE}_{R}$ (rad) & \textbf{0.078}& \textbf{0.067}&  \textbf{0.069} & \textbf{0.087} & \textbf{0.101} \\
& $\mathrm{RMSE}_{T}$ (m) & \textbf{0.043}&  \textbf{0.019}&  \textbf{0.082} &  \textbf{0.069} & \textbf{0.077} \\
        
& Time (s) & \textbf{43.5}& \textbf{53.8} &  \textbf{39.9} & \textbf{24.9}  &\textbf{42.2} \\

\hline\hline
\end{tabular}
}
\label{tabnew}

\end{table*} 
\par
\clearpage
\subsection{Application 1: 3D Asset Collection from Sparse Scans} \label{test2}
Collecting the complete shape of a novel object \cite{mending} from sparse observations is advantageous, as sparse scans require less labor and offer better efficiency. However, it is also a challenging task due to the low overlap between scans. In this test, we evaluate the instance reconstruction quality of the proposed method. \par
\begin{figure}[H] 
	\centering
\includegraphics[width=0.8\linewidth]{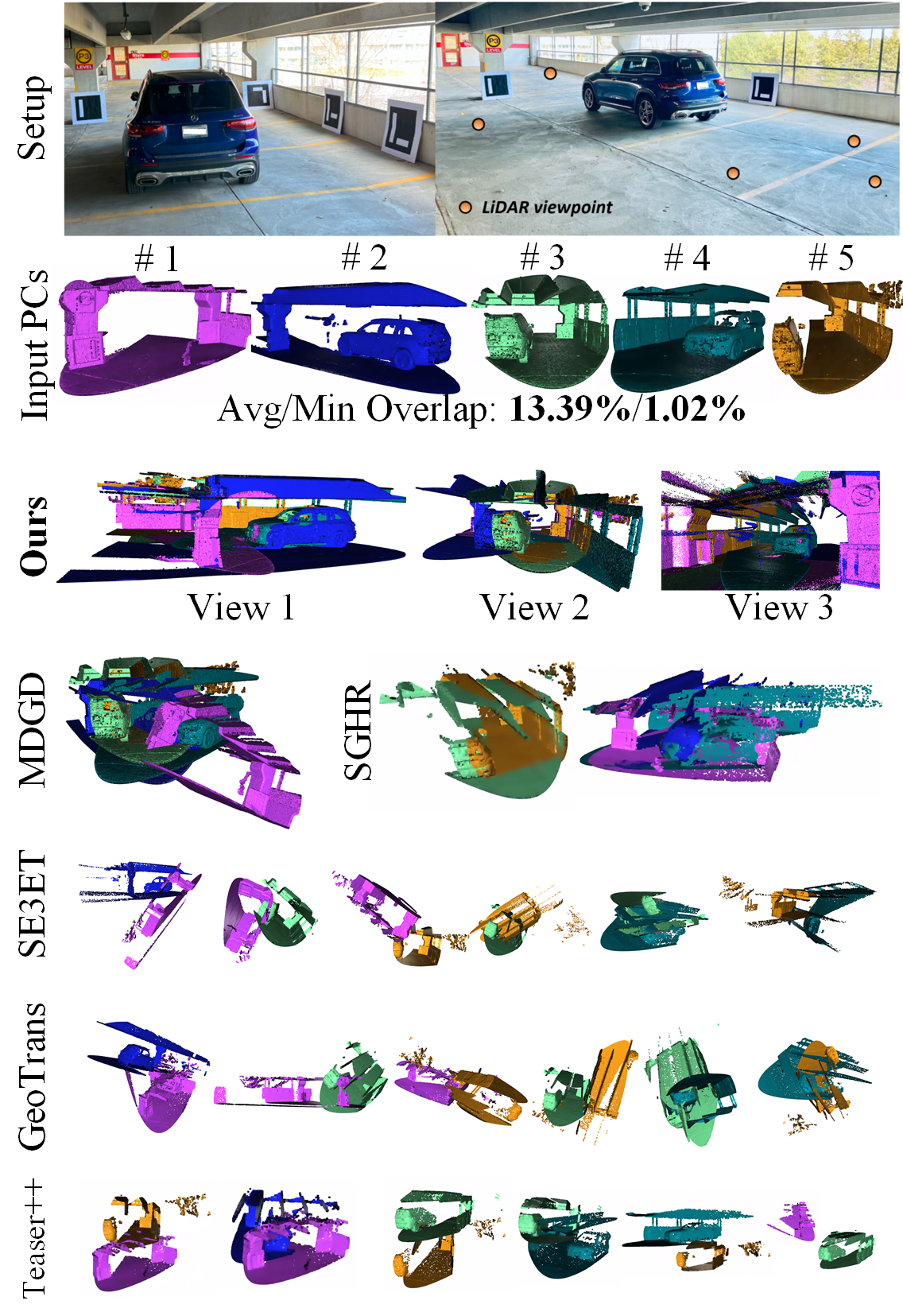}
	\caption{An illustration of the experimental setup and a visual comparison of the proposed method against the SOTA methods (MDGD \cite{mdgd}, SGHR \cite{sghr}, SE3ET \cite{se3et}, GeoTrans \cite{geotransformer}, and Teaser++ \cite{teaser}) regarding instance reconstruction from sparse scans.}
	\label{glbtraj}
\end{figure}
\begin{figure}[H] 
	\centering
\includegraphics[width=1.0\linewidth]{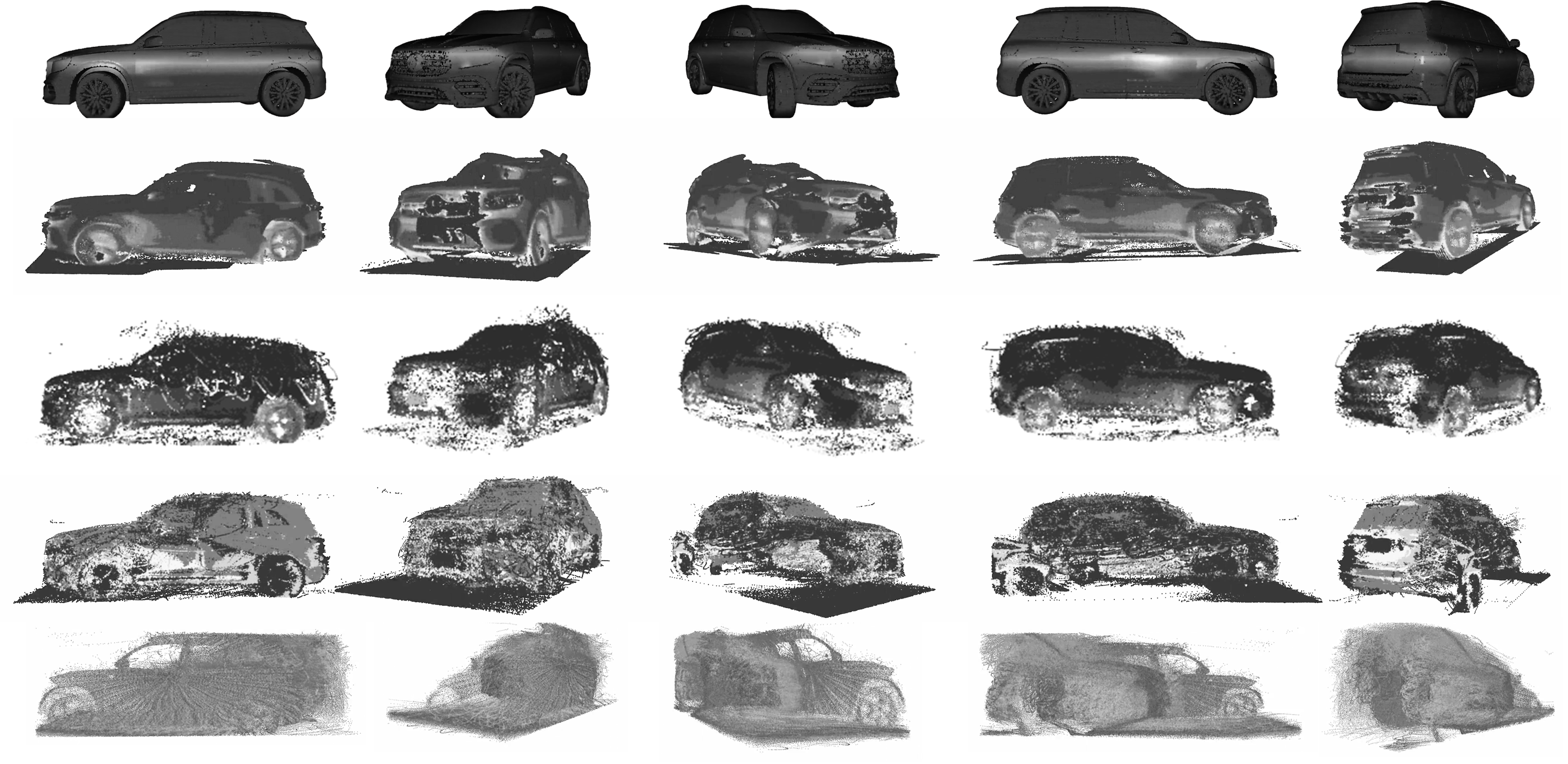}
	\caption{Visual comparison of the instance reconstruction. From top to bottom: ground truth, Ours, Traj LO \cite{traj}, Livox Mapping \cite{sdk}, and LOAM Livox \cite{lloam}.}
	\label{glb}
\end{figure}

\noindent\textbf{Data.}
The experimental setup is depicted in Fig. \ref{glbtraj}. Four 69.2 cm $\times$ 69.2 cm ArUco \cite{aruco} markers are placed in the environment. 
%
As depicted by the yellow trajectory, the Livox MID-40 LiDAR follows a looping path to scan the vehicle, pausing at five viewpoints for a few seconds to collect relatively dense point clouds due to its sparse scanning pattern \cite{lloam}. 
The rostopic of the point cloud published by the LiDAR sensor is recorded as a rosbag throughout the sampling process. Then, the same rosbag is provided to all competitors. 
However, point cloud registration methods \cite{mdgd,sghr,se3et,geotransformer,teaser}, including ours, utilize only a portion of the rosbag, \textit{i.e.}, five point clouds with significant viewpoint changes, as shown in Fig. \ref{glbtraj}. This is also a challenging low-overlap case, with the average and minimum overlap rates being 13.39\% and 1.02\%, respectively. To evaluate the instance reconstruction quality, \textit{Supervisely} \cite{super}, a popular 3D annotation tool, is used to extract the vehicle's point cloud from the reconstruction result.
Since the manufacture and model (Mercedes-Benz GLB 250) of the vehicle to be reconstructed are known, a high-fidelity 3D model acquired from a 3D assets website acts as the ground truth shape.
\\
\noindent\textbf{Competitors and Metrics.}
Competitors include the state-of-the-art point cloud registration methods (MDGD \cite{mdgd}, SGHR \cite{sghr}, SE3ET \cite{se3et}, GeoTrans \cite{geotransformer}, and Teaser++ \cite{teaser}) and LOAM methods (Traj LO \cite{traj}, Livox Mapping \cite{sdk}, and LOAM Livox \cite{lloam}). Pairs of point clouds with overlapping regions are manually provided to the pairwise methods \cite{se3et,geotransformer,teaser}. MDGD \cite{mdgd}, SGHR \cite{sghr}, and our method directly process the set of point clouds. The LOAM methods take the entire rosbag as input. Following \cite{cd}, the Chamfer Distance (CD) and Recall are computed between the ground truth shape and the reconstruction result. In particular, CD is defined as 
\begin{equation}
CD\left(\boldsymbol{X}, \boldsymbol{Y}\right)=\sum_{x \in \boldsymbol{X}} \min _{y \in \boldsymbol{Y}}\|x-y\|_2^2+\sum_{y \in \boldsymbol{Y}} \min _{x \in \boldsymbol{X}}\|x-y\|_2^2. \label{cd}
\end{equation} 
where $\boldsymbol{X}$ and $\boldsymbol{Y}$ are two point sets with $x$ and $y$ being the 3D points belonging to them, respectively. Recall is defined as follows:
\begin{equation}
\operatorname{Recall}(\boldsymbol{X}, \boldsymbol{Y})=\frac{1}{|\boldsymbol{X}|} \sum_{x \in \boldsymbol{X}}\left[\min _{y \in \boldsymbol{Y}}\|x-y\|_2^2<=thr\right],
\end{equation}
where $thr$ is a predefined threshold \cite{cd}.
\\
\noindent\textbf{Results and Analysis.}
The qualitative results for point cloud registration methods and LOAM methods are shown in Fig. \ref{glbtraj} and Fig. \ref{glb}, respectively. The quantitative results are presented in Table \ref{tabglb}. As shown in Fig. \ref{glbtraj}, MDGD \cite{mdgd} and SGHR \cite{sghr} successfully align the third and fifth scans. In particular, the third and fifth scans have the highest overlap ratio of 59.33\% among all pairs as they are captured from similar perspectives. However, when dealing with point clouds that have lower overlap ratios, the competitors \cite{mdgd,sghr,se3et,geotransformer,teaser} struggle to register them. This comparison illustrates that, although these existing methods can handle some unseen scenarios with high overlap, they are not generalizable to novel low overlap cases.
Moreover, the failure of these existing methods in the instance reconstruction task implies that they are not suitable for efficient 3D asset collection. By contrast, the proposed method successfully registers the unordered multiview point clouds. 
\par
In terms of instance reconstruction quality, as depicted in Fig. \ref{glb}, our reconstructed shape preserves intricate details well compared to the ground truth shape. The Traj LO \cite{traj} result shows almost no drift, though the surface is noisy. The Livox Mapping \cite{sdk} result exhibits noticeable drift and surface noise, while LOAM Livox \cite{lloam} demonstrates severe drift. Moreover, these LOAM methods require the entire rosbag instead of just five sparse scans. As shown in Table \ref{tabglb}, the proposed method achieves the highest reconstruction quality. Specifically, the CD and Recall of the reconstructed result are \textbf{0.003} and \textbf{96.22\%}, respectively. 
The decent performance in terms of reconstruction quality metrics \cite{mending,cd} indicates that our method can serve as a convenient, efficient, and low-cost tool for collecting high-fidelity 3D assets using the LiDAR sensor. 
In particular, we consider the proposed method efficient, as it only requires four or five scans from dramatically changed viewpoints to reconstruct the complete shape.

\par

\begin{table}[ht]
\caption{Comparison with LOAM methods regarding reconstruction }
	\centering
		\begin{tabular}{c|c|c}
			\hline\hline
				Method $\backslash$ Metric & CD $\downarrow$ & Recall ($\%$) $\uparrow$ \\ \hline
   Livox Mapping \cite{sdk} & 0.0106 & 75.27 \\ \hline
   LOAM Livox \cite{lloam} & 0.0335 & 78.82  \\ \hline
   Traj LO \cite{traj} & 0.0107 & 82.88  \\ \hline
   \textbf{Ours}  & \textbf{0.0030} & \textbf{96.22}
\\ \hline  \hline	
		\end{tabular}
		\label{tabglb}
\end{table}

\subsection{Application 2: Training Data Collection and Enhancement of\\ Existing Learning-Based Methods} \label{testadd}
Training data is crucial for learning-based methods. Most existing methods \cite{sghr,mdgd} are trained on the 3DMatch dataset \cite{3dmatch}. Augmenting the training data is beneficial for improving the generalization ability of learning-based methods. However, the existing methods cannot be utilized for collecting training data in unseen scenes, as unseen scenes imply out-of-distribution cases and the limited effectiveness of the existing methods.
The proposed method can serve as a valuable tool for collecting training data. 
In this test, we demonstrate that training data collected using our method can enhance the performance of the state-of-the-art methods across various benchmarks.\\
\noindent\textbf{Data.} Using the proposed method, all the point clouds shown in Figs. \ref{newbench} and \ref{glbtraj}, comprising 11 scenes with 33 scans, are aligned and processed into the format required for training by MDGD \cite{mdgd} and SGHR \cite{sghr}. The newly collected data is named Livox-3DMatch. We train the models \cite{mdgd,sghr} from scratch using only 3DMatch and a combination of 3DMatch and Livox-3DMatch.
\begin{figure}[H] 
	\centering
\includegraphics[width=1.0\linewidth]{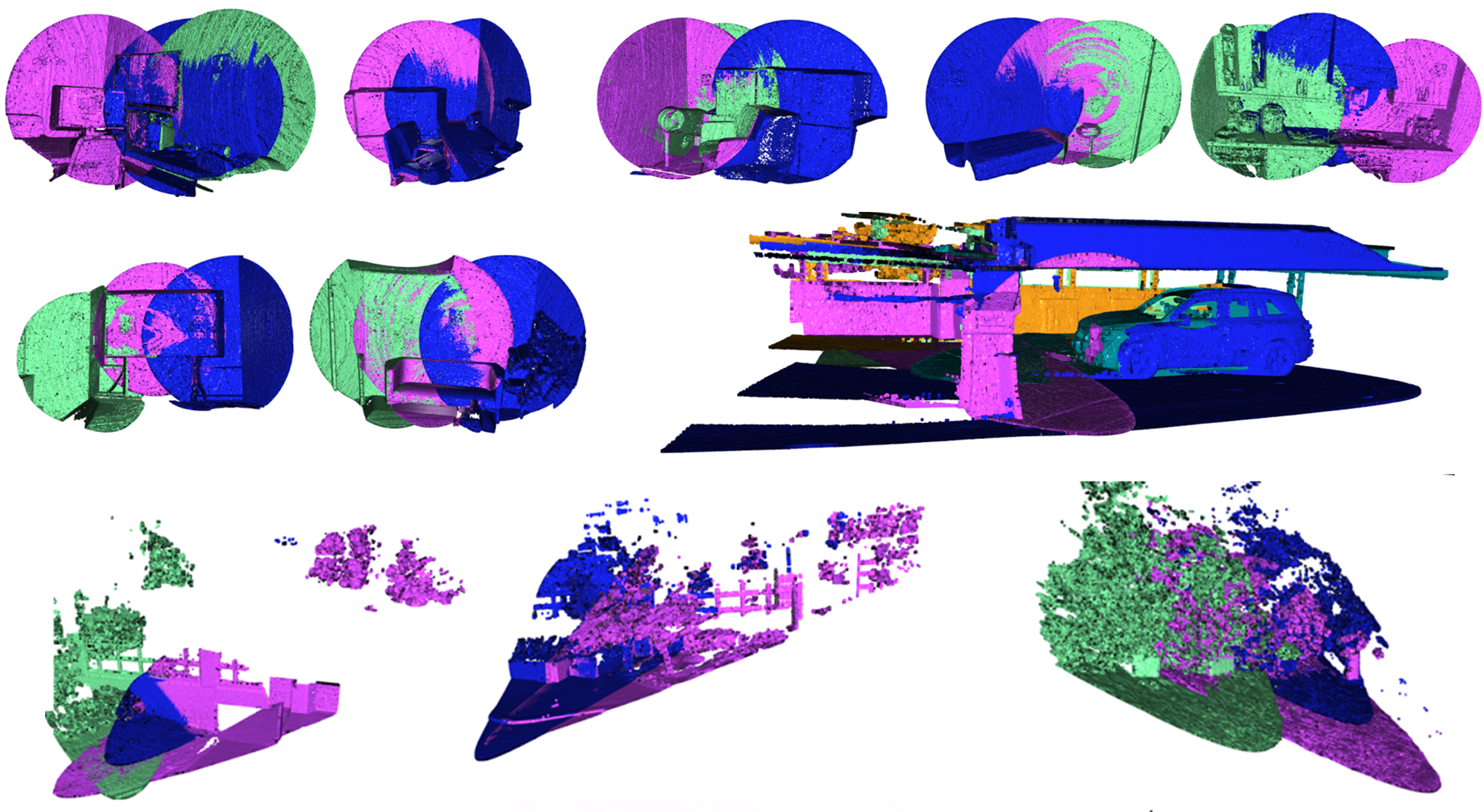}
	\caption{An illustration of the scenes in Livox-3DMatch.}
	\label{livox3dmatchfig}
\end{figure}
\noindent\textbf{Benchmarks and Metrics.}
We compare the performance of the models \cite{mdgd,sghr} trained with and without our data on three popular benchmarks: 3DMatch \cite{3dmatch}, ETH \cite{eth}, and ScanNet \cite{scan}. Following \cite{sghr}, Registration Recall (RR) is used to evaluate performance on 3DMatch \cite{3dmatch} and ETH \cite{eth}, while RMSEs are used to report performance on ScanNet \cite{scan}. RR is defined as follows:
\begin{equation}
{\rm RR} =\frac{N_{success}}{N_{total}}.
\end{equation}
where $N_{success}$ refers to the number of successful registrations, and $N_{total}$ be the total number of registration attempts. For the registration to be considered successful, the RMSEs (See Eq. (\ref{rmse})) must be below a predefined threshold.\par
\noindent\textbf{Results and Analysis.} First, the number of pairs in the training data increases from 14,400 to \textbf{17,700}, a boost of \textbf{22.91\%}, thanks to the introduction of our data. 
The quantitative results are reported in Table \ref{tabsghr}. 
As seen, the RR of SGHR \cite{sghr} is increased by \textbf{2.90\%} on 3DMatch \cite{3dmatch} and \textbf{4.29\%} on ETH \cite{eth}. The translation error and rotation error on ScanNet \cite{scan} are decreased by \textbf{22.72\%} and \textbf{11.19\%}, respectively.
The RR of MDGD \cite{mdgd} is improved by \textbf{1.71\%} on 3DMatch \cite{3dmatch} and \textbf{2.89\%} on ETH \cite{eth}. The translation error and rotation error on ScanNet \cite{scan} are decreased by \textbf{22.45\%} and \textbf{7.80\%}, respectively.
The reason behind these improvements is mentioned in Section \ref{dataset}: the introduction of our data enriches the learnable features during training, which benefits the generalizability of the models.

\begin{table*}[htb]

\caption{Quantitative evaluation of the enhancement of SGHR \cite{sghr} and MDGD \cite{mdgd} due to the addition of Livox-3DMatch to the training.}

\centering
	\resizebox{1.0\columnwidth}{!}{
\begin{tabular}{c|c|c|c|c|c|c|c}
\hline\hline
\multirow{2}{*}{Method} & \multirow{2}{*}{Dataset} & RR $\uparrow$  & \multirow{2}{*}{Dataset} &RR $\uparrow$ & \multirow{2}{*}{Dataset} &  $\mathrm{RMSE}_{T}$ (m) $\downarrow$ & $\mathrm{RMSE}_{R}$ (deg) $\downarrow$ \\ 
  & & (\%) &  &(\%) &  &  Mean/Med & Mean/Med\\ \hline
SGHR \cite{sghr} & \multirow{4}{*}{3DMatch \cite{3dmatch}}  & 92.68 & \multirow{4}{*}{ETH \cite{eth}} &93.74 & \multirow{4}{*}{ScanNet \cite{scan}} &  0.66/0.51 & 23.50/22.08 \\ 
SGHR \cite{sghr} + \textbf{Livox-3DMatch (our data)} &  & \textbf{95.58} &  &\textbf{98.03} &  &  \textbf{0.51/0.45} & \textbf{20.87/17.21} \\ 
MDGD \cite{mdgd}  &  & 94.26 &  & 96.06 &  &  0.49/0.44 & 20.51/19.82 \\ 
MDGD \cite{mdgd} + \textbf{Livox-3DMatch (our data)} &  & \textbf{95.97} &  &\textbf{98.95} &  &  \textbf{0.38/0.31} & \textbf{18.91/17.36} \\ \hline

\hline\hline
\end{tabular}
}
\label{tabsghr}

\end{table*}

\subsection{Application 3: Reconstructing a Degraded Scene} \label{test3}
An important application of fiducial markers in the real world is enhancing the robustness of reconstruction and localization in degraded scenes when additional sensors are unavailable. Therefore, in this test, we evaluate our method in a textureless, degraded scene.  \par
\begin{figure}[H] 
	\centering
\includegraphics[width=1.0\linewidth]{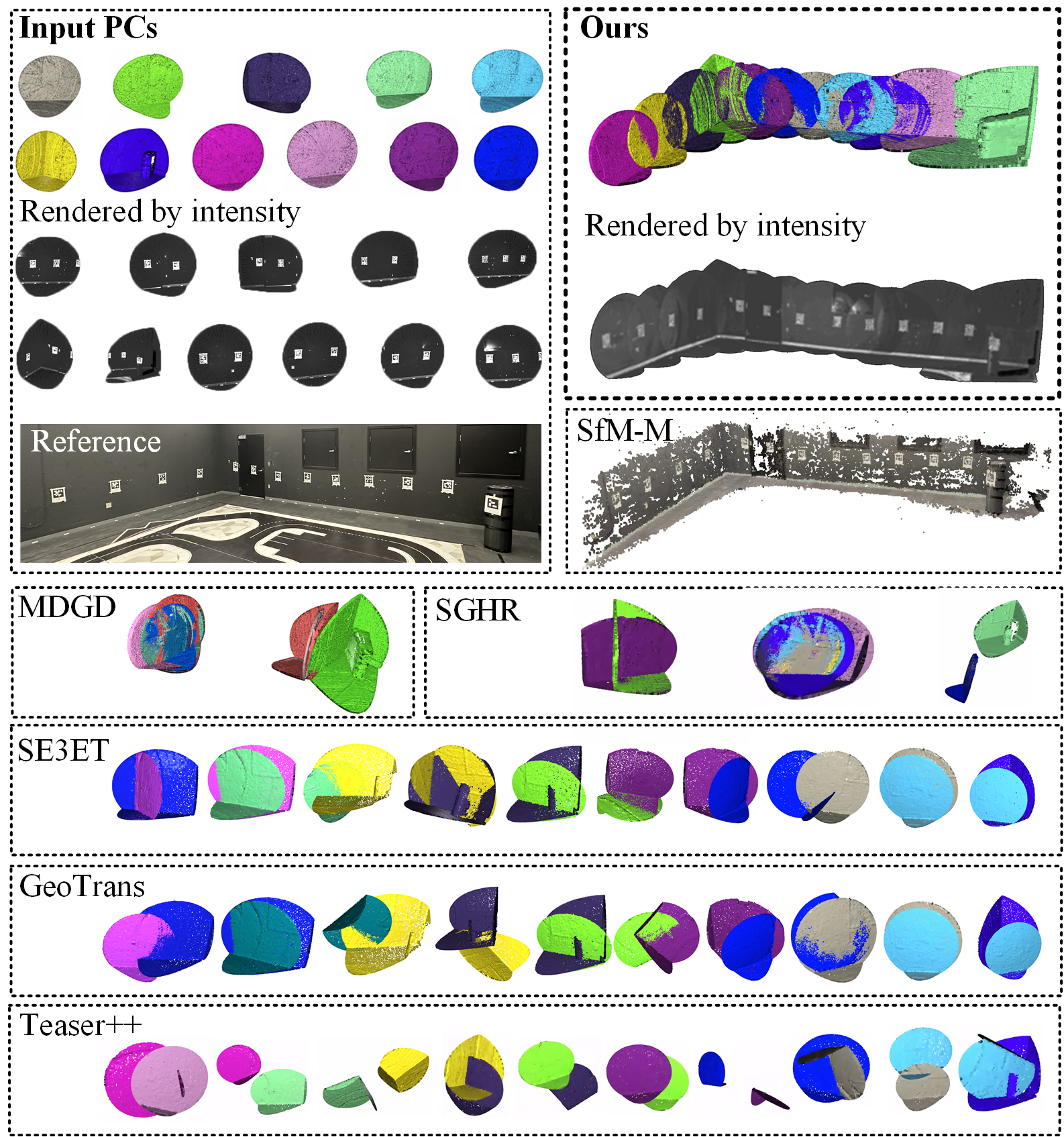}
	\caption{An illustration of the experimental setup and a visual comparison of the proposed method against the SOTA methods (MDGD \cite{mdgd}, SGHR \cite{sghr}, SE3ET \cite{se3et}, GeoTrans \cite{geotransformer}, Teaser++ \cite{teaser}, and SfM-M \cite{qingdao}) in a degraded scene.}
	\label{labpic}
\end{figure}

\noindent\textbf{Data.}
The setup is shown in the reference of Fig. \ref{labpic}. This scenario has repetitive structures and weak geometric features. We attach thirteen 16.4 cm $\times$ 16.4 cm AprilTags to the wall. The LiDAR scans the scene from 11 viewpoints. We also captured 72 images with an iPhone 13 to use as input for SfM-M \cite{qingdao}. The ground truth trajectories are given by an OptiTrack Motion Capture system. 
\par
\begin{figure}[H] 
	\centering
\includegraphics[width=0.6\linewidth]{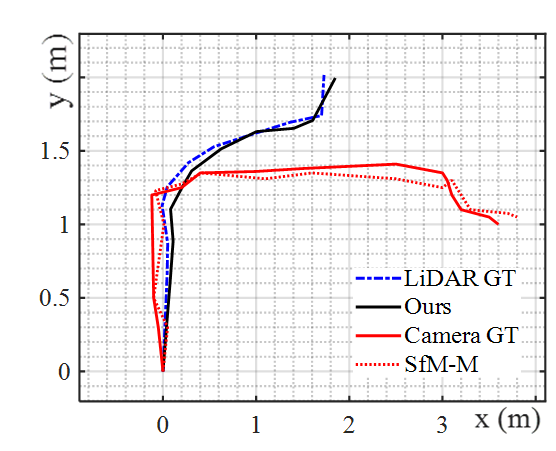}
	\caption{A comparison of the sensor trajectories obtained from the proposed method and SfM-M \cite{qingdao}. G.T. refers to the ground truth.}
	\label{labtraj}
\end{figure}

\noindent\textbf{Competitors and Metrics.}
The competitors are the SOTA point cloud registration methods \cite{mdgd,sghr,se3et,geotransformer,teaser}. However, considering these competitors struggle in the degraded scene, we add SfM-M \cite{qingdao}, the SOTA marker-based SfM method, as a competitor. This addition enables a comprehensive and meaningful comparison. We employ the RMSEs as the metric.\\
\noindent\textbf{Results and Analysis.}
The qualitative results are presented in Fig. \ref{labpic}. As seen, none of the point cloud registration methods \cite{mdgd,sghr,se3et,geotransformer,teaser} can align a single pair of point clouds. 
This is because these cases only have planar overlap regions and thus are even more challenging than those in Fig. \ref{newbench} and Fig. \ref{glbtraj}. 
In comparison, the proposed method and SfM-M \cite{qingdao} successfully reconstruct this degraded scene as they utilize the fiducial markers. The comparison of these two methods in terms of sensor trajectories is shown in Fig. \ref{labtraj}. The quantitative results shown in Table \ref{427} demonstrate that the proposed approach achieves better localization accuracy, which is expected given that LiDAR is a ranging sensor. 
\par

\begin{table}[th]
\caption{Comparison with SfM-M \cite{qingdao} regarding localization accuracy in a degraded scene. }
	\centering
		\begin{tabular}{c|c|c}
			\hline\hline
				Method $\backslash$ Metric & $\mathrm{RMSE}_{T}$ (m) $\downarrow$ & $\mathrm{RMSE}_{R}$ (rad) $\downarrow$\\ \hline
   SfM-M \cite{qingdao} & 0.0551 & 0.0491  \\ \hline
   \textbf{Ours} & 0.0490 & 0.0384
\\ \hline  \hline
			
		\end{tabular}
		\label{427}
  
\end{table}

\subsection{Application 4: Localization in a GPS-denied Environment} \label{test4}
Localization is also a crucial application of point cloud registration methods. Fiducial markers are a popular tool for providing localization information in GPS-denied environments, such as indoor parking lots. In this test, we evaluate the proposed method in this context. \par
\begin{figure}[H] 
	\centering
\includegraphics[width=0.8\linewidth]{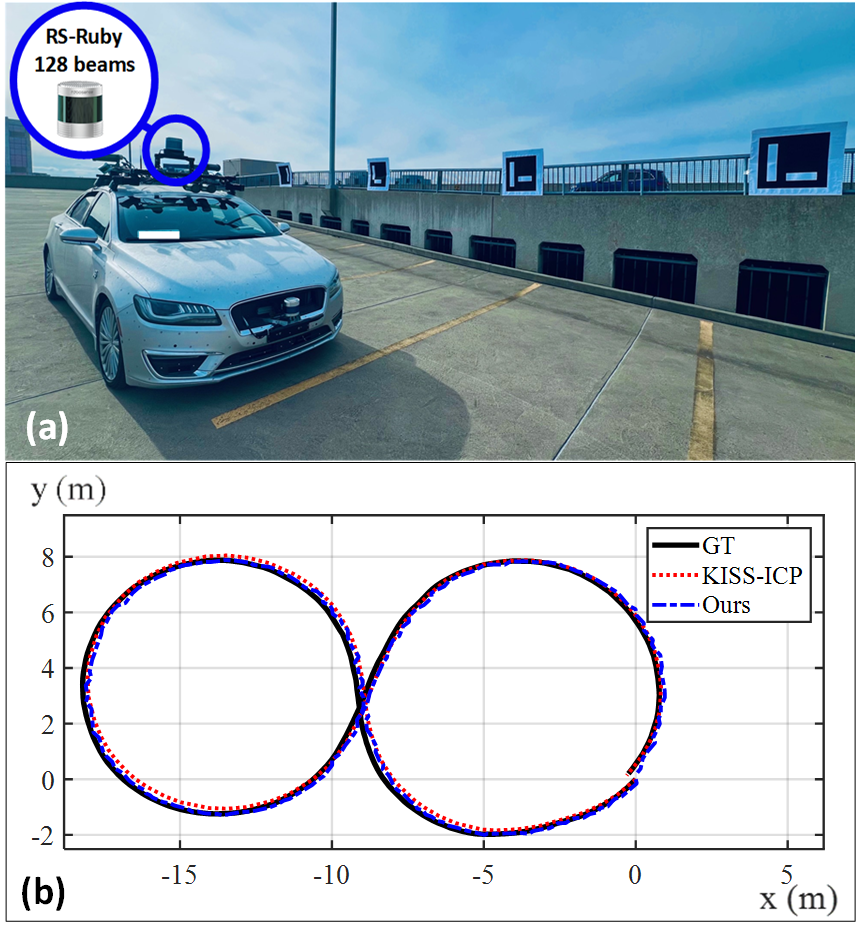}
	\caption{(a): An illustration of the experimental setup. (b): Comparison of the trajectories given by different methods.}
	\label{rob}
\end{figure}

\noindent\textbf{Data.}
The experimental setup is shown in Fig. \ref{rob}(a): four 69.2 cm $\times$ 69.2 cm ArUco \cite{aruco} markers are deployed in the environment. The vehicle, equipped with an RS-Ruby 128-beam mechanical LiDAR, follows an 8-shaped trajectory without pausing and samples 364 LiDAR scans. We conduct the experiment on the roof of a large parking lot to acquire the ground truth trajectory from the Real-Time Kinematic.  \par
\noindent\textbf{Competitor and Metric.}
Note that most LO methods are limited to specific LiDAR models, and modifying the method to accommodate the features of a particular LiDAR model is not trivial \cite{lloam}. 
Considering this, the choice is made to compare with KISS-ICP \cite{kiss}, the state-of-the-art general pure LO method, which can be directly applied to the RS-Ruby 128 LiDAR without requiring modifications. Again, RMSEs are employed as the metric.\par
\noindent\textbf{Results and Analysis.}
The visual comparison and quantitative comparison of localization results are presented in Fig. \ref{rob}(b) and Table \ref{outtab}, respectively. Our method exhibits less drift in the middle of the trajectory and demonstrates better overall localization accuracy.

\begin{table}[th]

\caption{Comparison with KISS-ICP \cite{kiss} regarding localization. }
	\centering
		\begin{tabular}{c|c|c}
			\hline\hline
				Method $\backslash$ Metric & $\mathrm{RMSE}_{T}$ (m) $\downarrow$ & $\mathrm{RMSE}_{R}$ (rad) $\downarrow$\\ \hline
   KISS-ICP \cite{kiss} & 0.1976 & 0.1617  \\ \hline
   \textbf{Ours} & 0.1715 & 0.1394
\\ \hline  \hline
		\end{tabular}
		\label{outtab}
\end{table}

\subsection{Application 5: 3D Map Merging} \label{test5}
\begin{figure}[H]
	\centering
	\includegraphics[width=6.0in]{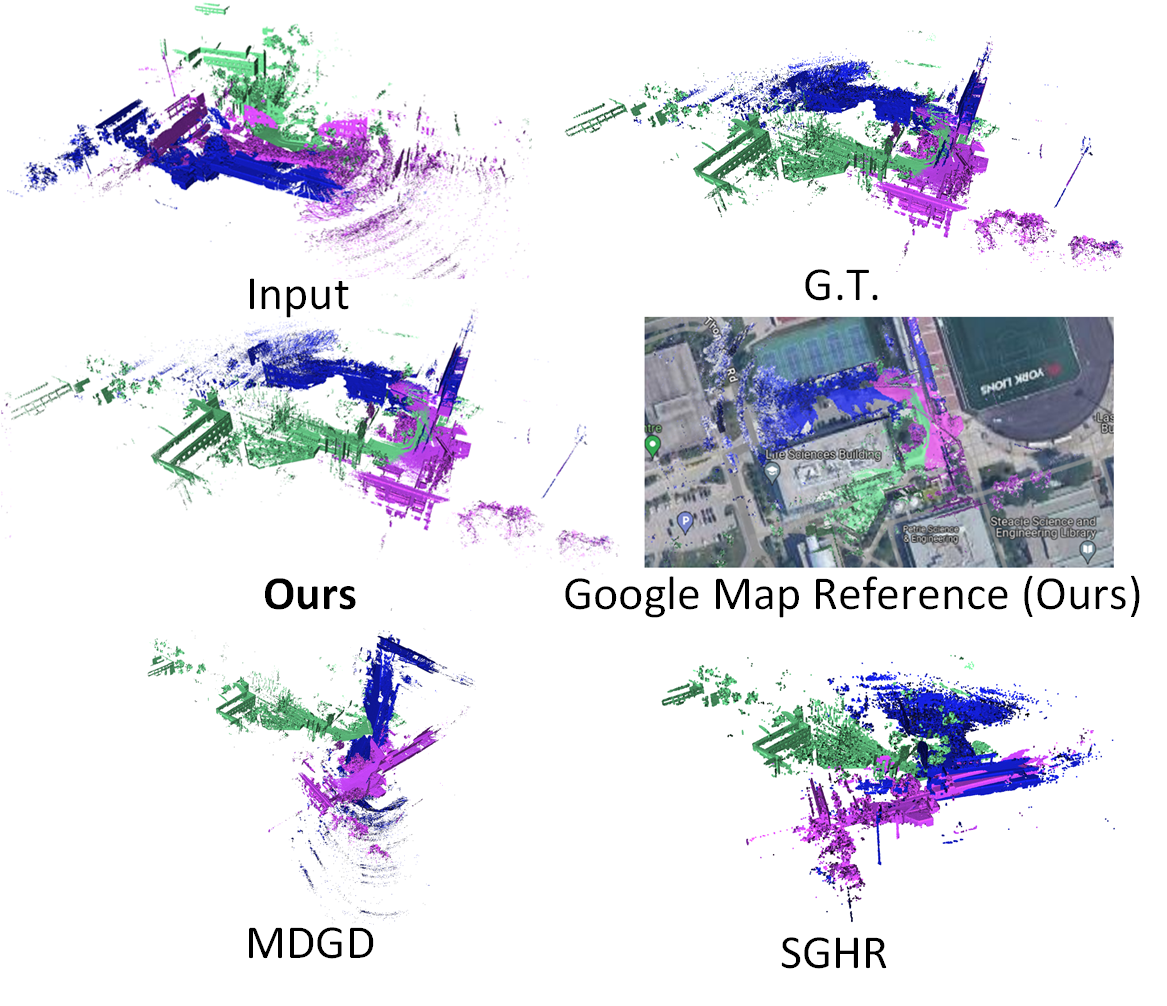}
\color{blue}
	\caption{A comparison of the 3D map merging results of MDGD \cite{mdgd}, SGHR \cite{sghr}, and our method. }
	\label{merge}

\end{figure}

To validate the performance of the proposed method in large-scale scenarios, we apply the proposed method to the 3D map merging task in this test, which involves merging multiple large-scale, low overlap 3D maps into a single frame.
\\
\noindent\textbf{Data.} We collected three large-scale LiDAR maps. They are constructed using the SOTA LiDAR-based SLAM method, Traj-LO \cite{traj}, by scanning the York University campus with a Livox MID-40 LiDAR. The ground truth poses between the maps are manually obtained using \textit{CloudCompare} \cite{cloudcompare}. \\
\noindent\textbf{Competitors and Metrics.}
The competitors are SOTA multiview point cloud registration methods, including MDGD \cite{mdgd} and SGHR \cite{sghr}. Moreover, unlike previous tests, the LiDAR fiducial markers on the 3D maps are localized using the algorithm from our previous work \cite{mapmerge}, and the marker detection results serve as input for the our pipeline.
We employ the RMSEs as the metric.
\\
\noindent\textbf{Results and Analysis.}
The visual and quantitative comparisons are shown in Fig. \ref{merge} and Table \ref{tabmap}, respectively. As shown in Fig. \ref{merge}, neither MDGD nor SGHR can address this challenging task. This is due to the fact that the overlap in large-scale maps is too scarce. Specifically, the overlap rates \cite{pre} are 4.87\% between map 1 and map 2, 3.96\% between map 1 and map 3, and 2.59\% between map 2 and map 3. On the other hand, both MDGD and SGHR start by analyzing the features of each individual point cloud. As the scale of the point cloud becomes larger, the portion of the features belonging to the overlap regions decreases, making them unsuitable for large-scale, low overlap map merging. In addition, as the scale of point clouds becomes larger, the absolute error values of MDGD and SGHR also increase. In comparison, even though the scales of the point clouds in this test are much larger than those in the previous test, the proposed method successfully merges these large-scale, low overlap 3D maps. This stems from the observation that the proposed method focuses on utilizing the thin-sheet LiDAR fiducial markers and is not sensitive to scale changes. As introduced in \cite{mapmerge}, the accuracy of marker localization degrades as the scale increases. Thus, the absolute error values of our method also increase compared to previous tests.

\begin{table}[H]

\caption{Comparison of 3D map merging results between MDGD \cite{mdgd}, SGHR \cite{sghr}, and our method.}
	\centering
	\resizebox{0.6\columnwidth}{!}{
		\begin{tabular}{c|c|c}
			\hline\hline
				Method $\backslash$ Metric & $\mathrm{RMSE}_{T}$ (m) $\downarrow$ & $\mathrm{RMSE}_{R}$ (rad) $\downarrow$\\ \hline
   MDGD \cite{mdgd} & 2.7883 & 1.9210  \\ \hline
   SGHR \cite{sghr} & 4.0933 & 2.3925  \\ \hline
   \textbf{Ours} & 0.2305 & 0.1771
\\ \hline  \hline
			
		\end{tabular}
	}
		\label{tabmap}
\end{table}

\subsection{Ablation Studies} \label{ab}
\begin{figure}[h]
	\centering
  
	\includegraphics[width=6.0in]{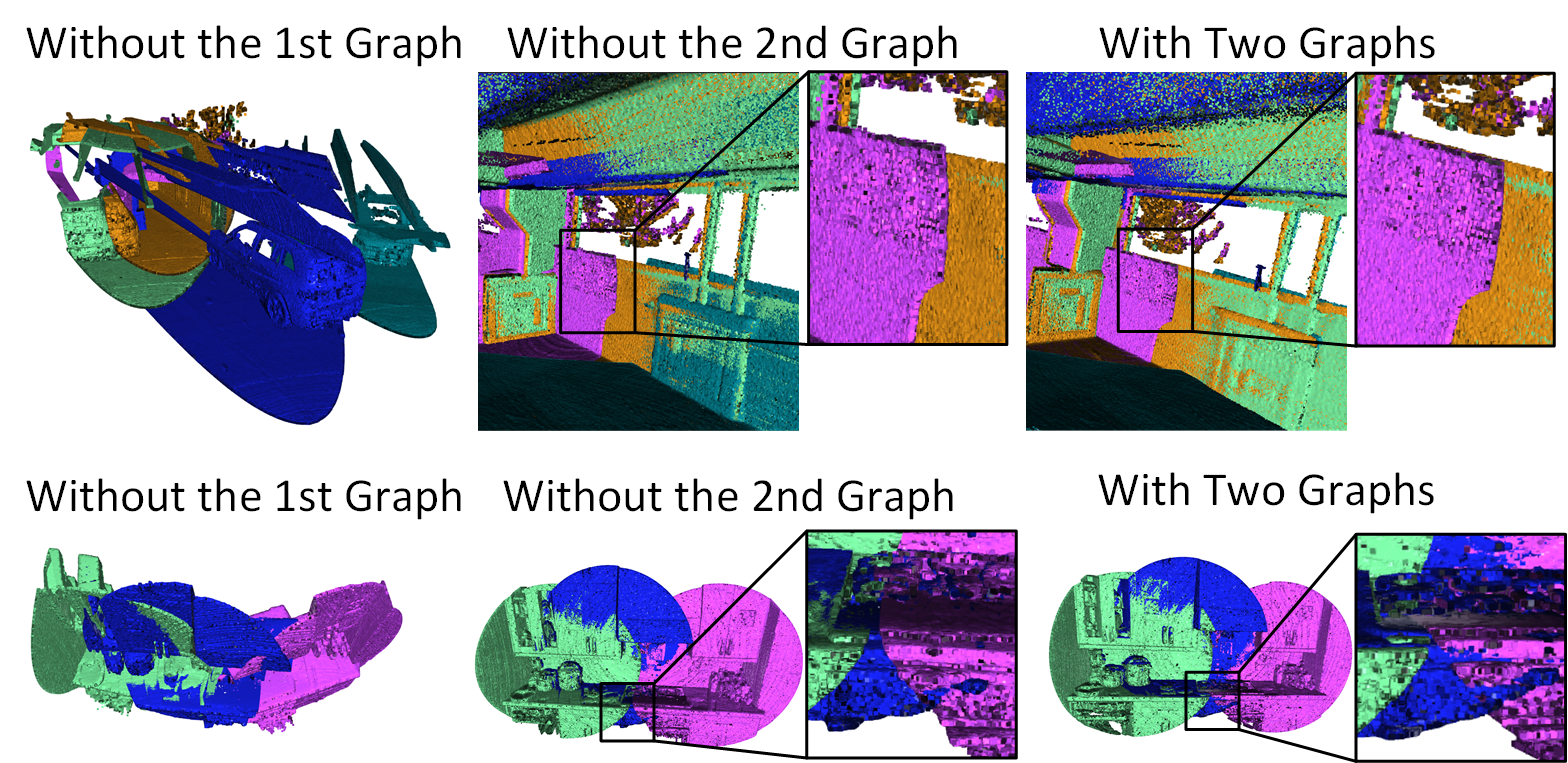}

	\caption{ A comparison of the point cloud registration results without the first graph, without the second graph, and with both graphs for two cases. }
	\label{abstudy}

\end{figure}

The proposed framework is composed of two levels of graphs. To demonstrate the necessity of this overall architecture, we conduct ablation studies in this section. In particular, we study the effects of removing the first and second graphs in two cases: Fig. \ref{glbtraj} and the kitchen scene in Fig. \ref{newbench}. The visual comparison and quantitative results are shown in Fig. \ref{abstudy} and Table \ref{abtab}, respectively. When the first graph is removed, the factors representing the relative poses between frames in Fig. \ref{second} have to be set to identity due to the lack of initial values provided by the first graph. 
However, the role of the second graph is to further optimize the variables based on their initial values rather than finding the optimal solution from scratch.
Consequently, as shown in Fig. \ref{abstudy} and Table \ref{abtab}, severe misalignment or degradation in registration accuracy occurs.
When the second graph is removed, the multiview point clouds are directly registered using the initial values obtained from the first graph, without any further refinement. 
As shown in Fig. \ref{abstudy} and Table \ref{abtab}, removing the second graph causes a slight degradation in registration accuracy.
Moreover, the degradation in the kitchen case of Fig. \ref{newbench} is slighter than that in Fig. \ref{glbtraj}.
This is because the effect of the second graph is case-by-case, determined by the quality of the initial values provided by the first graph. 
Namely, the better the initial values are, the less important the second graph becomes. However, in the real world, pose estimation of markers cannot be perfect. In addition, as seen in Fig. \ref{secondgraph}, we also add the marker corners to the second graph so that the pose estimation of each individual marker can be further optimized along with other variables in the graph.
Therefore, it is beneficial to apply the second graph in practice. In summary, the proposed framework adopts a coarse-to-fine pipeline, where the first graph corresponds to the coarse stage, while the second graph is the fine stage. Removing any of them will cause performance degradation.

\begin{table}[h]
\caption{Ablation studies on the first and second graphs.}
	\centering
	\resizebox{0.7\columnwidth}{!}{
		\begin{tabular}{c|c|c|c}
			\hline\hline
				 Scene & Framework & $\mathrm{RMSE}_{T}$ (m) $\downarrow$ & $\mathrm{RMSE}_{R}$ (rad) $\downarrow$\\ \hline
\multirow{3}{*}{Fig. \ref{glbtraj}} & without first graph & 0.431 & 0.459 \\
& without second graph & 0.078 & 0.084 \\
& with both graphs & 0.066 & 0.072 \\
\hline
\multirow{3}{*}{Fig. \ref{newbench}} & without first graph & 0.088 & 0.112 \\
& without second graph & 0.021 & 0.073 \\
(Kitchen) & with both graphs & 0.019 & 0.067 \\
\hline  \hline
			
		\end{tabular}
	}
		\label{abtab}
\end{table}

\subsection{Limitations} 
Despite the promising results of the proposed algorithm, there are potential limitations. Firstly, although LiDARs are robust to unideal illumination conditions, adverse weather phenomena such as rain, snow, and fog can affect LiDAR measurements, thereby reducing the effectiveness of the proposed method. Secondly, the adopted low-cost LFMs, made of thin-sheet paper or boards, may deform after long-term use in the wild. However, this is not a concern for one-time applications such as data collection. Finally, deploying the LFMs requires some labor, but their value is demonstrated in this dissertation.

\chapter{Conclusion and Future Work \label{conclusion}}

\section{Conclusion}
To facilitate robotics and computer vision applications, a LiDAR-based mapping and localization method is needed that can robustly handle unordered, low overlap, multiview point clouds simultaneously, similar to how Structure-from-Motion (SfM) for cameras processes unordered images all at once.
However, unlike in the field of SfM, where the use of visual fiducial markers (VFMs) has been widely studied, the development of the LiDAR fiducial marker (LFM) and its utilization in such a framework remains a technological gap. This dissertation addresses these pressing issues from the following perspectives.
\par
To address the lack of a general LFM system that can be applied to different LiDAR models and is compatible with popular VFM patterns, a novel intensity image-based LiDAR fiducial marker (IFM) system is proposed. 
\begin{itemize}
\item Unlike LiDARTag \cite{lt} which requires extra 3D objects to be added to the environment, the usage of the IFM is as convenient and easy as the VFM systems \cite{ap3,aruco}. Namely, the users can produce the marker by printing the VFM on regular letter-size paper with a regular printer and then place the marker anywhere they like. 
\item A novel marker detection method is proposed to detect 3D fiducials through the intensity image. Thanks to this, the VFM systems proposed in the past, present, and even future can be easily embedded into the IFM system. This is a superiority of the proposed system over LiDARTag \cite{lt} which only supports square markers with patterns from AprilTag 3 \cite{ap3}. 
\item A pose estimation approach for the LiDAR via the proposed IFM is introduced, which has better accuracy than the VFM-based pose estimation for the camera. In addition, unlike the VFM system, the proposed pose estimation is free from the rotation ambiguity problem \cite{ippe,yibo} and is robust to changes in ambient light. 
\end{itemize} \par
Due to the adoption of 3D-to-2D spherical projection, the vanilla IFM exhibits two limitations: (i) IFM can only detect fiducials in a single-view point cloud and does not apply to a 3D LiDAR map, and (ii) as the distance between the tag and the LiDAR increases, the projection size of the tag decreases until it is too small to be detected. In response to these limitations, a novel method has been developed to improve the localization of thin-sheet LFMs.
\begin{itemize}
	\item The LFM localization is extended from the single-view point cloud to the 3D LiDAR map and the detectable distance of LFMs is enhanced. 
	\item A new pipeline for jointly analyzing a 3D point cloud from both intensity and geometry perspectives is proposed, as fiducial tags are planar objects with high-intensity contrast and are indistinguishable from the plane to which they are attached. This differs from conventional 3D object detection methods that rely merely on 3D geometric features and can only detect spatially distinguishable objects.
\end{itemize} \par
Finally, building upon the developed algorithms for LFM detection, this dissertation discusses how to exploit LFMs for mapping and localization.
\begin{itemize}
	\item A novel framework for mapping and localization using LFMs is designed, where mapping and localization are achieved by registering unordered multiview point clouds, even with low overlap. The proposed framework serves as an efficient and reliable tool for diverse robotics and computer vision applications, including 3D asset collection, training data collection, reconstruction of degraded scenes, localization in GPS-denied environments, and 3D map merging.
	\item A new training dataset called Livox-3DMatch using the proposed framework is collected, which augments the original 3DMatch training data \cite{3dmatch} from 14,400 pairs to 17,700 pairs (a 22.91\% increase). Training on this augmented dataset improves the performance of the SOTA learning-based methods on various benchmarks. In particular, the RR of SGHR \cite{sghr} is increased by 2.90\% on 3DMatch \cite{3dmatch} and 4.29\% on ETH \cite{eth}. The translation error and rotation error on ScanNet \cite{scan} are decreased by 22.72\% and 11.19\%, respectively. The RR of MDGD \cite{mdgd} increases by 1.71\% on 3DMatch \cite{3dmatch} and 2.89\% on ETH \cite{eth}. The translation error and rotation error on ScanNet \cite{scan} decrease by 22.45\% and 7.80\%, respectively.
\item An adaptive threshold LFM detection algorithm that is robust to viewpoint changes in the wild is developed.
 
\end{itemize}

\section{Future Work}
In this dissertation, a framework for mapping and localization using LFMs is proposed to benefit various robotic and computer vision applications. While the results are promising, several issues remain that require further exploration. The following are potential future research directions to enhance this work.
\par	
The first issue is the utilization of learning-based methods in LFM detection. Since the proposed IFM is open-source, much inspiring user feedback has been received, which also motivated the reconsideration of LFM detection. For example, both LiDARTag \cite{lt} and IFM \cite{iilfm} propose introducing the encoding-decoding algorithms of VFM systems in LFM detection. However, since the 'code' of a marker's pattern typically exceeds 16 bits—meaning more than 16 black-and-white blocks—the LiDAR must have a high resolution to fully capture the code/bit information. Based on feedback from users in the autonomous driving industry who use LFM for calibration, marker detection sometimes fails as the bit information is lost when the distance increases. Unfortunately, this is caused by the sparse features of the LiDAR data, which have not been adequately addressed so far. Despite this, a notable technology called LiDAR intensity densification \cite{dense}, which learns to generate densified LiDAR intensity data from sparse and occluded LiDAR data, has the potential to mitigate this problem.
Another solution that has a similar effect is Neural LiDAR Fields \cite{lidarnerf}, which learns LiDAR-based novel view synthesis and supports the upsampling of LiDAR data. Moreover, considering that many 3D object detection and segmentation models are designed to learn directly from sparse and occluded data, it is possible to tackle this problem using a learning-based method. Nevertheless, it should be noted that the lack of corresponding training data presents a significant challenge. In particular, the training data should include a substantial number of annotated thin-sheet LFMs, and the model should be designed to learn detection by utilizing intensity information.
\par
The second issue is the need to collect more data using the proposed framework. Although a new dataset named Livox-3DMatch has been collected for point cloud registration research in this dissertation, it would be beneficial to use the proposed framework to collect additional data and a greater variety of datasets. For example, 3D assets of irregular vehicles, such as construction trucks equipped with various machines, are rare but crucial for autonomous driving simulation. Thus, collecting more 3D assets of irregular vehicles using the proposed framework is advantageous. LiDAR-based novel view synthesis is an emerging topic. In camera-based research, large-scale datasets like OmniObject3D \cite{omni} provide extensive object-level posed images. Developing a similar dataset for LiDAR that offers posed point clouds would enhance LiDAR-based novel view synthesis \cite{lidarnerf,lidarnerf2}, which is a significant downstream and future application of the proposed framework.
\par
The third issue is to explore other types of materials and formats to construct markers for better utilization of the intensity information. In this research, we only study black-and-white markers made of paper, as we aim to develop a low-cost solution. However, for scenes where the distance between the fiducial and the sensor is hundreds of meters, it is beneficial to introduce fiducial objects such as prisms. Moreover, by leveraging other materials and formats, such as prisms, to construct LFMs, it is feasible to eliminate the reliance on 2D space, while keeping the system as unified as prisms work for different LiDAR models.
\par
The fourth issue is to fuse LFM detection with VFM detection for a robust autonomous system. Although utilizing the algorithms proposed in this research, it is feasible to detect thin-sheet markers with LiDARs alone, redundant design is considered necessary in autonomous systems nowadays. As introduced in the dissertation, LFM detection can help address rotational ambiguity and unideal illumination conditions, which are challenging for VFM detection. On the other hand, there are many mature computer vision algorithms to derain, defog, and desnow, which are challenging for LFM detection. In addition, fusing LFM and VFM detection and constructing the pose estimation of an autonomous system as a global optimization problem has the potential to provide better accuracy.

\bibliographystyle{IEEEtran}
\bibliography{RefDissertation}

\begin{thebibliography}{10}
\providecommand{\url}[1]{#1}
\csname url@samestyle\endcsname
\providecommand{\newblock}{\relax}
\providecommand{\bibinfo}[2]{#2}
\providecommand{\BIBentrySTDinterwordspacing}{\spaceskip=0pt\relax}
\providecommand{\BIBentryALTinterwordstretchfactor}{4}
\providecommand{\BIBentryALTinterwordspacing}{\spaceskip=\fontdimen2\font plus
\BIBentryALTinterwordstretchfactor\fontdimen3\font minus \fontdimen4\font\relax}
\providecommand{\BIBforeignlanguage}[2]{{%
\expandafter\ifx\csname l@#1\endcsname\relax
\typeout{** WARNING: IEEEtran.bst: No hyphenation pattern has been}%
\typeout{** loaded for the language `#1'. Using the pattern for}%
\typeout{** the default language instead.}%
\else
\language=\csname l@#1\endcsname
\fi
#2}}
\providecommand{\BIBdecl}{\relax}
\BIBdecl

\bibitem{ap3}
M.~Krogius, A.~Haggenmiller, and E.~Olson, ``Flexible layouts for fiducial tags,'' in \emph{Proc. IEEE/RSJ International Conference on Intelligent Robots and Systems}, 2019, pp. 1898--1903.

\bibitem{mdgd}
S.~Li, J.~Zhu, Y.~Xie, N.~Hu, and D.~Wang, ``Matching distance and geometric distribution aided learning multiview point cloud registration,'' \emph{IEEE Robotics and Automation Letters}, vol.~9, no.~11, pp. 9319--9326, 2024.

\bibitem{sghr}
H.~Wang, Y.~Liu, Z.~Dong, Y.~Guo, Y.-S. Liu, W.~Wang, and B.~Yang, ``Robust multiview point cloud registration with reliable pose graph initialization and history reweighting,'' in \emph{Proc. of the IEEE/CVF Conference on Computer Vision and Pattern Recognition}, 2023, pp. 9506--9515.

\bibitem{se3et}
C.~E. Lin, M.~Zhu, and M.~Ghaffari, ``Se3et: Se(3)-equivariant transformer for low-overlap point cloud registration,'' \emph{IEEE Robotics and Automation Letters}, vol.~9, no.~11, pp. 9526--9533, 2024.

\bibitem{geotransformer}
Z.~Qin, H.~Yu, C.~Wang, Y.~Guo, Y.~Peng, S.~Ilic, D.~Hu, and K.~Xu, ``Geotransformer: Fast and robust point cloud registration with geometric transformer,'' \emph{IEEE Transactions on Pattern Analysis and Machine Intelligence}, vol.~45, no.~8, pp. 9806--9821, 2023.

\bibitem{teaser}
H.~Yang, J.~Shi, and L.~Carlone, ``Teaser: Fast and certifiable point cloud registration,'' \emph{IEEE Transactions on Robotics}, vol.~37, no.~2, pp. 314--333, 2020.

\bibitem{qingdao}
Z.~Jia, Y.~Rao, H.~Fan, and J.~Dong, ``An efficient visual sfm framework using planar markers,'' \emph{IEEE Transactions on Instrumentation and Measurement}, vol.~72, pp. 1--12, 2023.

\bibitem{kiss}
I.~Vizzo, T.~Guadagnino, B.~Mersch, L.~Wiesmann, J.~Behley, and C.~Stachniss, ``{KISS-ICP: In Defense of Point-to-Point ICP -- Simple, Accurate, and Robust Registration If Done the Right Way},'' \emph{IEEE Robotics and Automation Letters}, vol.~8, no.~2, pp. 1029--1036, 2023.

\bibitem{lt}
J.-K. Huang, S.~Wang, M.~Ghaffari, and J.~W. Grizzle, ``Lidartag: A real-time fiducial tag system for point clouds,'' \emph{IEEE Robotics and Automation Letters}, vol.~6, no.~3, pp. 4875--4882, 2021.

\bibitem{traj}
X.~Zheng and J.~Zhu, ``Traj-lo: In defense of lidar-only odometry using an effective continuous-time trajectory,'' \emph{IEEE Robotics and Automation Letters}, vol.~9, no.~2, pp. 1961--1968, 2024.

\bibitem{dij}
E.~W. Dijkstra, ``A note on two problems in connexion with graphs,'' in \emph{Edsger Wybe Dijkstra: His Life, Work, and Legacy}, 2022, pp. 287--290.

\bibitem{sdk}
\BIBentryALTinterwordspacing
Livox-SDK. {Livox Mapping}. (2020). [Online]. Available: \url{https://github.com/Livox-SDK/livox_mapping}
\BIBentrySTDinterwordspacing

\bibitem{lloam}
J.~Lin and F.~Zhang, ``Loam livox: A fast, robust, high-precision lidar odometry and mapping package for lidars of small fov,'' in \emph{Proc. IEEE International Conference on Robotics and Automation}, 2020, pp. 3126--3131.

\bibitem{loam}
J.~Zhang and S.~Singh, ``Loam: Lidar odometry and mapping in real-time.'' in \emph{Robotics: Science and systems}, vol.~2, no.~9.\hskip 1em plus 0.5em minus 0.4em\relax Berkeley, CA, 2014, pp. 1--9.

\bibitem{rangenet}
L.~Fan, X.~Xiong, F.~Wang, N.~Wang, and Z.~Zhang, ``Rangedet: In defense of range view for lidar-based 3d object detection,'' in \emph{Proc. IEEE/CVF International Conference on Computer Vision}, 2021.

\bibitem{wang}
J.~Wang and E.~Olson, ``Apriltag 2: Efficient and robust fiducial detection,'' in \emph{Proc. IEEE/RSJ International Conference on Intelligent Robots and Systems}, 2016, pp. 4193--4198.

\bibitem{olson}
E.~Olson, ``Apriltag: A robust and flexible visual fiducial system,'' in \emph{Proc. IEEE International Conference on Robotics and Automation}, 2011, pp. 3400--3407.

\bibitem{aruco}
F.~J. Romero-Ramirez, R.~Mu{\~n}oz-Salinas, and R.~Medina-Carnicer, ``Speeded up detection of squared fiducial markers,'' \emph{Image and vision Computing}, vol.~76, pp. 38--47, 2018.

\bibitem{cctag}
L.~Calvet, P.~Gurdjos, C.~Griwodz, and S.~Gasparini, ``{Detection and Accurate Localization of Circular Fiducials under Highly Challenging Conditions},'' in \emph{{Proc. IEEE Conference on Computer Vision and Pattern Recognition}}, 2016, pp. 562 -- 570.

\bibitem{ar}
D.~Avola, L.~Cinque, G.~L. Foresti, C.~Mercuri, and D.~Pannone, ``A practical framework for the development of augmented reality applications by using aruco markers,'' in \emph{Proc. International Conference on Pattern Recognition Applications and Methods}, vol.~2, 2016, pp. 645--654.

\bibitem{yibo1}
Y.~Liu, H.~Schofield, and J.~Shan, ``Navigation of a self-driving vehicle using one fiducial marker,'' in \emph{Proc. IEEE International Conference on Multisensor Fusion and Integration for Intelligent Systems}, 2021, pp. 1--6.

\bibitem{yibo2}
Y.~Liu, A.~Haridevan, H.~Schofield, and J.~Shan, ``Application of ghost-deblurgan to fiducial marker detection,'' in \emph{Proc. IEEE/RSJ International Conference on Intelligent Robots and Systems}, 2022, pp. 6827--6832.

\bibitem{liao}
T.~Liao, A.~Haridevan, Y.~Liu, and J.~Shan, ``Autonomous vision-based uav landing with collision avoidance using deep learning,'' in \emph{Science and Information Conference}.\hskip 1em plus 0.5em minus 0.4em\relax Springer, 2022, pp. 79--87.

\bibitem{kalibr}
J.~Rehder, J.~Nikolic, T.~Schneider, T.~Hinzmann, and R.~Siegwart, ``Extending kalibr: Calibrating the extrinsics of multiple imus and of individual axes,'' in \emph{Proc. IEEE International Conference on Robotics and Automation}, 2016, pp. 4304--4311.

\bibitem{lt2}
J.-K. Huang and J.~W. Grizzle, ``Improvements to target-based 3d lidar to camera calibration,'' \emph{IEEE Access}, vol.~8, pp. 134\,101--134\,110, 2020.

\bibitem{munoz}
R.~Mu{\~n}oz-Salinas, M.~J. Mar{\'\i}n-Jimenez, E.~Yeguas-Bolivar, and R.~Medina-Carnicer, ``Mapping and localization from planar markers,'' \emph{Pattern Recognition}, vol.~73, pp. 158--171, 2018.

\bibitem{munoz2019}
R.~Mu{\~n}oz-Salinas, M.~J. Marin-Jimenez, and R.~Medina-Carnicer, ``Spm-slam: Simultaneous localization and mapping with squared planar markers,'' \emph{Pattern Recognition}, vol.~86, pp. 156--171, 2019.

\bibitem{shuo}
S.~Zhang, J.~Shan, and Y.~Liu, ``Variational bayesian estimator for mobile robot localization with unknown noise covariance,'' \emph{IEEE/ASME Transactions on Mechatronics}, vol.~27, no.~4, pp. 2185--2193, 2022.

\bibitem{shuo2}
------, ``Particle filtering on lie group for mobile robot localization with range-bearing measurements,'' \emph{IEEE Control Systems Letters}, vol.~7, pp. 3753--3758, 2023.

\bibitem{multiway}
S.~Jin, I.~Armeni, M.~Pollefeys, and D.~Barath, ``Multiway point cloud mosaicking with diffusion and global optimization,'' in \emph{Proceedings of the IEEE/CVF Conference on Computer Vision and Pattern Recognition}, 2024, pp. 20\,838--20\,849.

\bibitem{shuo3}
S.~Zhang, J.~Shan, and Y.~Liu, ``Approximate inference particle filtering for mobile robot slam,'' \emph{IEEE Transactions on Automation Science and Engineering}, pp. 1--12, 2024.

\bibitem{waabi}
Z.~Yang, S.~Manivasagam, Y.~Chen, J.~Wang, R.~Hu, and R.~Urtasun, ``Reconstructing objects in-the-wild for realistic sensor simulation,'' in \emph{Proc. of IEEE International Conference on Robotics and Automation}, 2023, pp. 11\,661--11\,668.

\bibitem{waymo}
P.~Sun, H.~Kretzschmar, X.~Dotiwalla, A.~Chouard, V.~Patnaik, P.~Tsui, J.~Guo, Y.~Zhou, Y.~Chai, B.~Caine \emph{et~al.}, ``Scalability in perception for autonomous driving: Waymo open dataset,'' in \emph{Proc. of the IEEE/CVF conference on computer vision and pattern recognition}, 2020, pp. 2446--2454.

\bibitem{cd}
Y.~Liu, K.~Zhu, G.~Wu, Y.~Ren, B.~Liu, Y.~Liu, and J.~Shan, ``Mv-deepsdf: Implicit modeling with multi-sweep point clouds for 3d vehicle reconstruction in autonomous driving,'' in \emph{Proc. of the IEEE/CVF International Conference on Computer Vision}, 2023, pp. 8306--8316.

\bibitem{3dmatch}
A.~Zeng, S.~Song, M.~Nie{\ss}ner, M.~Fisher, J.~Xiao, and T.~Funkhouser, ``3dmatch: Learning local geometric descriptors from rgb-d reconstructions,'' in \emph{Proc. of the IEEE conference on computer vision and pattern recognition}, 2017, pp. 1802--1811.

\bibitem{eth}
F.~Pomerleau, M.~Liu, F.~Colas, and R.~Siegwart, ``Challenging data sets for point cloud registration algorithms,'' \emph{The International Journal of Robotics Research}, vol.~31, no.~14, pp. 1705--1711, 2012.

\bibitem{scan}
A.~Dai, A.~X. Chang, M.~Savva, M.~Halber, T.~Funkhouser, and M.~Nie{\ss}ner, ``Scannet: Richly-annotated 3d reconstructions of indoor scenes,'' in \emph{Proceedings of the IEEE conference on computer vision and pattern recognition}, 2017, pp. 5828--5839.

\bibitem{mending}
S.~Duggal, Z.~Wang, W.-C. Ma, S.~Manivasagam, J.~Liang, S.~Wang, and R.~Urtasun, ``Mending neural implicit modeling for 3d vehicle reconstruction in the wild,'' in \emph{Proc. of the IEEE/CVF Winter Conference on Applications of Computer Vision}, 2022, pp. 1900--1909.

\bibitem{iilfm}
Y.~Liu, H.~Schofield, and J.~Shan, ``Intensity image-based lidar fiducial marker system,'' \emph{IEEE Robotics and Automation Letters}, vol.~7, no.~3, pp. 6542--6549, 2022.

\bibitem{munoz2018}
R.~Mu{\~n}oz-Salinas, M.~J. Mar{\'\i}n-Jimenez, E.~Yeguas-Bolivar, and R.~Medina-Carnicer, ``Mapping and localization from planar markers,'' \emph{Pattern Recognition}, vol.~73, pp. 158--171, 2018.

\bibitem{cal}
J.~Beltr{\'a}n, C.~Guindel, A.~de~la Escalera, and F.~Garc{\'\i}a, ``Automatic extrinsic calibration method for lidar and camera sensor setups,'' \emph{IEEE Transactions on Intelligent Transportation Systems}, vol.~23, no.~10, pp. 17\,677--17\,689, 2022.

\bibitem{cal2}
L.~Tao, L.~Pei, T.~Li, D.~Zou, Q.~Wu, and S.~Xia, ``Cpi: Lidar-camera extrinsic calibration based on feature points with reflection intensity,'' in \emph{Spatial Data and Intelligence: First International Conference, SpatialDI 2020, Virtual Event, May 8--9, 2020, Proceedings 1}.\hskip 1em plus 0.5em minus 0.4em\relax Springer, 2021, pp. 281--290.

\bibitem{a4}
Y.~Xie, L.~Deng, T.~Sun, Y.~Fu, J.~Li, X.~Cui, H.~Yin, S.~Deng, J.~Xiao, and B.~Chen, ``A4lidartag: Depth-based fiducial marker for extrinsic calibration of solid-state lidar and camera,'' \emph{IEEE Robotics and Automation Letters}, vol.~7, no.~3, pp. 6487--6494, 2022.

\bibitem{random}
M.~A. Fischler and R.~C. Bolles, ``Random sample consensus: a paradigm for model fitting with applications to image analysis and automated cartography,'' \emph{Communications of the ACM}, vol.~24, no.~6, pp. 381--395, 1981.

\bibitem{pre}
S.~Huang, Z.~Gojcic, M.~Usvyatsov, A.~Wieser, and K.~Schindler, ``Predator: Registration of 3d point clouds with low overlap,'' in \emph{Proc. of the IEEE/CVF Conference on Computer Vision and Pattern Recognition}, June 2021, pp. 4267--4276.

\bibitem{degradation}
C.~Yang, Z.~Chai, X.~Yang, H.~Zhuang, and M.~Yang, ``Recognition of degradation scenarios for lidar slam applications,'' in \emph{Proc. IEEE International Conference on Robotics and Biomimetics}.\hskip 1em plus 0.5em minus 0.4em\relax IEEE, 2022, pp. 1726--1731.

\bibitem{colmap}
J.~L. Schonberger and J.-M. Frahm, ``Structure-from-motion revisited,'' in \emph{Proceedings of the IEEE conference on computer vision and pattern recognition}, 2016, pp. 4104--4113.

\bibitem{dust3r}
S.~Wang, V.~Leroy, Y.~Cabon, B.~Chidlovskii, and J.~Revaud, ``Dust3r: Geometric 3d vision made easy,'' in \emph{Proceedings of the IEEE/CVF Conference on Computer Vision and Pattern Recognition}, 2024, pp. 20\,697--20\,709.

\bibitem{omni}
T.~Wu, J.~Zhang, X.~Fu, Y.~Wang, J.~Ren, L.~Pan, W.~Wu, L.~Yang, J.~Wang, C.~Qian \emph{et~al.}, ``Omniobject3d: Large-vocabulary 3d object dataset for realistic perception, reconstruction and generation,'' in \emph{Proceedings of the IEEE/CVF Conference on Computer Vision and Pattern Recognition}, 2023, pp. 803--814.

\bibitem{arc}
C.~Lu, S.~Xia, M.~Shao, and Y.~Fu, ``Arc-support line segments revisited: An efficient high-quality ellipse detection,'' \emph{IEEE Transactions on Image Processing}, vol.~29, pp. 768--781, 2019.

\bibitem{arc2}
C.~Lu, S.~Xia, W.~Huang, M.~Shao, and Y.~Fu, ``Circle detection by arc-support line segments,'' in \emph{Proc. IEEE International Conference on Image Processing}, 2017, pp. 76--80.

\bibitem{ippe}
T.~Collins and A.~Bartoli, ``Infinitesimal plane-based pose estimation,'' \emph{International journal of computer vision}, vol. 109, no.~3, pp. 252--286, 2014.

\bibitem{yibo}
Y.~Liu, H.~Schofield, and J.~Shan, ``Navigation of a self-driving vehicle using one fiducial marker,'' in \emph{Proc. IEEE International Conference on Multisensor Fusion and Integration for Intelligent Systems}, 2021, pp. 1--6.

\bibitem{ch2020}
S.-F. Ch’ng, N.~Sogi, P.~Purkait, T.-J. Chin, and K.~Fukui, ``Resolving marker pose ambiguity by robust rotation averaging with clique constraints,'' in \emph{Proc. IEEE International Conference on Robotics and Automation}, 2020, pp. 9680--9686.

\bibitem{station}
M.~Vaidis, P.~Giguère, F.~Pomerleau, and V.~Kubelka, ``Accurate outdoor ground truth based on total stations,'' in \emph{2021 18th Conference on Robots and Vision (CRV)}, 2021, pp. 1--8.

\bibitem{cal1}
P.~Furgale, J.~Rehder, and R.~Siegwart, ``Unified temporal and spatial calibration for multi-sensor systems,'' in \emph{Porc. IEEE/RSJ International Conference on Intelligent Robots and Systems}, 2013, pp. 1280--1286.

\bibitem{edge}
A.~D. Sappa and M.~Devy, ``Fast range image segmentation by an edge detection strategy,'' in \emph{Proc. International Conference on 3-D Digital Imaging and Modeling}, 2001, pp. 292--299.

\bibitem{single}
S.~C. Johnson, ``Hierarchical clustering schemes,'' \emph{Psychometrika}, vol.~32, no.~3, pp. 241--254, 1967.

\bibitem{floam}
H.~Wang, C.~Wang, C.-L. Chen, and L.~Xie, ``F-loam: Fast lidar odometry and mapping,'' in \emph{2021 IEEE/RSJ International Conference on Intelligent Robots and Systems (IROS)}.\hskip 1em plus 0.5em minus 0.4em\relax IEEE, 2021, pp. 4390--4396.

\bibitem{aloam}
\BIBentryALTinterwordspacing
Q.~Tong and C.~Shaozu. {A-LOAM: Advanced implementation of loam}. (2019). [Online]. Available: \url{” https://github.com/HKUST-Aerial-Robotics/A-LOAM}
\BIBentrySTDinterwordspacing

\bibitem{multiview}
Z.~Gojcic, C.~Zhou, J.~D. Wegner, L.~J. Guibas, and T.~Birdal, ``Learning multiview 3d point cloud registration,'' in \emph{Proc. of the IEEE/CVF conference on computer vision and pattern recognition}, 2020, pp. 1759--1769.

\bibitem{features}
M.~Khoury, Q.-Y. Zhou, and V.~Koltun, ``Learning compact geometric features,'' in \emph{Proc. of the IEEE international conference on computer vision}, 2017, pp. 153--161.

\bibitem{access}
Y.~Liu, J.~Shan, and H.~Schofield, ``Improvements to thin-sheet 3d lidar fiducial tag localization,'' \emph{IEEE Access}, vol.~12, pp. 124\,907--124\,914, 2024.

\bibitem{lpr}
Y.~Liu, J.~Shan, A.~Haridevan, and S.~Zhang, ``L-pr: Exploiting lidar fiducial marker for unordered low overlap multiview point cloud registration,'' \emph{IEEE Transactions on Instrumentation and Measurement}, 2025.

\bibitem{barfoot}
T.~D. Barfoot, \emph{State estimation for robotics}.\hskip 1em plus 0.5em minus 0.4em\relax Cambridge University Press, 2017.

\bibitem{tagslam}
B.~Pfrommer and K.~Daniilidis, ``Tagslam: Robust slam with fiducial markers,'' \emph{arXiv preprint arXiv:1910.00679}, 2019.

\bibitem{lasernet}
G.~P. Meyer, A.~Laddha, E.~Kee, C.~Vallespi-Gonzalez, and C.~K. Wellington, ``Lasernet: An efficient probabilistic 3d object detector for autonomous driving,'' in \emph{Proc. IEEE/CVF Conference on Computer Vision and Pattern Recognition}, 2019, pp. 12\,677--12\,686.

\bibitem{colormap}
P.~Kovesi, ``Good colour maps: How to design them,'' \emph{arXiv preprint arXiv:1509.03700}, 2015.

\bibitem{PCL}
R.~B. Rusu and S.~Cousins, ``3d is here: Point cloud library (pcl),'' in \emph{Proc. IEEE International Conference on Robotics and Automation}, 2011, pp. 1--4.

\bibitem{gb}
R.~A. Haddad, A.~N. Akansu \emph{et~al.}, ``A class of fast gaussian binomial filters for speech and image processing,'' \emph{IEEE Transactions on Signal Processing}, vol.~39, no.~3, pp. 723--727, 1991.

\bibitem{nicp}
J.~Serafin and G.~Grisetti, ``Nicp: Dense normal based point cloud registration,'' in \emph{Proc. IEEE/RSJ International Conference on Intelligent Robots and Systems}, 2015, pp. 742--749.

\bibitem{icp}
K.~S. Arun, T.~S. Huang, and S.~D. Blostein, ``Least-squares fitting of two 3-d point sets,'' \emph{IEEE Transactions on pattern analysis and machine intelligence}, no.~5, pp. 698--700, 1987.

\bibitem{map3}
K.~Koide, S.~Oishi, M.~Yokozuka, and A.~Banno, ``Scalable fiducial tag localization on a 3d prior map via graph-theoretic global tag-map registration,'' in \emph{Proc. of IEEE/RSJ International Conference on Intelligent Robots and Systems}.\hskip 1em plus 0.5em minus 0.4em\relax IEEE, 2022, pp. 5347--5353.

\bibitem{vins}
T.~Qin, P.~Li, and S.~Shen, ``Vins-mono: A robust and versatile monocular visual-inertial state estimator,'' \emph{IEEE transactions on robotics}, vol.~34, no.~4, pp. 1004--1020, 2018.

\bibitem{rusu}
R.~B. Rusu, ``Semantic 3d object maps for everyday manipulation in human living environments,'' \emph{KI-K{\"u}nstliche Intelligenz}, vol.~24, no.~4, pp. 345--348, 2010.

\bibitem{bounding}
P.~Schneider and D.~H. Eberly, \emph{Geometric tools for computer graphics}.\hskip 1em plus 0.5em minus 0.4em\relax Elsevier, 2002.

\bibitem{cloudcompare}
D.~Girardeau-Montaut \emph{et~al.}, ``Cloudcompare,'' \emph{France: EDF R\&D Telecom ParisTech}, vol.~11, no.~5, 2016.

\bibitem{isam2}
M.~Kaess, H.~Johannsson, R.~Roberts, V.~Ila, J.~Leonard, and F.~Dellaert, ``isam2: Incremental smoothing and mapping with fluid relinearization and incremental variable reordering,'' in \emph{Proc. of IEEE International Conference on Robotics and Automation}, 2011, pp. 3281--3288.

\bibitem{gtsam}
\BIBentryALTinterwordspacing
F.~Dellaert and M.~Kaess, \emph{Factor Graphs for Robot Perception}.\hskip 1em plus 0.5em minus 0.4em\relax Foundations and Trends in Robotics, Vol. 6, 2017. [Online]. Available: \url{http://www.cs.cmu.edu/~kaess/pub/Dellaert17fnt.pdf}
\BIBentrySTDinterwordspacing

\bibitem{super}
\BIBentryALTinterwordspacing
Supervisely. {Supervisely Computer Vision platform}. (2023). [Online]. Available: \url{https://supervisely.com}
\BIBentrySTDinterwordspacing

\bibitem{mapmerge}
\BIBentryALTinterwordspacing
Y.~Liu, J.~Shan, and H.~Schofield, ``Fiducial tag localization on a 3d lidar prior map,'' 2024. [Online]. Available: \url{https://arxiv.org/abs/2209.01072}
\BIBentrySTDinterwordspacing

\bibitem{dense}
S.~Sato, Y.~Yao, T.~Yoshida, T.~Kaneko, S.~Ando, and J.~Shimamura, ``Unsupervised intrinsic image decomposition with lidar intensity,'' in \emph{Proceedings of the IEEE/CVF Conference on Computer Vision and Pattern Recognition}, 2023, pp. 13\,466--13\,475.

\bibitem{lidarnerf}
S.~Huang, Z.~Gojcic, Z.~Wang, F.~Williams, Y.~Kasten, S.~Fidler, K.~Schindler, and O.~Litany, ``Neural lidar fields for novel view synthesis,'' in \emph{Proceedings of the IEEE/CVF International Conference on Computer Vision}, 2023, pp. 18\,236--18\,246.

\bibitem{lidarnerf2}
Z.~Zheng, F.~Lu, W.~Xue, G.~Chen, and C.~Jiang, ``Lidar4d: Dynamic neural fields for novel space-time view lidar synthesis,'' in \emph{Proceedings of the IEEE/CVF Conference on Computer Vision and Pattern Recognition}, 2024, pp. 5145--5154.

\end{thebibliography}

\chapter*{List of Publications \label{pub}}
\addcontentsline{toc}{chapter}{List of Publications}

\begin{enumerate}
    \item \textbf{Y. Liu}, H. Schofield and J. Shan, "Navigation of a Self-Driving Vehicle Using One Fiducial Marker," IEEE International Conference on Multisensor Fusion and Integration for Intelligent Systems (\textbf{MFI}), Karlsruhe, Germany, 2021, pp. 1-6, 
    DOI: \url{10.1109/MFI52462.2021.9591194}. 
    
    \item \textbf{Y. Liu}, H. Schofield and J. Shan, "Intensity Image-Based LiDAR Fiducial Marker System," in IEEE Robotics and Automation Letters (\textbf{RA-L}), vol. 7, no. 3, pp. 6542-6549, 2022, DOI: \url{10.1109/LRA.2022.3174971}. 
    
    \item \textbf{Y. Liu}, A. Haridevan, H. Schofield and J. Shan, "Application of Ghost-DeblurGAN to Fiducial Marker Detection," IEEE/RSJ International Conference on Intelligent Robots and Systems (\textbf{IROS}), Kyoto, Japan, 2022, pp. 6827-6832, DOI: \url{10.1109/IROS47612.2022.9981701}.

    \item \textbf{Y. Liu}, K. Zhu, G. Wu. Y. Ren, B. Liu, Y. Liu, and J. Shan, "MV-DeepSDF: Implicit Modeling with Multi-Sweep Point Clouds for 3D Vehicle Reconstruction in Autonomous Driving," IEEE/CVF International Conference on Computer Vision (\textbf{ICCV}), Paris, France, 2023, pp. 8272-8282, DOI: \url{10.1109/ICCV51070.2023.00763}.  

    \item \textbf{Y. Liu*}, Z. Yang*, G. Wu. Y. Ren, K. Lin, B. Liu, Y. Liu, and J. Shan, "VQA-Diff: Exploiting VQA and Diffusion for Zero-Shot Image-to-3D Vehicle Asset Generation in Autonomous Driving," European Conference on Computer Vision (\textbf{ECCV}), Milan, Italy, 2024. pp. 323-340, DOI: \url{https://doi.org/10.1007/978-3-031-72848-8_19}.  

    \item \textbf{Y. Liu}, J. Shan, and H. Schofield, "Improvements to Thin-Sheet 3D LiDAR Fiducial Tag Localization,", \textbf{IEEE ACCESS}, vol. 12, pp. 124907-124914, 2024, DOI: \url{10.1109/ACCESS.2024.3451404}. 

    \item \textbf{Y. Liu}, J. Shan, A. Haridevan, and S. Zhang, “L-PR: Exploiting LiDAR Fiducial Marker for Unordered Low Overlap Multiview Point Cloud Registration,” IEEE Transactions on Instrumentation \& Measurement (\textbf{TIM}), in press, 2025

    \item S. Zhang, J. Shan, and \textbf{Y. Liu}, "Approximate Inference Particle Filtering for Mobile Robot SLAM," in IEEE Transactions on Automation Science and Engineering (\textbf{T-ASE}), pp. 1-12, 2024, DOI: \url{10.1109/TASE.2024.3475735}. 
    
     \item Z. Yang*, \textbf{Y. Liu*}, G. Wu, T. Cao, Y. Ren, Y. Liu, and B. Liu, “Learning Effective NeRFs and SDFs Representations with 3D GANs for Object Generation,” \textbf{NeurIPS Workshop}, 2024 (* co-first author).

    \item S. Zhang, J. Shan, and \textbf{Y. Liu}, "Variational Bayesian Estimator for Mobile Robot Localization With Unknown Noise Covariance," in IEEE/ASME Transactions on Mechatronics (\textbf{T-MECH}), vol. 27, no. 4, pp. 2185-2193, 2022, DOI: \url{10.1109/TMECH.2022.3161591}. 

    \item S. Zhang, J. Shan, and \textbf{Y. Liu}, "Particle Filtering on Lie Group for Mobile Robot Localization With Range-Bearing Measurements," in IEEE Control Systems Letters (\textbf{L-CSS}), vol. 7, pp. 3753-3758, 2023, DOI: \url{10.1109/LCSYS.2023.3340419}.

    \item T. Liao, A. Haridevan, \textbf{Y. Liu}, and J. Shan, “Autonomous Vision-Based UAV Landing with Collision Avoidance Using Deep Learning,” Science and Information Conference (\textbf{SAI}), pp. 79–87, 2022, DOI: \url{10.1007/978-3-031-10464-0_6}
    
\end{enumerate}




    





\end{document}